\documentclass{article}
\usepackage{graphicx}
\usepackage{authblk}
\usepackage{verbatim}
\usepackage{multirow}
\usepackage{array, tabularx}
\usepackage{subfigure}
\usepackage{caption}
\usepackage{url}
\usepackage{booktabs} 
\usepackage{xspace}
\usepackage{float}
\usepackage{multirow}
\usepackage{alltt}
\usepackage{longtable}
\usepackage{color}
\usepackage{xcolor,colortbl}
\usepackage{pifont}
\usepackage{makecell}
\usepackage{multicol}
\usepackage{comment}

\usepackage[a4paper, total={6.7in, 8in}]{geometry}
\usepackage{xr}
\usepackage[utf8]{inputenc}
\usepackage[T1]{fontenc}
\usepackage{amsthm}
\usepackage{amsmath}
\usepackage{amssymb}
\usepackage{siunitx}

\usepackage{url}
\usepackage{booktabs} 
\usepackage{xspace}
\usepackage{float}
\usepackage{multirow}
\usepackage{alltt}
\usepackage{color}
\usepackage{pifont}
\usepackage{makecell}
\usepackage{multicol}
\usepackage[super]{nth}
\newcommand\cparagraph[1]{\vspace{0.6mm}\noindent\textbf{#1.}}
\let\origref\ref
\def\ref#1{\textnormal{\origref{#1}}}

\newcommand{\cmark}{\ding{51}}
\newcommand{\xmark}{\ding{55}}

\title{Generative AI voting: fair collective choice is resilient to LLM biases and inconsistencies}

\author[1]{Srijoni Majumdar$^*$}
\author[2]{Edith Elkind}
\author[1]{Evangelos Pournaras}

\affil[1]{School of Computer Science, University of Leeds, Leeds,  LS29JT UK\ \ \ \ \ \ \ \ \ \ \ \ \ \ \ \ \ \ \ \ \   \ \ \ \ \ \ \ \ \ \ \ \ \ \ \ \ \ \ \ \ \ \ \ \ \   \ \ \ \ E-mails: \{s.majumdar,e.pournaras\}@leeds.ac.uk}

\affil[2]{Department of Computer Science, Northwestern University, Evanston,  IL 60208 US\ \ \ \ \ \   \ \ \ \ \ \ \ \ \ \ \ \ \ \ \ \ \ \ \ \ \ \ \ \ \ \ \ \ \   \ \ \ E-mails:\ edith.elkind@northwestern.edu}

\date{} 

\begin{document}
	\maketitle
	
	\begin{abstract}
		\footnotetext[1]{$^*$ Corresponding author: Srijoni Majumdar, School of Computer Science, University of Leeds, Leeds, UK, E-mail: s.majumdar@leeds.ac.uk}

Recent breakthroughs in generative artificial intelligence (AI) and large language models (LLMs) unravel new capabilities for AI personal assistants to overcome cognitive bandwidth limitations of humans, providing decision support or even direct representation of abstained human voters at large scale.  However, the quality of this representation and what underlying biases manifest when delegating collective decision making to LLMs is an alarming and timely challenge to tackle.  By rigorously emulating more than \textcolor{black}{>50K LLM voting personas in 363 real-world voting elections}, we disentangle how AI-generated choices differ from human choices and how this affects collective decision outcomes. Complex preferential ballot formats show significant inconsistencies compared to simpler majoritarian elections, which demonstrate higher consistency. Strikingly, proportional ballot aggregation methods such as equal shares prove to be a win-win: \textcolor{black}{fairer voting outcomes for humans and fairer AI representation, especially for voters likely to abstain.} This novel underlying relationship proves paramount for building democratic resilience in scenarios of low voters turnout by voter fatigue: abstained voters are mitigated via AI representatives that recover representative and fair voting outcomes. These interdisciplinary insights provide decision support to policymakers and citizens for developing safeguards and policies for risks of using AI in democratic innovations.  

\end{abstract}



\noindent

\section{Introduction} \label{sec:introduction}

Recent advances in artificial intelligence (AI) provide new, unprecedented opportunities for citizens to scale up participation in digital democracy~\cite{Helbing2015society,helbing2023democracy,Pournaras2020}. Generative AI in particular, such as large language models (LLMs), has the potential to overcome human cognitive bandwidth limitations and digitally assist citizens to deliberate and decide about public matters at scale~\cite{koster2022human,heersmink2024use,Gudino2024,Small2023,rattanasevee2024direct}. This is by articulating, summarizing and even providing syntheses of complex opinions~\cite{argyle2023leveraging,dryzek2019crisis,Small2023}, with a potential to mitigate for the voter fatigue and reduced voter turnout~\cite{ercan2013democratic,dryzek2019crisis,aichholzer2020experience}, while fostering common ground for compromises, consensus and lower polarization~\cite{ercan2013democratic,navarrete2024understanding,dryzek2019crisis,park2023electoral,koster2022human}. However, understanding the implications and risks of using large language models for decision support, recommendations or even direct representation of human voters is a pressing challenge~\cite{Pournaras2023,gunjal2024detecting,ram2023context}. 

\begin{figure}[!htb]
    \centering
    \includegraphics[scale = 0.328]{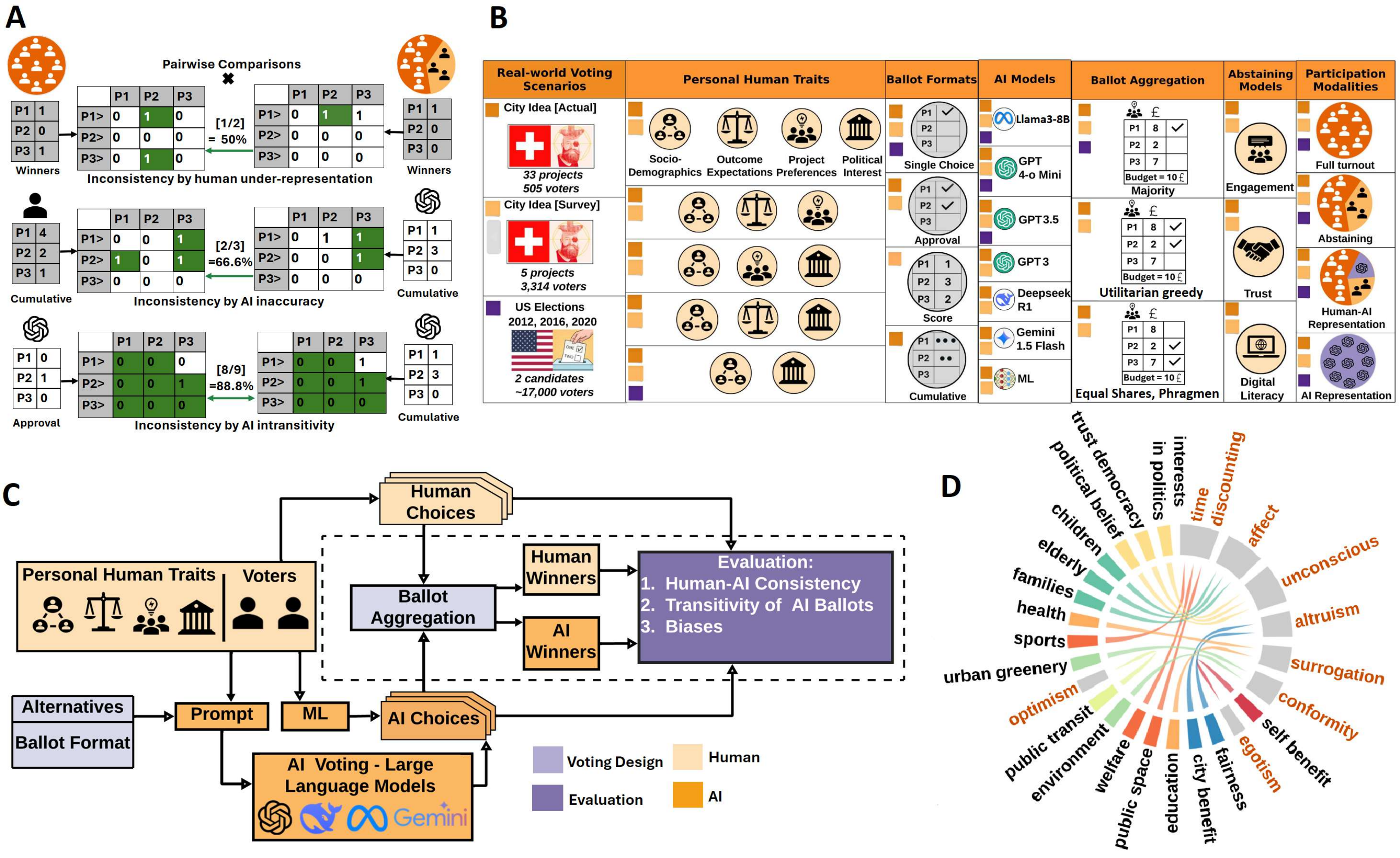}
    \caption{\textbf{An overview of the studied generative AI  voting framework.} (A) Three manifestations of choice inconsistency are distinguished, measured using Condorcet pairwise matches: (i) inconsistencies by under-representation as a result of low voters turnout, (ii) inconsistencies by inaccuracy of AI choice to approximate human choice and (iii) inconsistency by intransitivity of AI choices over different ballot formats. P1, P2, and P3 are projects put up for voting and received the score 4, 2, and 1 respectively by a voter; in the case of approval voting, the scores are 0 or 1. (B) The factorial design with the 7 studied dimensions: (i) Real-world voting scenarios in the context of participatory budgeting and national elections. (ii) Various combinations of personal human traits (features) based on which AI voting personas are created. (iii) Four ballot formats. (iv) Seven AI models, six large language models and a predictive machine learning model (benchmark). (v) Ballot aggregation methods for elections and participatory budgeting. (vi) The three abstaining models that are based for engagement, digital literacy and trust. (vii) Participation modalities ranging from exclusive human participation of varying turnout to mixed populations of humans and AI representatives of abstained voters. The studied combinations for each voting scenario are marked with different colors, see also Table~\ref{table:persona1}. (C) The framework of generative AI voting. For each voter in the real-world voting scenario, a prompt is given to large language models to construct the voting persona. The input is the personal human traits, the voting options, and the ballot format, with instructions for the voting persona on how to make a choice. This choice is the output of the persona. Both human and AI choices are aggregated using a ballot aggregation method. The inconsistencies of individual and collective choices for humans and AI personas are assessed, along with potential biases that explain these inconsistencies.  (D) The personal human traits are mapped to cognitive biases. Section~\ref{sec:bias} illustrates the origin of choice inconsistencies to potential cognitive biases. }
    \label{fig:factorial}
\end{figure}

\noindent \cparagraph{Unraveling inconsistencies  in generative AI voting}
We disentangle the inconsistencies of large language models when employed to generate {\em individual voter choices} and assess the ways in which these inconsistencies shape the {\em collective choice}. In particular, we study three manifestations of choice inconsistency as shown in Figure~\ref{fig:factorial}a: 

\begin{enumerate}
\item \textbf{Inconsistency in voting outcomes by under-representation due to low human voters turnout}.\\It is measured by the dissimilarity in collective choices  when voters abstain compared to when they participate;
\item \textbf{Inconsistency by inaccurate approximation of human choice by AI}. \\ It is measured by the dissimilarity between AI and human choices, and;
\item \textbf{Inconsistency by intransitivity~\cite{Tversky1969,hausladen2023legitimacy} of AI choice}. \\It is measured by the dissimilarity in AI choices across different ballot formats.
\end{enumerate}

\noindent Since intransitivity is also present in human choices, particularly in polarized contexts~\cite{hausladen2023legitimacy} that are often shaped by biases~\cite{Schwalbe2020,Dandekar2013}, it is reasonable to expect similar inconsistencies to appear in LLM choices. Whether potential biases that explain the inconsistencies between human and AI choices are of a different nature than the ones between different input voting methods is an open question studied in this article. We rigorously measure such inconsistencies with a single universal approach grounded in social choice theory~\cite{navarrete2024understanding,kulakowski2022similarity}. It exhaustively characterizes the similarity of the two choices (individual or collective) by counting the relative number of Condorcet pairwise matches; see Section~\ref{sec:consistency} for further information. We also explore causal links of these inconsistencies to potential cognitive biases triggered by the input to large language models based on which choices are made.

\cparagraph{Generative AI voting: a converging technological advance with inevitable challenges} Large language models have been applied to predict election outcomes using sensitive demographic information reflecting the political profile of individuals~\cite{argyle2023out}. \textcolor{black}{They have also been employed to predict pairwise comparisons of proposals for constitutional changes~\cite{Gudino2024} and to facilitate deliberation by summarizing opinions expressed in free-form text~\cite{Fish2023,gudino2025prompt}.} \textcolor{black}{However, little is known about whether this AI predictive capability can expand to voting with complex ballot formats that involve more options to choose from~\cite{heersmink2024use}}. Participatory budgeting~\cite{pierczynski2021proportional} is one such process put under scrutiny in this article. Here city authorities distribute a public budget by letting citizens propose their own project ideas, which they vote for and often implement themselves~\cite{Bartocci2023}. Projects may be pertinent to different impact areas (e.g., environment, culture, welfare), beneficiaries (e.g., elderly, children) and can have different costs~\cite{maharjan2024fair}. Voters can approve, rank or distribute points over their preferred projects, while winners are elected based on the popularity of the projects (\emph{utilitarian greedy}) or based on a proportional representation of the voters' preferences (\emph{equal shares} or Phragmen's rule)~\cite{Fairstein2023,Yang2023}. So far, AI assistance for such processes is limited. \textcolor{black}{
A participatory budgeting process has been emulated using AI agents to examine the feasibility of consensus building by assisting voters in electing winners through a reinforcement learning framework~\cite{Majumdar2023}. This work  focus  on promoting compromises using rewards to reach consensus instead of applying a ballot aggregation method.
In the context of vote prediction, Yang et al. recently conducted a study in which large language models (LLMs) emulate voters to generate preferences and to examine the diversity of preference generation through a lab experiment involving 180 university students~\cite{yang2024llm}. However, the study does not evaluate the impact of LLM-based voting on real-world participatory processes. It does not also address the influence of voters who are more likely to abstain on voting outcomes.} Moreover, the scope and citizens' engagement in participatory budgeting campaigns remain to a large extent a one-shot and and rooted in local civic cultures~\cite{gersbach2024forms,ercan2013democratic}.  With such complexity and degree of design freedom, scaling up participatory budgeting turns into the ultimate democratic blueprint to assess capabilities and risks of generative AI voting. We do not make a normative statement about the use of (generative) AI voting, although prominent scholars have explored this plausible future; for instance, Augmented Democracy by Hidalgo et al.~\cite{augmented}, along with recent research~\cite{yang2024llm,argyle2023out} that explore the potential for scaling up direct citizen participation in decision-making, rather than over-relying on human representatives. This scenario though seems highly relevant as a result of an inevitable technological convergence of AI and digital voting, for instance, allowing personal and localized AI assistants to interoperate with the Application Programming Interfaces (APIs) of digital voting platforms. Understanding the implications of such capabilities and preparing safeguards to protect democracy and mitigate the consequences of AI risks comes with merit and urgency, which we address in our work. 

 \cparagraph{How resilient representative voting outcomes are with generative AI} \textcolor{black}{We hypothesize that a proportional ballot aggregation method can build up {\em resilience} for representative voting outcomes if AI representatives are used for human voters who would otherwise abstain or lack the capacity to actively participate (see participation modalities in Figure~\ref{fig:factorial}b). In other words, {\em we examine whether inconsistencies in collective voting outcomes resulting from low voter turnout (see Figure~\ref{fig:factorial}a) are greater than those arising from generative AI representatives of abstaining voters using different ballot aggregation methods}.} This process of consistency recovery through AI representation indicates the degree to which the original outcome can be preserved. We refer to this as the {\em resilience} of a voting outcome in scenarios of low voter turnout and mixed populations composed of humans and AI representatives of abstaining voters.
 
\cparagraph{Disentangling the role of voting design in generative AI voting} \textcolor{black} {The inconsistencies of generative AI voting, their association with ballot formats and aggregation methods, along with the potential AI and human biases explaining these inconsistencies, are systematically studied here for the first time using a novel factorial design based on real-world empirical evidence. It consists of seven dimensions (see Figure~\ref{fig:factorial}b) designed to emulate AI voting representation, generate individual choices, and aggregate them into a collective voting outcome.}  

\begin{enumerate}
\item \emph{Real-world voting scenarios} - election datasets from the 2012, 2016, and 2020 US national elections~\cite{debell2009computing} as well as data from the 2023 participatory budgeting campaign of `City Idea' in Aarau, Switzerland~\cite{CityIdeaReport2025} are studied. The latter dataset includes two voting scenarios: a hypothetical one provided to voters before voting via a \emph{survey}, and the \emph{actual} voting data. The datasets from Aarau also contain demographic data and personal information traits collected before and after voting through pre-voting and post-voting surveys. This information is used to capture individual voter context when emulating AI representations through prompt engineering in large language models. These three datasets cover a wide range of ballot types (e.g., single choice ballots for US elections and approval or score/cumulative ballots for the Aarau voting), voting alternatives and numbers of voters to experiment with; see Figure~\ref{fig:factorial}b and Section~\ref{sec:emul}.
\item \emph{Personal human traits} - for each voter, multiple incremental levels of additional information are provided as input to large language models to generate ballots. This includes (i) socio-demographic characteristics (e.g., gender, age, education, household size), (ii) political interests (e.g., ideological profile, political beliefs), (iii) personal attitudes toward project preferences (e.g., prioritization of green initiatives, sustainable transport, elderly care facilities), and (iv) expectations for the qualities of voting outcomes (e.g., favoring cost-effective winning projects, popular projects, or projects with proportional representation of citizens’ preferences). These traits are obtained from voter feedback surveys, which are linked to actual voting behavior in the Aarau voting scenarios, or collected during voter registration for the US elections (Tables~\ref{table:prologue1}–\ref{table:prologue5}). Not all traits are available across all datasets (see the distribution of extracted human traits in Figure~\ref{fig:factorial}b).
\item \emph{Ballot formats} - four methods with incremental levels of complexity and expressiveness are compared~\cite{hausladen2023legitimacy,Welling2023fair}. These include single choice for all voting scenarios, n-{\em approvals} (`n' of projects approved), score (assigning a preference score from a specified range [1 to 5] to each option) and cumulative voting (distributing a number of points (i.e., 10) over the options)~\cite{ebrahimnejad2017survey,Cooper2007,Skowron2020} for the participatory budgeting scenarios.
\item \emph{AI models} -  generative and predictive AI methods have been used to emulate AI representation. Six large language models~\cite{brown2020language,touvron2023llama} are assessed along with a more mainstream predictive machine learning (ML) model used as a benchmark. \texttt{GPT 4-o Mini}, \texttt{GPT3}, \texttt{GPT3.5} , \texttt{Deepseek R1}, \texttt{Gemini 1.5 Flash}, and \texttt{Llama3-8B} are chosen, covering a wide spectrum of capabilities in open-source and proprietary generative AI (more details on prompts and choice generation in Sections~\ref{sec:prompt})~\cite{zhao2024explainability}. The predictive ML benchmark is built by using personal human traits as features to predict ballots using neural networks (more details in Section~\ref{sec:ml})~\cite{rocha2007evolution}.
\item \emph{Ballot aggregation methods} - majority aggregation is used to determine the collective outcome of the US elections. For the participatory budgeting scenarios, the  utilitarian greedy method, the method of equal shares~\cite{peters2021proportional}, and Phragmén's sequential rule~\cite{brill2024phragmen} are employed. {\em Utilitarian greedy} simply selects the next most popular project, the one with the highest number of votes, provided the available budget is not exhausted. {\em Equal shares} ensures proportional representation of voters' preferences by dividing the budget equally among voters as endowments. Voters can only use their share to fund projects they voted for. The method evaluates all project options, starting with those receiving the most votes, and selects a project if it can be funded using the budget shares of its supporters. A full explanation of equal shares is beyond the scope of this article and can be found in earlier work~\cite{peters2021proportional,aziz2021participatory,faliszewski2023participatory}. In practice, equal shares may sacrifice an expensive popular project in favor of several low-cost projects that collectively satisfy more voters’ preferences~\cite{maharjan2024fair,Fairstein2023}. Because of this effect, it is likely that consistency measurements based on the pairwise similarity yield higher values for equal shares. This is the reason we control for the number of winning projects in equal shares by counting a subset of the most popular winning projects, which is equal in number with the winners of the utilitarian greedy method.  Phragmén's sequential rule is another proportional aggregation method that balances fairness and representation between groups, in contrast to equal shares, which emphasizes fair representation within groups by ensuring that at least one voter from each group is represented~\cite{peters2020proportional}. Equal shares was the method actually used in the City Idea campaign to select winners~\cite{maharjan2024fair}, providing strong realism for the findings of this study.
\item \emph{Abstaining models} -  three types of abstaining voters: (i) those with low digital skills, limiting their ability to participate online~\cite{aichholzer2020experience,vassil2011bottleneck} and often leading to low turnouts~\cite{rattanasevee2024direct,Gudino2024,lorenz2023systematic,devine2024does}; (ii) those with low political engagement~\cite{hylton2023voter,lane2018planning}; and (iii) those who distrust institutions~\cite{wang2016political,belanger2017political,devine2024does}. Using pre-voting and post-voting survey questions from the City Idea participatory budgeting campaign (Tables~\ref{table:prologue3}–\ref{table:prologue5}), we identify proxies for these abstaining profiles~\cite{lorenz2023systematic,halpern2023representation} and divide voters into quartiles to distinguish voters who are likely to abstain. The share of the population that meet the criteria of the three abstaining models is 36.1\%, 48.3\% and 27.4\% respectively.
\item \emph{Participation modalities} - we assess the consistency of voting scenarios with full and low turnouts of human voters, partial/full AI representation of abstained voters, and AI representation of the whole human population. 
\end{enumerate}

\noindent The dimensions of the factorial design are illustrated in Table~\ref{table:persona1} and the studied combinations are marked with the colored boxes in Figure~\ref{fig:factorial}b. This broad spectrum of analysis based on real-world evidence allows us to generalize the findings of the study and make them relevant for a broad spectrum of research communities and policymakers. 


\cparagraph{Assessing generative AI voting in action} Voting personas are constructed using input prompts of large language models as depicted in Figure~\ref{fig:factorial}c. This designed process aims to emulate the three voting scenarios with the different settings of Figure~\ref{fig:factorial}b. Each input prompt consists of a standardized description of the voter's profile (see Section~\ref{sec:study}) and an instruction to vote according to the ballot format. The consistency between the individual and collective real-world choices of humans and AI personas is compared for the first time by measuring the Condorcet pairwise matches as shown in Figure~\ref{fig:factorial}a~\cite{navarrete2024understanding,kulakowski2022similarity,hausladen2023legitimacy}. These consistency values are then becoming the dependent variable to predict using the personal human traits as independent variables (features), fed into a neural network (see Section~\ref{sec:interpret-AI}). Based on a systematic mapping of human personal traits to cognitive biases as illustrated in Figure~\ref{fig:factorial}c (see Section~\ref{sec:bias} for more detail), this prediction model causally explains the human traits that contribute to inconsistencies and the potential underlying biases that explain these inconsistencies. This novel analysis is designed to provide a significant conceptual advance in understanding how voting design reinforces or mitigates different AI biases in real-world practice.

\section{Results}\label{sec:results}

The following three key results are illustrated in this article: 

\cparagraph{1} \textcolor{black}{Fair voting methods to elect winners are more resilient to inconsistencies of AI to accurately estimate human choice, demonstrating a striking underlying win-win relationship: fairer voting outcomes for humans with fairer human representation by AI (Figure~\ref{fig:consistency-biases} and~\ref{fig:consistency-recovery}). These inconsistencies are particularly prominent in complex ballot formats with a large number of alternatives, while simple majoritarian voting tends to be highly consistent. AI intransitivity across ballot formats is higher than that of humans, with a greater impact on collective choice when the number of alternatives is large (Figure~\ref{fig:transitivity}).}

\cparagraph{2} \textcolor{black}{AI representation is more effective for a voter who is likely to abstain than for an arbitrary voter, particularly under fair collective choice (Figure~\ref{fig:consistency-recovery}). Abstaining voters result in a representation deficit that is restored by AI, while AI representation over arbitrary voters mainly has a noise-reduction effect on the voting outcome. }

\cparagraph{3} \textcolor{black}{Features of abstaining voters related to their low engagement, digital literacy, and trust explain the consistency of their AI representation and the transitivity of ballot formats (Figure~\ref{fig:biases}). Affect and unconscious biases explain the (in)consistency of human-AI choice, while  time-discounting biases explain the transitivity of AI choice across ballot formats.}

\subsection{Fair collective choice is resilient to human-AI inconsistencies}\label{sec:inconsistency-resilience}

\begin{figure}[!htb]
    \centering
    
    \includegraphics[scale = 0.18]{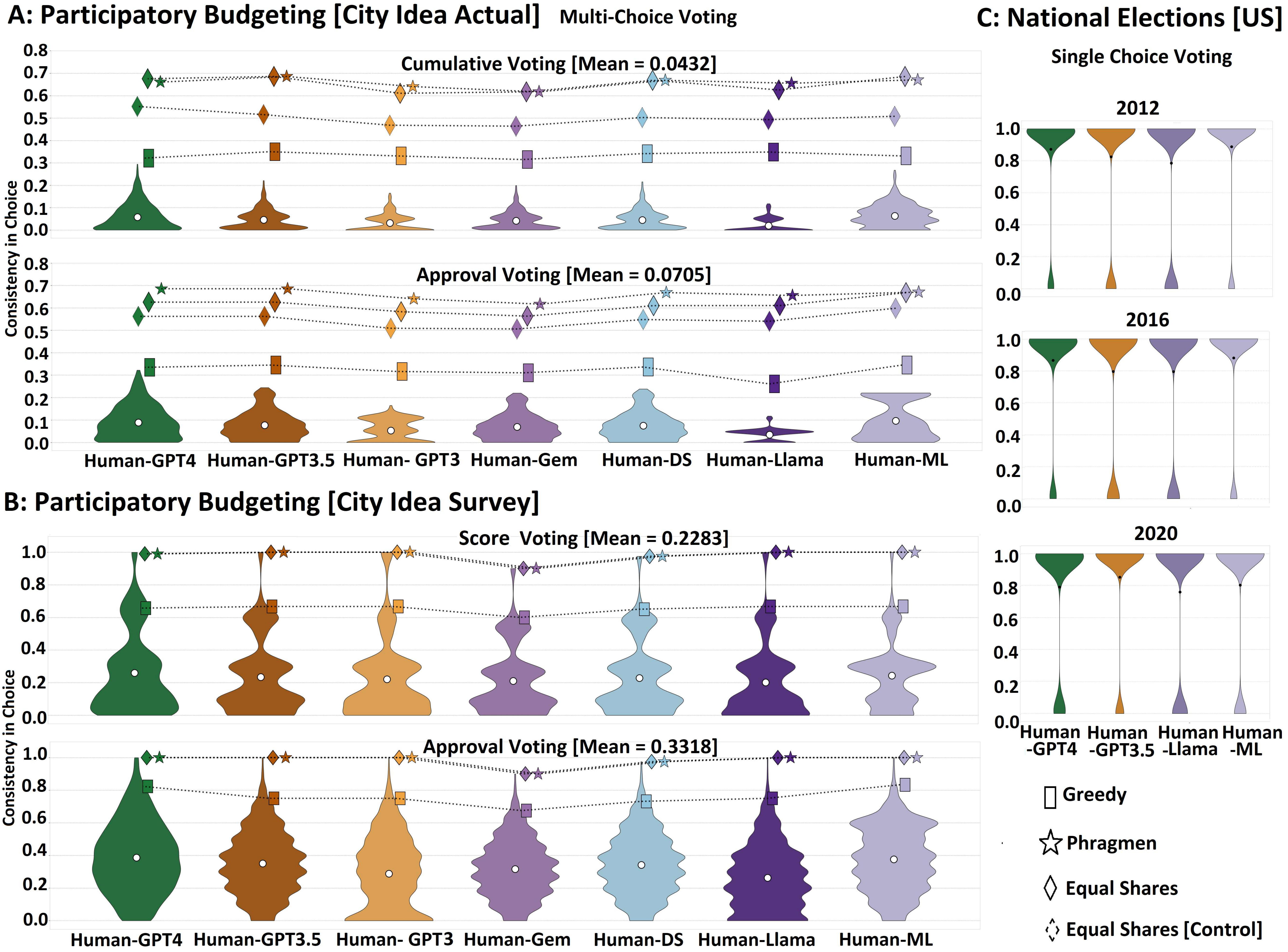}
    \caption{\textbf{Choice by large language models is consistent to humans for single choice majoritarian elections, however accuracy drops for more complex ballots with larger number of alternatives as in the case of participatory budgeting. Strikingly, accuracy  of collective choice is significantly higher than individual choice, particularly for the fairer ballot aggregation rules of equal shares and Phragmén's. \texttt{GPT 4-0 Mini} shows the highest consistency and \texttt{Llama3-8B} the lowest among the large language models, which though remain inferior to a predictive machine learning model}. The mean consistency (y-axis) for different population of voters (10\%, 25\%, 75\% and 100\%) in individual and collective choice is shown for six large language models ({\texttt{GPT 4-o Mini} (GPT 4) }, {\texttt{GPT3.5} },  {\texttt{GPT3} }, \texttt{Gemini 1.5 Flash}  (Gem),  { \texttt{Deepseek R1}  (DS)}  and {\texttt{Llama3-8B} (Llama)})   along with the predictive AI model ({\em ML})(x-axis), across three real-world voting scenarios: The participatory budgeting campaign of City Idea, (A) actual and (B) survey, as well as (C) the US national elections of 2012, 2016 and 2020. For participatory budgeting, the ballot formats of cumulative/score (top) and approval (bottom) are shown, including the ballot aggregation methods of equal shares, Phragmén's and utilitarian greedy. For the actual voting of City Idea, the accuracy of equal shares is calculated for all winners and a controlled number of winners (as many as utilitarian greedy) for a fairer comparison.}
    \label{fig:consistency-biases}
\end{figure}

\noindent {\bf Voting design and choice context have an impact on human-AI inconsistencies.} Figure~\ref{fig:consistency-biases} illustrates the human-AI consistency in individual and collective choices for single choice and multi-choice voting. For multi-choice voting, the individual and collective consistency of human and AI choices are measured as the average consistency across various sampled population sizes of 25\%, 50\%, and 75\% (shown separately in Figure~\ref{fig:consistency-biases_actual_division}).
The consistency of individual choice remains poor in complex ballot formats with several alternatives.  On average, it is 5.68\% and 28.005\% for the actual and survey voting scenarios of City Idea, yet it is 84.5\% for the binary majoritarian US elections. \texttt{GPT 4-o Mini} shows the highest consistency of individual choice among the seven proprietary and open-source large language models, which is 4.85\% and 7.85\% higher than \texttt{GPT3.5}  and \texttt{Llama3-8B}, respectively. We observe that \texttt{Gemini 1.5 Flash}  and \texttt{Deepseek R1}  have comparable performance with \texttt{GPT3.5}. On the contrary, the consistency of collective choice increases by 46.78\% in overall. Strikingly, the consistency of equal shares and Phragmen's is on average 69.9\%, which is 31.6\% higher than utilitarian greedy. Even when reducing the number of winners in equal shares to that of utilitarian greedy, the consistency remains 22.8\% higher.
The consistency differences of the proportional methods compared to utilitarian greedy, without or with the same number of winners, are statistically significant with (p $<$ 0.03) and (p $<$ 0.04), respectively. Compared to large language models, the machine learning model shows 1.7\% higher consistency in individual choice and 2.9\% higher in collective choice. The consistency values shown here are based on the AI emulations using all personal human traits, as shown in Figure~\ref{fig:factorial}a. Removing project preferences from the context of AI choice generation results in the highest consistency reduction of 18.1\%, whereas political interest leads to the lowest reduction of 3.5\%. 


\begin{figure}[!htb]
    \centering
    \includegraphics[scale = 0.152]{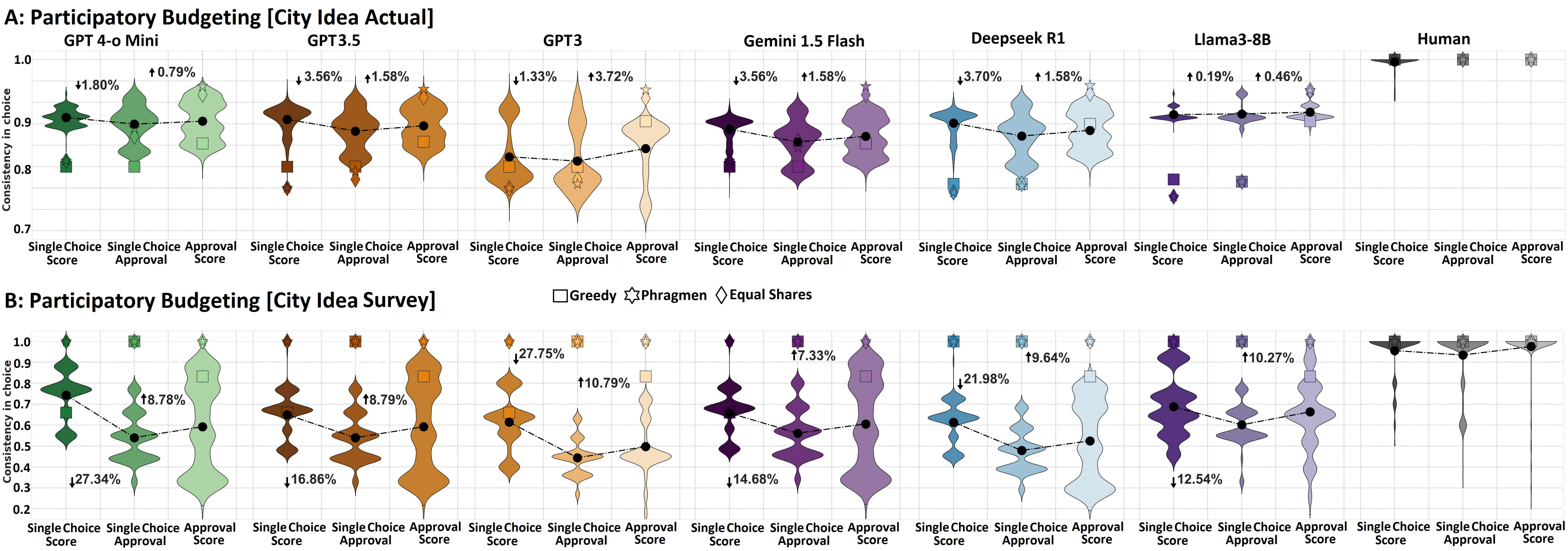}
    \caption{\textbf{Intransitivity of AI across different pairs of ballot formats is higher than the one of humans, which remains negligible. AI intransitivities have a higher influence on the consistency of voting outcomes over a large number of alternatives. \texttt{Llama3-8B} predicts ballots that are not very diverse and selects a limited set of projects, which results in higher transitivity compared to other language models. Equal shares and Phragmén's also show here higher capacity to mitigate the ballot intransitivities. It achieves more than 80\% consistency in preserving voting outcomes between cumulative and approval ballots.} The consistency (y-axis) in individual choice among different pairs of ballot formats (x-axis) is shown for six large language models ({\texttt{GPT 4-o Mini} }, {\texttt{GPT3.5} },  {\texttt{GPT3} }, \texttt{Gemini 1.5 Flash},  { \texttt{Deepseek R1} }  and {\texttt{Llama3-8B}}), humans and the two voting scenarios in the participatory budgeting campaign of City Idea: (A) actual vs. (B) survey. Mean consistency values are calculated across randomly sampled population of 25\%, 50\%, and 75\%.}
    \label{fig:transitivity}
\end{figure}

\noindent \cparagraph{Intransitivity: higher for AI with impact on collective choice among many alternatives} Figure~\ref{fig:transitivity} illustrate the transitivity of preferences in different large language models and humans by measuring the consistency of individual and collective choices across different pairs of ballot formats (see Section~\ref{sec:introduction} and Figure~\ref{fig:factorial}a). While human transitivity averages 97.1\%, AI transitivity is 74.3\% for \texttt{GPT 4-o Mini}, 72.1\% for \texttt{GPT3.5}, 76.2\% for \texttt{Llama3-8B}, and 71.23\% for \texttt{Gemini 1.5 Flash}. In terms of collective choice, equal shares shows 12.2\% higher consistency than utilitarian greedy (p < 0.04). Equal shares achieves more than 80\% consistency among winners based on cumulative, score, and approval ballots. For the actual voting of City Idea, `approval-cumulative' voting demonstrates the highest transitivity, which is 4.4\% higher than `single choice-approval' and 3.2\% higher than `single choice-cumulative'. However, for the survey of City Idea, `single choice-cumulative' shows the highest transitivity, which is 25.2\% higher than `single choice-approval' and 15.1\% higher than `approval-cumulative' (p < 0.03).

\subsection{AI representatives to recover from low voters turnout}\label{sec:AI-representatives}

\begin{figure}[!htb]
    \centering
    \includegraphics[scale = 0.285]{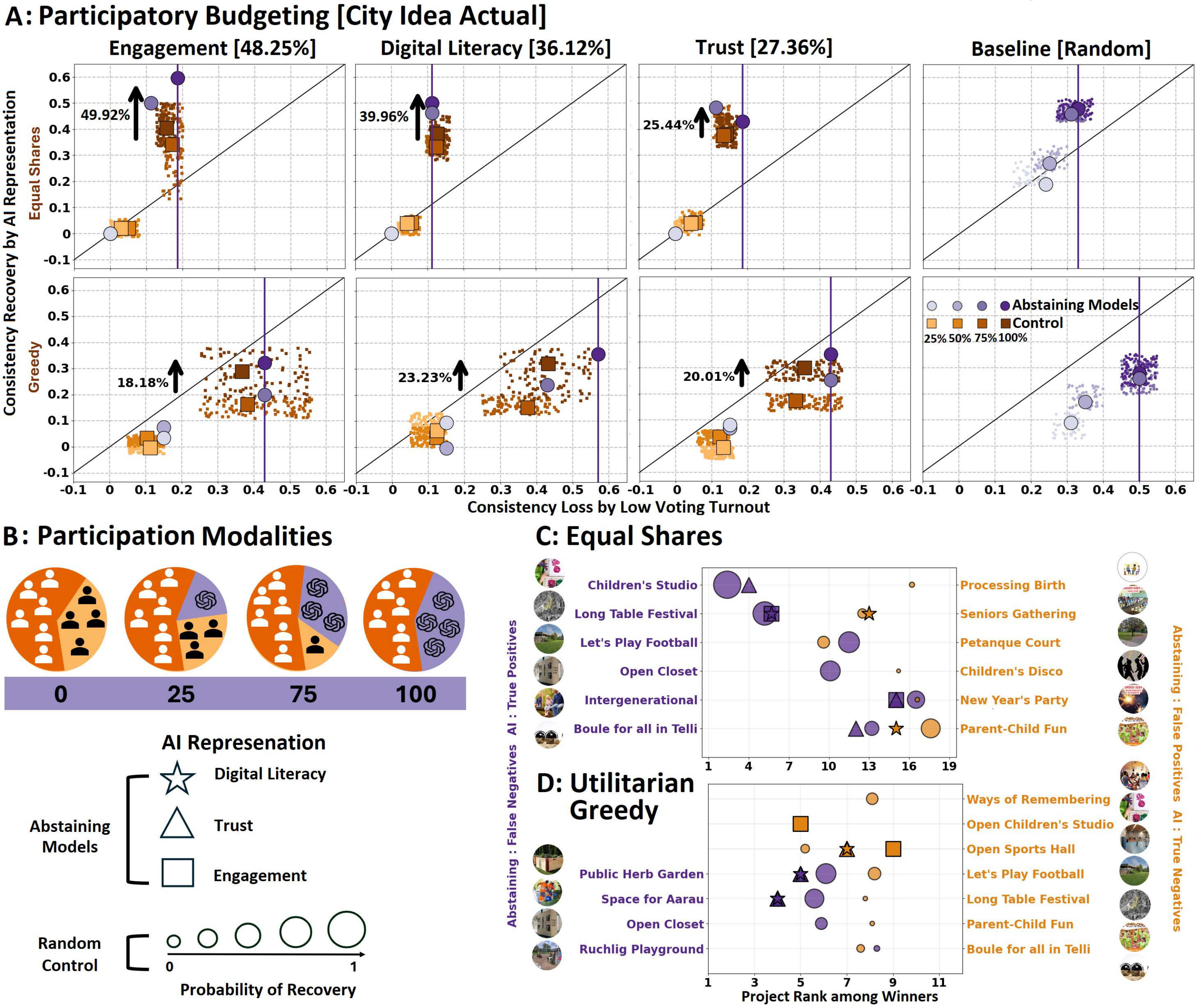}
    \caption{\textcolor{black}{\textbf{Representing more than half of human abstaining voters with AI results in significant consistency recovery, in particular for fair ballot aggregation methods. Strikingly, AI representation of abstained voters is more effective than representing arbitrary voters (random control). Consistency recovery is at two levels: (i) False negative projects removed under abstaining but added back by AI representatives, which are  higher in ranking and number than (ii) false positive projects added under abstaining but removed by AI representatives.}
    The consistency loss in voting outcomes by low voters turnout (x-axis) is emulated by removing different ratios of human voters (25\%, 50\%, 75\% and 100\%) among the whole population (baseline) and those who are likely to abstain: low engagement, trust and digital literacy profile (\% of the abstaining populations in the brackets on top). A consistency recovery (y-axis) is hypothesized by AI representation using \texttt{GPT3.5}. (A) Actual participatory budgeting campaign of City Idea. (B) Studied participation modalities. \textcolor{black}{(C)-(D) Depict which projects are recovered or added by AI representation of abstaining voters (digital literacy, trust, low engagement). When voters abstain, some projects are sacrificed, and purple markers represent the projects added back using AI representation. The orange projects are those that newly emerge as winners with AI representation. The projects and their probability to recover consistency under random control (recovery of 30 groups of random voters, each of size equal to the abstaining voters) are shown for comparison. } }}
    \label{fig:consistency-recovery}
\end{figure}

\cparagraph{Assessing consistency recovery by AI representatives} Figure~\ref{fig:consistency-recovery} illustrates the capability of AI representatives to recover the consistency of voting outcomes lost by low voter turnout. For a certain set of projects that are winners in the final voting outcome when all voters participate, abstaining can therefore lead to an outcome with fewer or more projects. The winning projects removed due to the abstaining population represent a loss of consistency in the voting outcome. Consistency recovery using AI representation for voters who are likely to abstain is electing winning projects that contain or remove the projects that would be erroneously removed or added respectively while abstaining. It is calculated as the difference of consistency of the two scenarios, see Section~\ref{sec:consistency}. The three abstaining models (low engagement, trust, and digital literacy) are assessed along with the baseline that determines random abstaining voters across the whole population. Four participation modalities (Figure~\ref{fig:factorial}a) are studied: (i) Human voters exclusively with 100\% of voters' turnout. (ii) Human voters exclusively with varying turnout levels in the range [25\%,100\%] with a step of 25\%. The maximum number of abstaining voters is either the total voters (baseline in Figure~\ref{fig:consistency-recovery}b) or the number of voters with low digital literacy, engagement and trust as determined by the abstaining models. (iii) Mixed populations of human voters and AI representatives of abstaining voters in the range [25\%,100\%] with a step of 25\%. (iv) AI representatives exclusively. We show the consistency recovery in the actual voting scenario for the City Idea campaign using \texttt{GPT3.5} in Figure~\ref{fig:consistency-recovery} and using the other large language models (\texttt{GPT 4-o Mini}, \texttt{Llama3-8B}, \texttt{Gemini 1.5 Flash}, \texttt{GPT3} and
\texttt{Deepseek R1}) in Section~\ref{sec:recoverysupple}. The results on consistency recovery by AI representatives in the survey voting scenario of City Idea have been shown in Figure~\ref{fig:consistency-recovery-survey-US}.

\cparagraph{Can AI representatives mitigate for abstained voters?}
Strikingly, up to 75\% of AI representation of low-engaged abstaining voters (94 representatives out of 126 abstaining voters in a population of 252 voters, see Section~\ref{sec:abstainingStats}) is sufficient to recover up to 50\% higher lost consistency than the random control population using equal shares. This superior consistency recovery is also observed for abstaining voters with low digital literacy (39.96\%) and trust (25.44\%). The fair aggregation rules of equal shares and Phragmén's achieve, on average 7.53\% higher recovery compared to utilitarian greedy for all the abstaining models. Even when controlling for the same number of winners, fair ballot aggregation methods achieve higher recovery than utilitarian greedy by 6.72\% (p < 0.05). Comparing the different AI models, we earlier observed in Figure~\ref{fig:consistency-recovery} that the collective consistency, that is the one between voting outcomes corresponding to humans and those corresponding to 100\% AI representation (Figure~\ref{fig:consistency-biases}), is comparable for \texttt{GPT 4-o Mini} and \texttt{GPT3.5}, with no statistically significant difference. We notice a similar trend here, where AI representation by \texttt{GPT 4-o Mini} achieves 2.1\% higher recovery than \texttt{GPT3.5}, which is though not statistically significant (p=0.092) (Figures~\ref{fig:consistency-recovery-GPT4},~\ref{fig:consistency-recovery_remain}). However, AI representation by \texttt{GPT 4-o Mini} shows significant differences in consistency recovery compared to \texttt{Llama3-8B} and \texttt{GPT3}, outperforming them by 6.4\% and 8.2\% respectively (Figures~\ref{fig:consistency-recovery_remain},~\ref{fig:consistency-recovery_llama} and~\ref{fig:consistency-recovery_GPT3}). \texttt{GPT3.5} performs better than \texttt{Llama3-8B} and \texttt{GPT3}, achieving recovery gains of 4.61\% and 5.97\%, respectively (Figures~\ref{fig:consistency-recovery_remain}~\ref{fig:consistency-recovery_llama},~\ref{fig:consistency-recovery_GPT3}, and Table~\ref{tab:false_positives_negatives_all_models}).

\noindent \textcolor{black}{
 \cparagraph{AI representation of arbitrary vs. abstaining voters: from removing noise to restoring representation deficit} Figures~\ref{fig:consistency-recovery}c and~\ref{fig:consistency-recovery}d show the origin of inconsistency under utilitarian greedy and equal shares when voters abstain and how AI representatives recover from this. The figures show which projects are involved in consistency recovery and their ranking: (i) erroneously removed projects (false negatives, left) that are correctly added back by AI representatives (true positives) and (ii) erroneously added projects (false positives, right) that are correctly removed by AI representatives (true negatives). Compared to true negative projects, true positive ones are higher in ranking by an average of 7.2 and 2.5 positions for equal shares and utilitarian greedy, respectively. The higher consistency recovery by the abstaining models compared to the random control population originates from an average of 0.71 and 0.47 additional projects involved in consistency  recovery for the two ballot aggregation methods, respectively. Moreover, the origin of consistency recovery by abstaining models is more prominent to true positive projects (mean of 1.66 over 1.0 for true negatives), while it is more prominent to true negative projects in the random control populations (mean of 2.27 over 1.89 for true positives). See Table~\ref{tab:recovery} for a complete outline based on all the AI models. This result demonstrates a distinguishing quality of targeting the AI representation to abstaining voters: representation deficit is restored by adding back winners who would not be there otherwise, while a non-targeted AI representation has a noise-removal effect by removing erroneous winners.  
The district wise consistency recovery for the Aarau has been enumerated in Table~\ref{table:region}.}

\subsection{Biases explaining AI (in)consistencies in choice and preference transitivity}\label{sec:explaining-inconsistencies}

\begin{figure}[!htb]
    \centering
    \includegraphics[scale = 0.22]{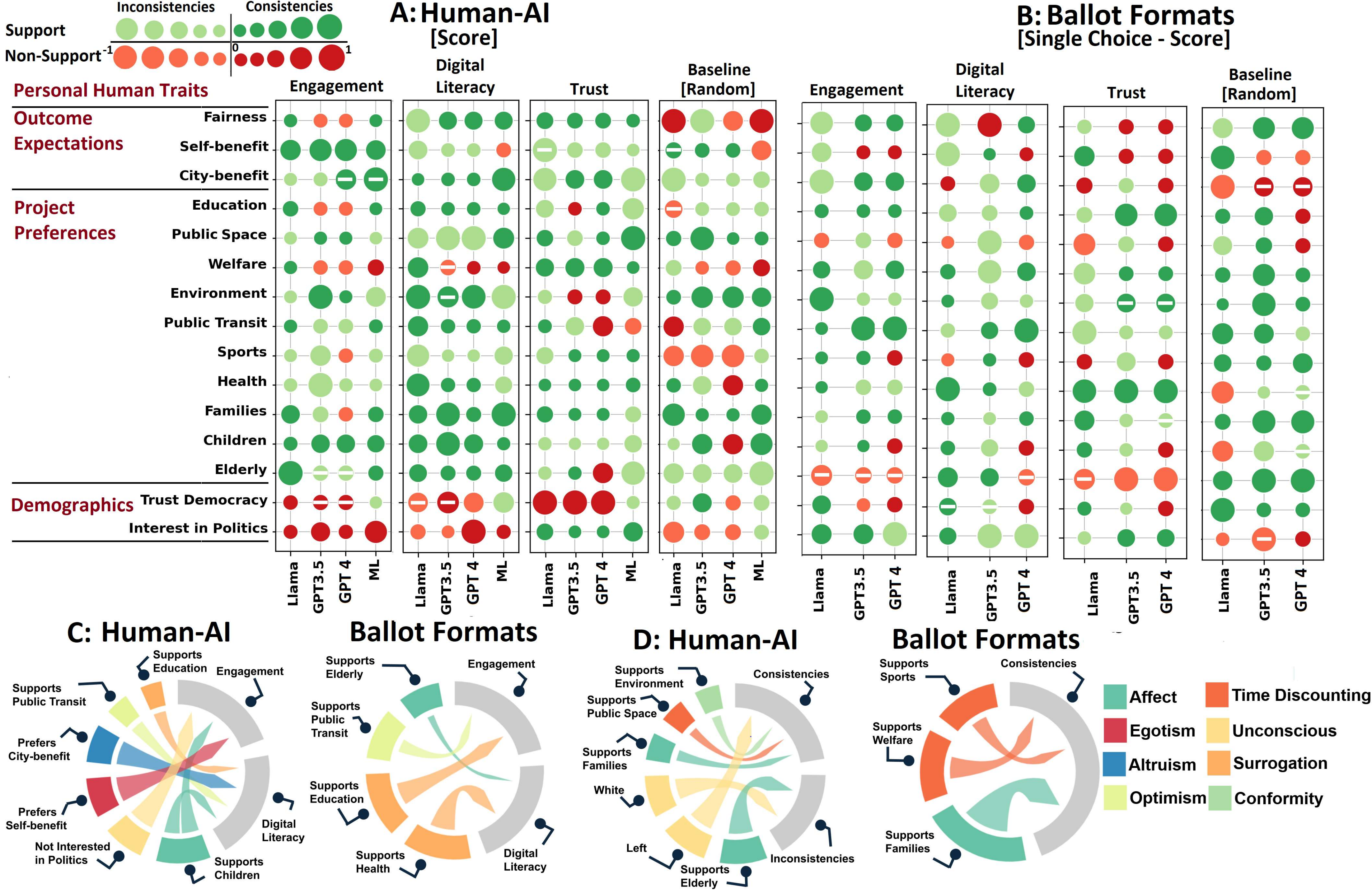}
    \caption{\textcolor{black}{\textbf{Compared to an arbitrary abstaining voter, those with low engagement and digital literacy exhibit characteristics that explain the consistency of human-AI representation and ballot formats, for instance, no interest in politics and support to education/health projects related to unconscious and surrogation biases. Time discounting, affect and conformity biases, such as preference for public space and environmental projects as well as support to families contribute to the consistency of human-AI choice. Time discounting factors such as preference for sport, and welfare projects as well as affect heuristics such as preference for projects that benefit families explain AI consistency among ballot formats.} 
The relative importance of the personal human traits (y axis) for the actual participatory budgeting campaign of City Idea, using \texttt{GPT 4-o Mini} (GPT 4), \texttt{GPT3.5}, \texttt{Llama3-8B} (Llama), and the predictive AI model ({\em ML}) on the x axis, is depicted by the size of the bubbles and is calculated using shapley additive explanations. The consistency of (A) human-AI representation and (B) ballot formats (single choice vs. cumulative) is assessed. For each of these, the personal human traits explain the following: (i) The consistency difference between the three abstaining models and their random control. (ii) The (in)consistency of AI representation and transitivity for the whole population. The `-' sign indicates non-significant values (p>0.05). (C)-(D) The statistically significant biases present in all AI models and datasets are summarized by chord diagrams. }}\label{fig:biases}
\end{figure}

\textcolor{black}{\cparagraph{Unraveling biases that explain AI inconsistencies} Figure~\ref{fig:biases} illustrates the biases that explain the (in)consistency of human-AI choice and the AI transitivity among different ballot formats (single choice vs. cumulative).  
We mainly show the results of the actual participatory budgeting of City Idea, while the results of the other datasets are shown in Figure~\ref{fig:ML_supple}. We distinguish between (i) the inconsistencies originated by the three identified subpopulations of abstained voters and (ii) the inconsistencies by the AI representation of the entire population. Prediction models are constructed using recurrent neural networks (see Sections~\ref{sec:fair} and~\ref{sec:ml}), demonstrating robust performance with F1 scores averaging over 80\% for abstaining groups and 74\% for the entire population across all large language models (Table~\ref{tab:result_all}). The different personal human traits are used as features to predict the consistency between human and AI choices or between AI choices corresponding to different ballots. The relative importance of the personal human traits (independent variables) that explain the AI consistency for individual voters (dependent variable) is calculated using model agnostic shapley additive explanations and local interpretable model-agnostic explanations (see Section~\ref{sec:interpret-AI}, Figures~\ref{fig:Lime_score},~\ref{fig:Lime_approval},~\ref{fig:score_SHAP_DS_Gem_v1})~\cite{framling2021comparison}. The features that are statistically significant and have high importance scores are then analyzed to understand the types of biases based on existing literature evidence (see Section~\ref{sec:bias}). } For the abstaining models, the dependent variable is the difference (plotted in Figure~\ref{fig:consistency-gain}) between the consistency of abstaining voters and the mean of 10 random control subpopulations as shown in Section~\ref{sec:AI-representatives}. This allows us to isolate the biases on the voters who are likely to abstain rather than on arbitrary voters. To provide more robust evidence, we distinguish in Figures~\ref{fig:biases}c and \ref{fig:biases}d those personal human traits that explain AI consistency (i) in all datasets, (ii) for \texttt{GPT 4-o Mini}, \texttt{GPT3.5}, and \texttt{Llama3-8B}, and (iii) those which are statistically significant (p<0.05).

\noindent \cparagraph{Affect and unconscious biases explain the (in)consistency of human-AI choice, while time discounting biases explain transitivity of AI choice over ballot formats} The consistency of human-AI choice is explained by support to families (affect, 7.43\%, p=0.03), public space (time discounting, 8.91\%, p=0.01) and environment (conformity, 8.46\%, p=0.03), while inconsistency is explained by support to elderly (affect, 11.39\%, p=0.04). 
 For the US elections, a political profile of left explains consistency of human-AI choice, while white voters explain inconsistency (33\% higher than consistency, p<0.019, see Figure~\ref{fig:ML_supple}). Affect and time discounting biases also explain the transitivity of AI representation over different ballot formats, in particular the support to families (18.22\%, p=0.01), welfare (16.78\%, p=0.02) and sport projects (17.05\%, p=0.02). Figures~\ref{fig:approval},~\ref{fig:Lime_score},~\ref{fig:Lime_approval},~\ref{fig:score_SHAP_DS_Gem_v1}, Table~\ref{tab:approval_tab} and Section~\ref{sec:explain} illustrate additional insights about how personal human traits explain the AI top choice and the human-AI consistency of the individual choices for six large language models: \texttt{GPT 4-o Mini}, \texttt{GPT3.5}, \texttt{GPT3}, \texttt{Llama3-8B}, \texttt{Gemini 1.5 Flash} and \texttt{Deepseek R1}.

\section{Discussion}

\textcolor{black}{\cparagraph{The {\em inevitability} of generative AI voting and the race to safeguard democracy}} Generative AI voting is likely to emerge as an inevitable technological convergence of AI and electronic voting solutions that are already being adopted in the real world. Our research does not imply or advocate the use of AI as a substitute for human voters who may choose to abstain. AI cannot replicate the human decision-making process in voting, which is shaped by socio-cultural and economic backgrounds, life experiences, and personal choices. However, generative AI and large language models are expected to become more open, pervasive and accessible to citizens~\cite{Bail2024,Pournaras2023}. AI personal assistants are already part of everyday life~\cite{Krishna2022,Heer2019,Asikis2021}, with their generative version expected to follow. On the other hand, the mandate of more direct, secure and active participation in decision-making for public matters is expected to further scale up electronic voting solutions and digital platforms. For instance, participatory budgeting elections are mainly conducted digitally, while Estonia has already institutionalized a digital identity for 99\% of its citizens as well as electronic voting since 2005~\cite{Kattel2019}. As the former president of Estonia emphasized "\emph{with the digital signature and the machine-readable ID card, we created the e-citizen}". In the light of these converging technological advancements, the inter-operation of a generative personal voting assistant with digital voting platforms becomes technologically feasible, along with the citizens' need to have a more direct say in several public matters and consultations. Therefore, the findings of this study become spot-on to understand the implications of such a future, while they are significant to prepare timely safeguards for digital democracy. Another inevitable risk for the integrity of elections is the use of AI representatives for running opinion and election polls at lower cost and larger scale. This is particularly alarming given the influential role of polls to shape voting behavior and how they can be often instrumentalized to influence election results~\cite{dahlgaard2017election,stoetzer2024learning}. Section~\ref{sec:recoverysupple} and Table~\ref{tab:USpre} evaluates the consistency of the voting results using different sampling strategies of voters represented by AI models.

\cparagraph{What we can optimize for: Fair voting design as a democratic safeguard to generative AI voting} We show that large language models currently have limitations in accurately representing individual human preferences in complex voting scenarios, such as participatory budgeting. They are also susceptible to multi-faceted biases. However, we also show that in voting scenarios involving AI representatives, voting design can play a crucial role by preserving the consistency of choices and elections as well as maximizing the recovery of consistency lost by abstaining voters. This is particularly the case for ballot aggregation methods that promote proportional representation such as equal shares. 
Therefore, this motivates a huge opportunity to get democracy "right" in the digital era of AI: move to alternative voting methods that yield fairer voting outcomes for all, while shielding democratic outcomes from AI biases and inconsistencies. \textcolor{black}{How to scale up these democratic blueprints remains, though, an open question. In particular, voting turnouts in participatory budgeting remain very low and far lower than in other elections, such as referenda or national elections. Despite the eminent ethical and legal challenges of engaging AI representatives in democratic processes, our findings show that a more ethically aligned AI representation of abstaining voters recovers consistency of voting outcomes, which would be lost in any case. This consistency loss can be to such a large extent that it is currently posing a long-standing barrier for participatory initiatives to take off. Note that we focus on the AI representation of abstaining voters who intend to participate but their low engagement~\cite{hylton2023voter,lane2018planning}, digital literacy~\cite{aichholzer2020experience,vassil2011bottleneck,rattanasevee2024direct,Gudino2024,lorenz2023systematic}, or trust~\cite{wang2016political,belanger2017political,devine2024does} are barriers for them. This is to distinguish voters whose abstention is a conscious, deliberate act, and their AI representation would not be relevant or even desired in this context. Last but not least, our findings also demonstrate that AI representation alone does not suffice - a fair voting design is imperative to materialize significant recoveries from low voting turnouts.}

\noindent \textcolor{black}{\cparagraph{Why fair collective choice is resilient to AI biases and inconsistencies} We provide an explanation of this significant finding. There is evidence that the equal shares method has an inherent stability in the resulting voting outcomes~\cite{Fairstein2023}. Low- and middle-cost projects require very minimal support to get elected, and as a result, these winning projects are likely to be retained in the winning set, even with different choices or groups of voters. Such projects are expected to be a source of consistency.} Indeed, this effect is also observed in the real-world voting scenario of City Idea, as with 80\% abstaining voters, 84\% of the winners are retained with equal shares, see Figure~\ref{fig:mixed populations-abs} and Table~\ref{table:change_gr_ES} that shows the origin of this stability in terms of new projects added and removed in the winning set. Nevertheless, equal shares is still affected by low voter turnout, especially, for participation rates $<50\%$~\cite{Fairstein2023}. As voter turnout in participatory budgeting is typically very low, these inconsistencies are both relevant and prevalent. Note that any comparison of stability between equal shares and utilitarian greedy should be made with caution, as the number of winning projects under equal shares is much larger than utilitarian greedy. When we control for the number of winning projects between the two methods, equal shares remains more robust than utilitarian greedy, but to a much lower extent (see baseline [random] in Figures~\ref{fig:consistency-recovery}a and~\ref{fig:consistency-recovery_remain}.)

\noindent \textcolor{black}{\cparagraph{Real-world testing of equal shares: overcoming a {\em validity barrier} and addressing data limitations}}
As City Idea promoted equal shares already in the project ideation phase and made use of equal shares for the aggregation of the ballots, this study becomes the first of its kind:   significant findings are illustrated that come with compelling realism and merit for their validity. This comes in stark contrast to other earlier studies~\cite{Fairstein2023,maharjan2024fair} that hypothesize the application of equal shares over proposed projects and ballots aggregated with the standard method of utilitarian greedy. Access to voters' profiles and preferences to emulate AI representation for voters is particularly limited. The City Idea participatory budgeting campaign overcame these limitations by collecting relevant data that captures such preferences to a meaningful extent. However, we also acknowledge that a broader collection of participatory budgeting elections from Pabulib~\cite{faliszewski2023participatory} could not be used in our study to analyze potential biases due to the unavailability of preference data.

\cparagraph{ Trustworthy generative AI voting: a call for research and policy action} What information large language models use to reason about voting decisions is influential for different types of biases to manifest. \textcolor{black}{This is particularly the case for affect, unconscious and time discounting biases involved in AI representation of human choices and the transitivity of AI choices over different ballot formats. Abstaining voters with low engagement, digital literacy and trust also possess related personal human traits that explain the consistency of their AI representation. Voters who come with a more active participation profile, without typical features of abstaining voters, appear irreplaceable, as the reasoning of large language models cannot accurately estimate their choices (Figure~\ref{fig:consistency-recovery_opp}). This motivates a tailored, purposeful and finite use of AI representation with the aim to make itself obsolete by recovering participation of abstaining voters, while mitigating for the consistency loss as long as voters abstain.} Training data in generative AI voting are expected to play a key role for representative voting outcomes of the voter population. Ethical and democratic guidelines are urgently needed, particularly for the use of (generative) AI in voting processes. For instance, who shall determine the input training data of AI representatives? Should the training data involve only self-determined personal information of voters, or shall these be augmented with more universal knowledge and experts' opinions? How to protect the privacy and autonomy of voters when training such AI representatives~\cite{Asikis2021}? Will citizens retain power to control AI representatives that reflect their values and beliefs while remaining accountable? These are some key questions as a basis of a call for action on research and policy in an emerging era of generative AI voting.


\section{Methods}\label{sec:method}

We show here how AI representatives are emulated and the real-world data based on which the voting scenarios are constructed. We also illustrate the evaluation approach and the studied human cognitive biases. Finally, the approach to explain the inconsistencies and biases of generative AI is outlined. 

Note that p values reported for statistical significance in Section~\ref{sec:results} (Results) are combined p values, which are based on summing log-transformed individual p values from all different runs (corresponding to different hyper-parameters) using the Fisher Method~\cite{yoon2021powerful}.

\subsection{Emulating AI representatives}\label{sec:emul}

The process of AI emulation using personal human traits, ballot formats, aggregation methods, and AI models is demonstrated for each of the real-world voting scenarios. Table~\ref{table:persona1} outlines the characteristics of the emulated voting scenarios.

\noindent \cparagraph{US elections} The 2012, 2016, and 2020 survey waves of the American National Election Study (ANES)~\cite{herrick2020gender} are used. The dataset for three years contains 20,650 voters together with the respective voter socio-demographic characteristics for each of these three years: (i) racial/ethnic self-identification [white, black, Asian, Hispanic, or others], (ii) gender [male, female, others], (iii) age, (iv) ideology [extremely liberal, liberal, slightly liberal, moderate, slightly conservative, conservative, or extremely conservative], (v) political belief [democrat, republican, or independent], (vi) political interest [very interested, somewhat interested, not very interested, or not at all interested], (vii) church attendance [yes, no], (viii) whether the respondent reported discussing politics with family and friends [yes, no], (ix) feelings of patriotism associated with the American flag [extremely good, moderately good, a little good, neither good nor bad, a little bad, moderately bad, or extremely bad], and (x) state of residence. A total of 18 elections are emulated using two combinations of human traits and three AI models, including \texttt{Llama3-8B}, \texttt{GPT3.5}, and a predictive ML model based on single choice ballots, with winners determined by majority aggregation for 2012, 2016 and 2020. Another 3 elections were emulated for \texttt{GPT 4-o Mini} based on single choice ballots, majority aggregation, and for one combination of human trait (see Table~\ref{table:persona1}). 
We systematically removed responses containing missing data, resulting in a refined subset of 17,010 voters with complete responses. These voters were emulated using three large language models - \texttt{GPT 4-o Mini}, \texttt{GPT3.5}, and \texttt{Llama3-8B} for generating a total of 51,030 AI representatives. Additionally, we incorporated 3,640 voters from the original dataset who had provided partial responses specifically related to personal human traits. These incomplete cases were similarly emulated using \texttt{GPT 4-o Mini}, \texttt{GPT3.5} and \texttt{Llama3-8B}, yielding 7,280 AI representatives. While these partially complete responses were utilized for vote aggregation, they were excluded from consistency prediction due to data limitations. The examples of the prompts used to generate the AI choices can be found in Table~\ref{table:Anes}.

\cparagraph{City Idea: Participatory budgeting campaign in Aarau, Switzerland}
The data from a recent innovative participatory budgeting campaign are used~\cite{CityIdeaReport2025,maharjan2024fair}, which was conducted with ethical approval from University of Fribourg (\#2021-680). It run in 2023 and is rigorously designed to assess the application of equal shares for the first time in real world, in combination with cumulative voting, using the open-source Stanford Participatory Budgeting platform~\cite{Welling2023fair}.
\textcolor{black}{ The  campaign was structured into six phases over a period of nine months. As part of this process, both pre-voting and post-voting surveys were conducted to capture the personal traits of the participant as well as their perspectives before and after the voting phase. The pre-voting survey was disseminated through physical invitation letters sent by the city council to all citizens, yielding 3,592 respondents. Of these, 808 individuals voluntarily participated in the post-voting survey. In total, 1,703 citizens participated in the voting process, of whom 252 also completed both the pre-voting and post-voting surveys. The participation achieved a gender balance, proportional representation of citizens and non-citizens, and equitable representation across the 18 districts of Aarau.
As such, the field study includes a survey conducted before voting linking the choices of survey respondents and voters. We use the following personal human traits from the survey (more information in Tables~\ref{table:prologue1} and~\ref{table:prologue2}): (i) 9 key socio-demographic characteristics (e.g., age, citizenship, education) and 2 political interests (political beliefs and trust in democracy),
(ii) preferences for 9 different types of projects and 6 beneficiaries and (iii) 4 types of preferences / expectations for qualities of the voting outcome.}

\begin{table}[!htb]
		\caption{The studied dimensions across three real-world voting scenarios. They provide the necessary diversity to generalize the findings of this study as they include different  number of voters, different ballot formats and aggregation methods, low and high numbers of alternatives, different personal human traits for studying a broad spectrum of biases including both generative and predictive AI methods.}\label{table:persona1}
		\centering
		\resizebox{0.97\textwidth}{!}{%
			\begin{tabular}{llll}\toprule
				\multicolumn{1}{l}{\centering  Studied factors} &  US elections   &  City Idea  & City Idea   \\
 &  2012, 2016, 2020  &  [Survey] &  [Actual]\\\midrule
    {{\bf Ballot input}} & & & \\\midrule

    {\bf Personal human traits}  & & & \\
  Socio-demographics, Outcome expectations, Project preferences, Political interests & \xmark & \cmark & \cmark \\
 Socio-demographics, Outcome expectations, Project preferences & \xmark & 
\cmark & \cmark \\
 Socio-demographics,  Project preferences, Political interests & \xmark & \cmark & \cmark \\
 Socio-demographics, Outcome expectations, Political interests & \xmark & 
 
\cmark & \xmark \\
 Socio-demographics, Political interests & \cmark & 
 
\cmark & \xmark \\
 Socio-demographics, Political interests (only 1 feature) & \cmark & 
 
\xmark & \xmark \\
{\bf Ballot formats}  & & & \\
  Single choice & \cmark & \cmark & \cmark \\
    Approval & \xmark & \cmark & \cmark \\
   Score & \xmark & \cmark & \xmark \\ 
   Cumulative & \xmark & \xmark & \cmark \\
      {\bf Alternatives for voting} & 2  &  5  & 33\\\midrule
{{\bf Ballot generation}} & & & \\\midrule
{\bf Generative AI} & & & \\
 \texttt{GPT 4-o Mini}{\Large $^*$} & \cmark & \cmark & \cmark \\
     \texttt{GPT3.5}  & \cmark & \cmark & \cmark \\
     \texttt{\texttt{GPT3} } & \xmark & \cmark & \cmark \\
               \texttt{Llama3-8B} & \cmark & \cmark & \cmark \\
                \texttt{Deepseek R1} {\Large $^*$} & \xmark & \cmark & \cmark \\
                 \texttt{Gemini 1.5 Flash} {\Large $^*$} & \xmark & \cmark & \cmark \\
               {\bf Predictive AI (ML)} & & & \\
         Neural Networks  & \cmark & \cmark & \cmark  \\\midrule
        
   {{\bf Ballot aggregation}} & & & \\\midrule
   Majority & \cmark & \cmark & \cmark \\
    Utilitarian greedy & \xmark & \cmark & \cmark \\
  Equal shares & \xmark & \cmark & \cmark \\\midrule

   {\bf Voters} & $\sim$17,010 (across 3 years)  &  3,314  & 505\\ \midrule
   {\bf Emulated elections} & 21  & 207 & 135 \\   
   
   \hline
   \multicolumn{4}{c}{{$*$} Only for 1 combination of personal human traits - Socio-demographics, Outcome Expectations, Project Preferences, Political Interests} \\\hline
			\end{tabular}
		}
	\end{table}

\textcolor{black}{Two participatory budgeting voting scenarios are studied in the context of the real-world campaign of City Idea.  The examples of the prompts used to generate the AI choices in both survey and actual voting can be found in Table~\ref{table:prologuePersona}, with more details in Section~\ref{sec:study}.}

\begin{itemize}
\item[(i)]{\em Survey Voting}: Five hypothetically costed projects belonging in different categories are put for choice as part of the initial survey. Table~\ref{table:prologueS} illustrates the project alternatives and their cost. The choice of 3,314 voters over the same alternatives is tested with three different ballot formats in a sequence, starting with the simplest one of single choice to the most complex ones of approvals and score voting.  The set of 3,314 voters also provided their personal trait information in the survey. This allows us to emulate 180 elections = 3 ballot formats x 4 AI models x 5 combinations of personal traits  x 3 ballot aggregation methods. An additional 27 elections were emulated using all human traits, the three ballot format - (single choice, approval and score) and majority, utilitarian greedy and equal shares  ballot aggregation for the \texttt{GPT 4-o Mini}, \texttt{Deepseek R1}  and \texttt{Gemini 1.5 Flash}. Hence a total of 207 elections have been emulated.  Based on the various combinations of personal traits, we then
 emulated a total of 19,884 corresponding AI representatives. This included 3,314 representatives for each of the six large language models: \texttt{GPT 4-o Mini}, \texttt{GPT3.5}, \texttt{GPT3}, \texttt{Deepseek R1}, \texttt{Gemini 1.5 Flash} and \texttt{Llama3-8B}.  The AI representatives have then been used to emulate elections based on the combinations of ballot format and ballot aggregation methods.

\item[(ii)]  {\em Actual Voting}: Using the Stanford Participatory Budgeting platform~\cite{Welling2023fair}, 1,703 voters cast their vote using cumulative ballots by distributing 10 points to at least 3 projects of their preference, out of 33 projects in total (see Table~\ref{table:prologueA} for project descriptions). A subset of 505 of these voters, which participated in the initial survey and provided their personal human traits, are used to construct the AI representatives. The ballot formats of single choice and approvals are derived from the cumulative ballots by taking the project with the most points and the projects that received any point respectively. This allows us to emulate 108 elections  = 3 ballot formats x 4  AI models x 3 combinations of personal traits  x 3 ballot aggregation methods. An additional 27 elections were emulated using all human traits, the three ballot format - (single choice, approval and score) and majority, utilitarian greedy and equal shares  ballot aggregation for the \texttt{GPT 4-o Mini}, \texttt{Deepseek R1}  and \texttt{Gemini 1.5 Flash} . Hence a total of 135 elections have been emulated. Of the 1,703 voters, 505 also completed the voting surveys, providing personal trait information for AI emulation. Using various combinations of these traits, we generated 3,030 AI representatives, comprising 505 representatives for each of six large language models: \texttt{GPT 4-o Mini}, \texttt{GPT3.5}, \texttt{GPT3}, \texttt{Deepseek R1}, \texttt{Gemini 1.5 Flash} and \texttt{Llama3-8B}.

\end{itemize}

\cparagraph{Data collection infrastructure} 
Generative AI choices were collected through API prompts to large language models over two periods: from June 16, 2023, to November 8, 2023, and from April 1, 2025, to August 31, 2025. We prompted the large language models using the zero-shot learning feature~\cite{brown2020language}, which does not require any specific fine-tuning. We use chain of thought prompting~\cite{holtzman2019curious} along with context-based prompting~\cite{wei2022chain} to provide a comprehensive and systematic flow of information for better interpretability. A detailed explanation for the prompt designing is provided in Section~\ref{sec:prompt}.

\subsection{Evaluation of choices by AI representatives} \label{sec:consistency}

The emulated elections with AI representatives are compared to the real-world elections of human voters at two levels: (i) \emph{individual choice}, i.e. the ballots, and (ii) \emph{collective choice}, i.e. the resulting voting outcomes. Consistency is the key assessment measure, derived from the \emph{accuracy} of individual and collective AI choices compared to human decisions and the \emph{transitivity} across different ballot formats (see Figure~\ref{fig:factorial}a).

\cparagraph{Consistency of individual choice} 
Single choice ballots for both AI and human voters are represented as binary sequences, where a value of \emph{1} indicates approval of a specific project, and  \emph{0} denotes disapproval of all remaining alternatives. In approval voting, each alternative is assigned either \emph{1} (approved)  or \emph{0} (not approved). In contrast, in score voting and cumulative voting each alternative receives a score or an number of distributed points (integer numbers) reflecting voter preference. To compare AI-generated and human choices, we employ a single method, the Condorcet pairwise comparison method~\cite{kulakowski2022similarity,navarrete2024understanding}, which is a generic approach to characterize the overall similarity of two ballots (or voting outcomes). A preference matrix is constructed, where rows and columns correspond to alternatives, and each matrix element records the outcome of a pairwise comparison.  If project $P_{i}$ is ranked higher than project $P_{j}$, or if $P_{i}$ is approved while $P_{j}$ is not, the corresponding matrix cell $P_{i} > P_{j}$ is assigned a value of 1; otherwise, it is set to 0. Ties are excluded from the analysis.

\begin{itemize}
\item[(i)] {\em Human-AI consistency (accuracy) of  individual choices}: 
The human ballots serve as the reference point for evaluating the ones generated by the AI representatives, see Figure~\ref{fig:factorial}a. The elements of `1' in the matrix of AI representatives that match the elements of `1' in the matrix of human choices determine the consistency~\cite{navarrete2024understanding}.

\item[(ii)] {\em Consistency (transitivity) of AI and human individual choice across ballot formats}: 
\textcolor{black}{
Ballot formats are standardized as follows: For cumulative/score vs. single choice ballots, the highest-scoring projects are set to `1' and the others to `0'. For cumulative/score vs. approval ballots, scored projects are set to `1',  while projects without score are set to `0'. The elements of `1' and `0' in the two matrices of the ballot formats that match determine the consistency.} 
\end{itemize}

\noindent \textcolor{black}{We also compare the choices based on preference reordering using the Kemeny distance~\cite{anshelevich2018approximating} as illustrated in Section~\ref{sec:supllemethod}.}

\cparagraph{Consistency of collective choice} 
\textcolor{black}{This follows the same approach of Condorcet pairwise comparisons for individual choices. However, before calculations of consistency are made, voting outcomes are turned into binary sequences to distinguish winners (`1') from losers (`0') as determined by a ballot aggregation method.}

\cparagraph{Consistency recovery in collective choice with AI representatives} It is determined here for voting scenarios with varying voters turnout, in which abstained voters result in collective consistency loss, which can be recovered if a portion of these abstained voters are represented by AI. \textcolor{black}{This recovery takes place at two levels: (i) False negative projects that are erroneously removed under abstaining but added back by AI representatives. (ii) False positive projects that are erroneously added under abstaining but correctly removed by AI representatives.} Consistency recovery is measured as follows:

{\footnotesize
$$\frac{consistency\ [\text{all\ human\ voters - abstained\ voters + AI\ representatives}] - consistency\ [\text{all\ human\ voters - abstained\ voters}]}{1 - consistency\ [\text{all\ human\ voters - abstained\ voters}]},\ $$
}

\noindent where the voters turnout {\footnotesize $\frac{\text{human\ voters - abstained\ voters}}{\text{human\ voters + abstained\ voters}}$ } varies in the range {\footnotesize[20\%,75\%]} with a step of {\footnotesize 25\%}, and AI representation $\frac{\text{AI\ representatives}}{\text{abstained\ voters}}$ varies in the range {\footnotesize [25\%,100\%]} with a step of {\footnotesize 25\%}.

\subsection{Explainability of generative AI voting}\label{sec:interpret-AI}

\textcolor{black}{The accuracy of the individual AI choices (see Section~\ref{sec:explain}) with human choices as well as the transitivity of AI choices over different ballot formats are modeled as the dependent variable in a predictive machine learning framework. We study causal relationships explaining how personal human traits (independent variables) influence consistency (both accuracy and transitivity). We model the problem of explaining inconsistencies as a classification problem, where 10 uniform consistency levels are defined as the ranges $[0.0,0.1],(0.1,0.2],...,(0.9,1.0]$. Further details about how we account for imbalances of features, their co-linearity and hyperparameter optimization of the model are illustrated in Section~\ref{sec:ml} and Table~\ref{tab:set}. }

\noindent \textcolor{black}{\cparagraph{Explainability of choices} We introduce a two-dimensional feature importance analysis framework to determine the impact of the personal human traits on the consistency of individual choices. 
For a given performance of the prediction model, we employ explainable AI methods to analyze the contribution of each individual human trait (feature) to the outcome. The approach to enhance the performance (accuracy, precision, recall) of the prediction model is illustrated in Section~\ref{sec:ml}. We then use the model agnostic
 Shapley Additive Explanations (SHAP) and Local Interpretable Model Agnostic Explanations (LIME)~\cite{framling2021comparison} to extract the individual contributions of each trait. Results are shown in Figure~\ref{fig:ML_supple},~\ref{fig:approval},~\ref{fig:Lime_score},~\ref{fig:Lime_approval} and Table~\ref{table:error}. A feature ablation study~\cite{hameed2022based} is used to calculate the error (loss) in the overall prediction accuracy of the model when a feature is removed (results in Table~\ref{table:error}).}

\newpage

\section*{Declarations}

\subsection*{Availability of data and materials}
The datasets generated and/or analyzed during the current study are available in Figshare:\\ \url{https://figshare.com/collections/Generative_AI_Voting_-_ANES/7261288}\\ and Github Repository:\\ \url{https://github.com/TDI-Lab/Generative-AI-Voting}

\subsection*{Competing interests}
The authors declare that they have no competing interests.

\subsection*{Funding}
This work is funded by a UKRI Future Leaders Fellowship (MR\-/W009560\-/1): \emph{Digitally Assisted Collective Governance of Smart City Commons--ARTIO}'. The participatory budgeting data were earlier collected in the context of the City Idea campaign  with support from Swiss National Science Foundation NRP77 ‘Digital Transformation’ project (\#407740\_187249): \emph{Digital Democracy: Innovations in Decision-making Processes}.

\subsection*{Authors' contributions}
S.M. wrote the manuscript, collected the data, designed and developed
the AI models, and analyzed the data. E.E. edited the manuscript and analyzed the data. E.P. wrote the manuscript, conceived the study, designed the AI models and analyzed the data. 

\subsection*{Acknowledgements}
The authors would like to thank Regula Hänggli and Dirk Helbing for constructive discussions.

\bibliographystyle{unsrt}
\bibliography{sample}

\makeatletter\@input{yy.tex}\makeatother
\end{document}


\maketitle
    	
    		\footnotetext[1]{Corresponding author: Srijoni Majumdar, School of Computer Science, University of Leeds, Leeds, UK, E-mail: s.majumdar@leeds.ac.uk}
    \tableofcontents

\section{Field study for multi-winner voting }\label{sec:study}

This section outlines the details of the pre-voting and post-voting surveys from the 2023 participatory budgeting campaign of City Idea in Aarau.
We also elaborate on the prompt design that has been used to emulate an AI representation of voters using the data collected from the surveys.

\subsection{Pre-voting and Post-voting surveys}
  The voting scenarios including the projects (alternatives) put up for voting and their characteristics, are presented in Tables~\ref{table:prologueS} and~\ref{table:prologueA}. The personal human traits collected from the pre-voting and post-voting surveys  are provided in Tables~\ref{table:prologue1}–\ref{table:prologue5}.

\subsection{Prompt design for AI representation}\label{sec:prompt}

We highlight the prompt design techniques, along with the approaches employed to mitigate biases introduced by the prompt specifications. Examples of prompts used to generate AI voting personas and their choices are shown in Table~\ref{table:prologuePersona} (survey and actual voting of the City Idea participatory campaign) and Table~\ref{table:Anes} (American National Election Studies).

\cparagraph{Prompt Design} We have designed the prompts using context based prompting~\cite{wei2022chain} with the details of the voting scenarios as the voting context. The voting context primarily includes project descriptions, detailing the type of project, its location, and its impact on citizens, in addition to the ballot formats. The project descriptions are clear and unambiguous. We have further incorporated chain of thought prompting~\cite{chen2023cp}, where individual voter information is provided so that the language model can apply {\em common sense} reasoning considering the global voting context and the individual information. In addition, these models have leveraged high dimensional word embeddings~\cite{naseem2021comprehensive} to effectively analyze semantic similarities between terms such as ''trash cans'' and bins.'' We run the models with temperature settings from 0.4 to 0, performing 20 runs for each setting. We calculate the consistency at each temperature setting and take the mean across all runs~\cite{wei2022chain,chen2023cp}. As we are dealing with a significantly large decision space, particularly the 33 projects in the actual voting, running with very high temperature settings can lead to randomness in the generation of choices~\cite{chen2023cp}. Hence we limit the range of the temperature setting from 0.4 to 0~\cite{wei2022chain,chen2023cp}.

\cparagraph{Prompt induced bias} We employ the following techniques~\cite{marvin2023prompt} to detect and mitigate knowledge, position, and format biases, which are commonly observed in large language model generation and reasoning~\cite{chen2024large,yu2024mitigate}. 

\begin{itemize}
  \setlength\itemsep{0pt} 
  \setlength\parskip{0pt}
\item  {\em Knowledge biases~\cite{chen2024large}}: To analyze and mitigate this bias, we design multiple runs in which we vary (a) the individual voter information, using different combinations of personal traits related to project preferences, voting outcome expectations, socio-demographics, and political interests, and (b) the voting context by providing projects with and without detailed descriptions. We observe that, on average, large language models generate ballots with 3 more projects in the actual voting scenario when all personal traits for individual voter information and project descriptions in the voting context are considered. This indicates that greater knowledge support helps the models generate less sparse ballots, facilitating more legitimate decision making.
 \item {\em Format biases~\cite{chen2024large}}: We experimented by providing the projects and descriptions in both tabular and list formats in the prompt, but the ballots generated did not differ in most cases. However, we observed that in 2\% of ballots generated by \texttt{GPT3.5}  and 4.13\% of ballots generated by \texttt{GPT-4 Mini}, the tabular format produced one less project on average for the actual City Idea voting scenario. Even though this change occurred in a very small subset of the generated ballots, we still proceeded with the list format to mitigate such scenarios.
 \item {\em Position biases~\cite{yu2024mitigate,wang2023primacy}}: 
We tested different project orderings (ascending and descending) based on project ID and cost, as well as the original order used to present the projects for voting. The original order was not sorted by project ID or cost and was mostly based on the sequence in which the projects were proposed and registered. In most cases, the project selections in the generated ballots remained unaffected. However, for \texttt{Llama3-8B}, presenting projects in the original order resulted in ballot generation with 2 more projects on average compared to other order configurations. Therefore, we adopted the original order to include the projects in the prompt context.

\end{itemize}

\begin{table}[!htb]
		\caption{{\bf Participatory budgeting campaign - City Idea [Survey] in Aarau}. A total of {5 projects} were proposed for the survey voting which  were related to urban greenery, public space, public transit and health. The total budget was set to  50,000 CHF. }\label{table:prologueS}
		\centering
		\resizebox{0.59\textwidth}{!}{%
			\begin{tabular}{lll}\toprule
				\textbf{ID} & \textbf{Project Descriptions} & \textbf{Cost (in CHF)}  \\ \midrule

    P1 & Bins placed in local woodland to reduce litter &  5000 \\
    P2 & Recreational activities for elderly & 10,000\\
    P3 & Refurbishment of local park & 30,000\\
    P4 & Mental health counseling at local school & 15,000\\
    P5 & Bike lane improvements & 40,000\\ \midrule
			\end{tabular}
		}
	\end{table}

\begin{table}[!htb]
		\caption{{\bf Participatory budgeting campaign - City Idea [Actual] in Aarau}.  Citizens proposed more than 161 project ideas out of which { 33  projects} are selected  to put for voting~\cite{CityIdeaReport2025}. The proposed projects were  related to education, culture, environment, welfare, urban greenery, public space, public transit, and health. The total budget was set to  50,000 CHF.}\label{table:prologueA}
		\centering
		\resizebox{0.58\textwidth}{!}{%
			\begin{tabular}{lll}\toprule
				\textbf{ID} & \textbf{Project Descriptions} & \textbf{Cost (in CHF)}  \\ \midrule

      P1	&	Upgrade Ruchlig soccer field	&	15,000	\\
    P2	&	Boule for all in Telli	&	2800	\\
    P3	&	Intergenerational project	&	1600	\\
    P4	&	Wild bees' paradise	&	20,000	\\
    P5	&	Parent-Child Fun and Action Day	&	3100	\\
    P6	&	Gruezi 2024 - New Year's Party	&	4000	\\
    P7	&	Children's Disco	&	4330	\\
    P8	&	Long Table Festival	&	3400	\\
    P9	&	Let's Play Football	&	2300	\\
    P10	&	LGBTQIA+ monthly party	&	20,000	\\
    P11	&	Open sports hall	&	2300	\\
    P12	&	Open closet	&	7000	\\
    P13	&	Open children's studio	&	10,000	\\
    P14	&	Petanque court	&	8000	\\
    P15	&	Pfasyl Aargau	&	3600	\\
    P16	&	Sponsoring a space for Aarau	&	1000	\\
    P17	&	Seniors gathering 70+	&	3500	\\
    P18	&	Processing birth	&	5000	\\
    P19	&	Ways of remembering	&	500	\\
    P20	&	Bread tour	&	1500	\\
    P21	&	Public bicycle pumps	&	4000	\\
    P22	&	CufA - Cultural Festival Aarau	&	15,000	\\
    P23	&	One Place for all	&	17,000	\\
    P24	&	Public herb garden	&	800	\\
    P25	&	Aarau Future Acre	&	3600	\\
    P26	&	Summer fun in the Sonnmatt summer garden	&	1500	\\
    P27	&	New edition of the Telli Map	&	4000	\\
    P28	&	Climate days for Aarau	&	24,000	\\
    P29	&	A Garden for All	&	2500	\\
    P30	&	Summery cinema nights in the Badi	&	10,000	\\
    P31	&	Ruchlig water playground	&	25,000	\\
    P32	&	Usable space with a hedge	&	1000	\\
    P33	&	Playground extension Oehlerpark	&	20,000	\\ \midrule
			\end{tabular}
		}
	\end{table}

\clearpage
    {\scriptsize
    \captionsetup[longtable]{
  labelfont={normalfont,normalsize},
  textfont={normalfont,normalsize}
}
    \setlength{\LTcapwidth}{\linewidth}
    \begin{longtable}{p{0.1\textwidth}p{0.35\textwidth}p{0.1\textwidth}p{0.3\textwidth}}
        \caption{\normalsize Pre-voting (Pr) survey : {Socio-demographics, political interests} and {outcome expectations} }\label{table:prologue1} \\
        \toprule
        \textbf{ID} & \textbf{Question} & \textbf{Type}& \textbf{Options} \\
        \midrule
        \endfirsthead 
        \multicolumn{4}{c}%
        {{\bfseries \tablename\ \thetable{} -- continued from previous page}} \\
        \toprule
        \textbf{ID} & \textbf{Question} & \textbf{Type}& \textbf{Options} \\
        \midrule
        \endhead 
    
        \bottomrule
        \multicolumn{4}{r}{{Continued on next page}} \\
        \endfoot
    
        \bottomrule
        \endlastfoot
    \midrule
       \multicolumn{4}{c}{{\bf Socio-demographic characteristics}}\\ \midrule  \midrule

        SPr.1 & What is your gender? & Single Choice & 3 [man, woman, various/ other] \\\midrule
    
        SPr.2 & What is your age? & Number & String \\\midrule
        SPr.3 & What is your location? & Text & String \\\midrule
        SPr.4 & Are you entitled to vote in Switzerland? & Single Choice & 2 [yes, no] \\\midrule
    
    SPr.5 & What is the highest education you have completed so far? & Single Choice & 5 [school level, bachelors, masters, doctorate and above] \\\midrule

        SPr.6 & Were you born in Switzerland? &  Single Choice & 4 [no, yes, don't know, no answer] \\ \midrule
        
        SPr.7 & Did your parents migrate to Switzerland? & Single Choice & 5 [yes both, only one, no both parents immigrated, don't know, no answer] \\\midrule
       
        SPr.8 & Do you have children? &  Single Choice & 3 [no, yes, no answer] \\\midrule
           
        SPr.9 & Do you have trust in political parties &  Single Choice & 3 [no, yes, no answer] \\\midrule \midrule
    
           \multicolumn{4}{c}{{\bf Political interests}}\\ \midrule \midrule
    
               IPr.1 &  Where would you place yourself on a scale from 0 to 10, on which 0 means "left" and 10 means "right"? & Ratio Scale & 12 [extremely left to extremely right, don't know, no answer] \\\midrule
    
                IPr.2 &  How interested are you in politics in general? & Ratio Scale & 6 [not interested at all, rather not interested, somewhat interested, very interested, don't know, no answer]   \\ \midrule
      IPr.3 & On a scale from 0 (no trust) to 10 (full trust), how much do you trust the following institutions, organizations and groups? & Group & 2 questions\\
        \qquad IPr.3.1 & \qquad City council (government)  & Ratio Scale & 10 [no trust,
    very low trust,
    low trust,
    moderate trust,
    neutral,
    moderate trust,
    moderate high trust,
    high trust,
    very high trust,
    full trust
    ] \\ 
        
        \qquad IPr.3.2 & \qquad Social media  & Ratio Scale & 10 [no trust,
    very low trust,
    low trust,
    moderate trust,
    neutral,
    moderate trust,
    moderate high trust,
    high trust,
    very high trust,
    full trust
    ]\\ \midrule \midrule
    
     \multicolumn{4}{c}{{\bf Outcome expectations}}\\ \midrule \midrule

        VPr.1 & Which method to you prefer for the selection of the projects? Please rank them from 1 to 3. Options are Method 1: most votes, Method 2: most of the budget, Method 3: satisfy most voters & Multiple - Ratio Scale & 5 [most preferred, second most preferred, third most preferred, don't know, no answer] \\\midrule
    
       VPr.2 & On a scale of 1 to 5, how important do you think these criteria are for the selection of projects to implement at a local level? (such as measures for climate adaptation or economic promotion)? & Group & 4 questions\\
        \qquad  VPr.2.1 & \qquad Cost efficiency & Ratio Scale & 7 [not important,
very less important,
moderately important,
important,
highly important, don't know, no answer] \\ 
        \qquad  VPr.2.2 & \qquad Environmental impact & Ratio Scale & 7 [not important,
very less important,
moderately important,
important,
highly important, don't know, no answer] \\ 
        \qquad  VPr.2.3 & \qquad Benefit for city & Ratio Scale & 7 [not important,
very less important,
moderately important,
important,
highly important, don't know, no answer] \\ 
        \qquad  VPr.2.4 & \qquad Benefit for myself & Ratio Scale & 7 [not important,
very less important,
moderately important,
important,
highly important, don't know, no answer] \\ \midrule
    \end{longtable}}

    \newpage
    
    {\scriptsize
    \begin{longtable}
 {p{0.1\textwidth}p{0.35\textwidth}p{0.1\textwidth}p{0.3\textwidth}}
        \caption{Pre-voting (Pr) survey : Project preferences }\label{table:prologue2} \\
        \toprule
        \textbf{ID} & \textbf{Question} & \textbf{Type}& \textbf{Options} \\
        \midrule
        \endfirsthead 
        \multicolumn{4}{c}%
        {{\bfseries \tablename\ \thetable{} -- continued from previous page}} \\
        \toprule
        \textbf{ID} & \textbf{Question} & \textbf{Type}& \textbf{Options} \\
        \midrule
        \endhead 
    
        \bottomrule
        \multicolumn{4}{r}{{Continued on next page}} \\
        \endfoot
    
        \bottomrule
        \endlastfoot

          PPr.1 & You now see nine thematic areas in which urban projects can be realized. Please select the ones you support.  The nine areas are Education, Urban greenery (e.g. parks, greenery), Public space (e.g. squares), Welfare (for people living below the poverty line), Culture, Environmental protection, Public transit and roads, Sports, and Health & Multiple choice & 2 [no, yes] \\ \midrule
          
          PPr.2 & On a scale of 1 to 5, how important is it to you that the following group benefits from urban projects? & Group & 6 questions \\
        \qquad PPr.2.1 & \qquad Families with children & Ratio Scale & 7 [not important,
very less important,
moderately important,
important,
highly important, don't know, no answer]  \\ 
        \qquad PPr.2.2 & \qquad Children & Ratio Scale & 7 [not important,
very less important,
moderately important,
important,
highly important, don't know, no answer]  \\ 
        \qquad PPr.2.3 & \qquad Youth & Ratio Scale & 7 [not important,
very less important,
moderately important,
important,
highly important, don't know, no answer]  \\ 
        \qquad PPr.2.4 & \qquad Adults & Ratio Scale & 7 [not important,
very less important,
moderately important,
important,
highly important, don't know, no answer]  \\ 
        \qquad PPr.2.5 & \qquad People with disabilities & Ratio Scale & 7 [not important,
very less important,
moderately important,
important,
highly important, don't know, no answer]  \\ 
        \qquad PPr.2.6 & \qquad Elderly & Ratio Scale & 7 [not important,
very less important,
moderately important,
important,
highly important, don't know, no answer] 
        \\\midrule

    \end{longtable}}

 {\scriptsize
\begin{longtable}{>{\color{black}}p{0.1\textwidth} >{\color{black}}p{0.35\textwidth} >{\color{black}}p{0.1\textwidth} >{\color{black}}p{0.3\textwidth}}

    \caption{\textcolor{black}{Pre-voting (Pr) survey: Digital  literacy}}\label{table:prologue3} \\
    \toprule

    \textbf{ID} & \textbf{Question} & \textbf{Type} & \textbf{Options} \\
    \midrule
    \endfirsthead 
    \multicolumn{4}{>{\color{black}}c}%
    {{\bfseries \tablename\ \thetable{} -- continued from previous page}} \\
    \toprule
    \textbf{ID} & \textbf{Question} & \textbf{Type} & \textbf{Options} \\
    \midrule
    \endhead 

    \bottomrule
    \multicolumn{4}{>{\color{black}}r}{{continued on next page}} \\
    \endfoot

    \bottomrule
    \endlastfoot

    DPr.1 & To what degree do the following statements apply to you? & Group & 2 questions \\
	
    \qquad DPr.1.1 & \qquad I know how to adjust the privacy settings on a mobile phone or tablet & Ratio scale & 7 [completely disagree, disagree, neutral, agree, completely agree, don't know, no answer] \\

    \qquad DPr.1.2 & \qquad I tend to shy away from using digital technologies where possible. & Ratio scale & 7 [completely disagree, disagree, neutral, agree, completely agree, don't know, no answer] \\\midrule 
		
    DPr.2 & In general, how much trust do you have in online voting / e-voting solutions? & Ratio scale & 6 [no trust at all, rather no trust, rather trust, a lot of trust, don't know, no answer] \\\midrule

\end{longtable}}

{\scriptsize
\begin{longtable}{>{\color{black}}p{0.1\textwidth} >{\color{black}}p{0.35\textwidth} >{\color{black}}p{0.1\textwidth} >{\color{black}}p{0.3\textwidth}}
    \caption{\textcolor{black}{Pre-voting (Pr) and Post-voting (Po) survey: Engagement profile}}\label{table:prologue4} \\
    \toprule
    \textbf{ID} & \textbf{Question} & \textbf{Type} & \textbf{Options} \\
    \midrule
    \endfirsthead
    \multicolumn{4}{>{\color{black}}c}{{\bfseries \tablename\ \thetable{} -- continued from previous page}} \\
    \toprule
    \textbf{ID} & \textbf{Question} & \textbf{Type} & \textbf{Options} \\
    \midrule
    \endhead

    \bottomrule
    \multicolumn{4}{>{\color{black}}r}{{Continued on next page}} \\
    \endfoot

    \bottomrule
    \endlastfoot

    EPr.1 & How often do you interact with the following persons? & Group  & 2 questions \\
    \qquad EPr.1.1 & \qquad Other inhabitants of Aarau & Ratio Scale & 7 [daily, weekly, quarterly, annually, never, don't know, no answer] \\
    \qquad EPr.1.2 & \qquad Members of Residents' Council & Ratio Scale & 7 [daily, weekly, quarterly, annually, never, don't know, no answer] \\ \midrule

    EPo.6 & What were your reasons to participate in the Stadtidee vote? You may tick more than one answer. & Multiple choice & 11 [support for one or more projects, interest in a new form of participation, civic duty, to have my say on how the local budget is spent, to know what Stadtidee is about, to experience the online voting platform, someone encouraged me, many others have also participated, other reason (please state), don't know, no answer] \\ \midrule
\end{longtable}}

 {\scriptsize
\begin{longtable}{>{\color{black}}p{0.1\textwidth} >{\color{black}}p{0.35\textwidth} >{\color{black}}p{0.1\textwidth} >{\color{black}}p{0.3\textwidth}}

    \caption{\textcolor{black}{Post-voting (Po) survey: Trust}}\label{table:prologue5} \\
    \toprule

    \textbf{ID} & \textbf{Question} & \textbf{Type} & \textbf{Options} \\
    \midrule
    \endfirsthead 
    \multicolumn{4}{>{\color{black}}c}%
    {{\bfseries \tablename\ \thetable{} -- continued from previous page}} \\
    \toprule
    \textbf{ID} & \textbf{Question} & \textbf{Type} & \textbf{Options} \\
    \midrule
    \endhead 

    \bottomrule
    \multicolumn{4}{>{\color{black}}r}{{Continued on next page}} \\
    \endfoot

    \bottomrule
    \endlastfoot

    TPo.1 & What's your impression of the Stadtidee voting result? Rate the following statements on a scale from 0 (do not agree at all) to 10 (fully agree). & Group of questions & 4 questions \\
    \qquad TPo.1.1 & \qquad I am satisfied with the outcome & Ratio scale & 13 [do not agree at all [0] to fully agree [10], don't know, no answer]  \\
    \qquad TPo.1.2 & \qquad I accept the outcome & Ratio scale & 13 [do not agree at all [0] to fully agree [10], don't know, no answer] \\
    \qquad TPo.1.3 & \qquad I was able to influence the outcome & Ratio scale & 13 [do not agree at all [0] to fully agree [10], don't know, no answer] \\
    \qquad TPo.1.4 & \qquad I feel the outcome of the Stadtidee votes accurately represents the will of Aarau citizens & Ratio scale & 13 [do not agree at all [0] to fully agree [10], don't know, no answer] \\ \midrule

\end{longtable}}

\clearpage
   {\scriptsize
    \begin{longtable}{p{0.2\textwidth}p{0.73\textwidth}}
        \caption{ {\bf Prompt design to construct AI voting personas for participatory budgeting campaign - City Idea [Survey] and [Actual] in Aarau}. The prompts are shown for selected ballot formats and personal human traits, using projects from the survey voting scenario.}\label{table:prologuePersona} \\
        \toprule
       \textbf{Personal human traits} &  \textbf{Prompts}\\ \midrule
    
       Socio-demographics. \newline Approval ballot  & Among the following list of projects: P1:  \underline{Bins for Litter}, cost is  \underline{5000 CHF};  P2:  \underline{Elderly Fun}, cost is  \underline{10,000 CHF};  P3:  \underline{Local Park}, cost is  \underline{30,000 CHF}; P4:  \underline{Mental Health}, cost is  \underline{15,000 CHF}; P5: \underline{Bike Lane}, cost is  \underline{40,000 CHF}  with a total budget of \underline{50,000 CHF} \newline
       
    {\em Which projects are preferred for a person with the following profile?}  \newline
    
    \underline{male}, \underline{49.0 years old}, \underline{lives in Zelgli}, \underline{citizen of Switzerland}, has education at the level of \underline{ Master's degree}, \underline{not born} in Switzerland, whose both parents \underline{were born} in Switzerland,  does \underline{not have} children \\
    \midrule
      
       Political interests. \newline Score ballot & Among the following list of projects: P1:  \underline{Bins for Litter}, cost is  \underline{5000 CHF};  P2:  \underline{Elderly Fun}, cost is  \underline{10,000 CHF};  P3:  \underline{Local Park}, cost is  \underline{30,000 CHF}; P4:  \underline{Mental Health}, cost is  \underline{15,000 CHF}; P5: \underline{Bike Lane}, cost is  \underline{40,000 CHF}  with a total budget of \underline{50,000 CHF}  \newline
       
     {\em Assign a score of 1 to 5, 5 being the highest and 1 being the lowest to the projects for a person with the following profile} \newline
    
    has neutral political orientation (score  \underline{5}), where 1 is left wing orientation and 10 is right wing orientation, 
        \underline{not interested} in local politics of Aarau,  scores the trust in city administration with \underline{4 (moderate trust) }, scores the trust in social media with \underline{3 (low trust)}  where 1 is no trust and 10 is full trust.\\
    \midrule
    
    Project preferences.  \newline Single choice  & Among the following list of projects: P1:  \underline{Bins for Litter}, cost is  \underline{5000 CHF};  P2:  \underline{Elderly Fun}, cost is  \underline{10,000 CHF};  P3:  \underline{Local Park}, cost is  \underline{30,000 CHF}; P4:  \underline{Mental Health}, cost is  \underline{15,000 CHF}; P5: \underline{Bike Lane}, cost is  \underline{40,000 CHF}  with a total budget of \underline{50,000 CHF} \newline
    
    {\em Which one is the most preferred for a person with the following profile? } \newline

    considers projects related to education as \underline{not important},   urban greenery as \underline{not important}, public space as \underline{important}, welfare as \underline{not important}, culture as \underline{not important}, environmental protection as \underline{important}, public transit as \underline{not important}, sports as \underline{important}, health as \underline{not important}  \newline
    
    scores projects that impact the elderly population with \underline{ 3 (moderately important)},    children with \underline{4 (important)},  youth with  \underline{4 (important)},  the   adults with \underline{ 2 (very less important)},   people with disabilities with \underline{ 3 (moderately important)},    elderly population with \underline{ 3 (moderately important)}  where 1 is not important and 5 is highly important \\
    \midrule
    
    \end{longtable}}

 {\scriptsize
    \begin{longtable}{p{0.2\textwidth}p{0.7\textwidth}}
        \caption{{\bf Prompts to construct  AI voting personas for American National Election Studies - 2012, 2016 and 2020.} We have used the same prompts as used in the study of Arghyle et al.~\cite{argyle2023out}.}\label{table:Anes} \\
        \toprule
       \textbf{Personal human traits} &  \textbf{Prompts}\\ \midrule
    
       Socio-Demographics & \textcolor{black}{ Which candidate - Barack Obama or Mitt Romney would be most preferred in the US presidential elections 2012 for a person with the following profile? }\newline
    
    Racially the person is \underline{black}. Ideologically, the person is \underline{extremely liberal}. Politically, the person is a \underline{Democrat}. The person \underline{attends} church. The person is \underline{86 years old}. The person is a \underline{woman}. The person has \underline{no interest} in politics. The person feels \underline{a little good} while seeing  the American flag. \\
    \midrule

    \end{longtable}}

     \section{Consistency of AI choices}\label{sec:supllemethod}

\begin{figure}[!htb]
    \centering
    
    \includegraphics[scale = 0.35]{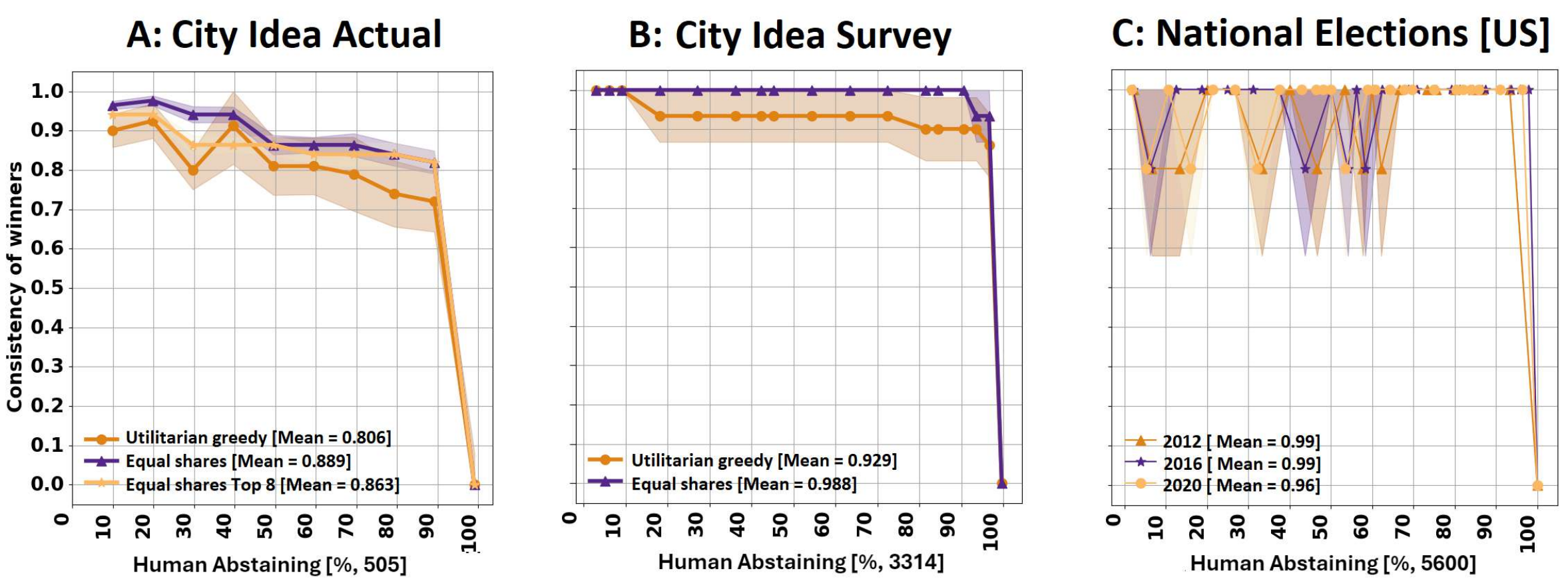}
    \caption{\textcolor{black}{\textbf{Equal shares preserve 83.1\% of the winners even with 80\% of the voters abstaining}.  The
abstaining voters are randomly sampled to analyze the change in the overall
decision outcomes. We use 40 iterations of random sampling and report the average consistency.}}
    \label{fig:mixed populations-abs}
\end{figure}

    \begin{table}[!htb]
    \caption{\small {\bf The winners elected by the equal shares method show greater resilience than utilitarian greedy in retaining projects from the original winner set corresponding to 100\% turnout.} We study project changes, including additions and deletions, by emulating election instances with abstaining voters under different aggregation methods. Abstaining voters are randomly sampled from the population using sizes of 10\%, 25\%, 40\%, 50\%, 75\%, and 85\%, and for each size, the random sampling process is repeated 40 times.
}\label{table:change_gr_ES}
    \centering
    \resizebox{0.8\textwidth}{!}{%
    \begin{tabular}{crrrr}\toprule
 
 &     \multicolumn{2}{c}{\textcolor{black}{Considering all elections}} &  \multicolumn{2}{c}{\textcolor{black}{Considering only elections}} \\ 
  &     \multicolumn{2}{c}{} &  \multicolumn{2}{c}{\textcolor{black}{where the winners change}} \\ \hline
     \multicolumn{5}{c}{\textcolor{black}{Consistency loss: Avg. changes in winners (addition + deletion) }}\\ \hline 
      \textcolor{black}{voters (\% who abstain)}  &  \textcolor{black}{equal shares}  &  \textcolor{black}{utilitarian greedy}  & \textcolor{black}{equal shares}  &  \textcolor{black}{utilitarian greedy}   \\ \hline

    \textcolor{black}{10}	&	\textcolor{black}{0.71}	&	\textcolor{black}{1.32}	&	\textcolor{black}{0.33}	&	\textcolor{black}{1.62}	\\ 
    \textcolor{black}{25}	&	\textcolor{black}{1.38}	&	\textcolor{black}{2.36}	&	\textcolor{black}{0.97}	&	\textcolor{black}{2.19}	\\ 
    \textcolor{black}{40}	&	\textcolor{black}{1.66}	&	\textcolor{black}{2.81}	&	\textcolor{black}{1.77}	&	\textcolor{black}{3.12}	\\ 
    \textcolor{black}{50}	&	\textcolor{black}{1.87}	&	\textcolor{black}{3.59}	&	\textcolor{black}{2.37}	&	\textcolor{black}{3.59}	\\ 
    \textcolor{black}{75}	&	\textcolor{black}{2.27}	&	\textcolor{black}{4.52}	&	\textcolor{black}{3.56}	&	\textcolor{black}{4.52}	\\ 
    \textcolor{black}{85}	&	\textcolor{black}{2.45}	&	\textcolor{black}{4.82}	&	\textcolor{black}{3.86}	&	\textcolor{black}{4.82}	\\ \hline

    \end{tabular}
    }
\end{table}
Voter abstention can influence voting outcomes in collective decision-making. Our analysis reveals that when more than 50\% of voters abstain, the average changes (additions and deletions) in winning projects  compared to original set of winners is 2.14 for  equal shares  and 3.31 for the utilitarian greedy aggregation (Table~\ref{table:change_gr_ES}).  The projects selected as winners with equal shares are highly resilient to abstentions; even with 80\% abstaining, around 83.1\% of the winners are retained  from the original project winner set corresponding to 100\% turnout (refer Figure~\ref{fig:mixed populations-abs}).

\subsection{Individual and collective consistency} \label{sec:condorcet}
\begin{figure}[!htb]
    \centering
    
    \includegraphics[scale = 0.122]{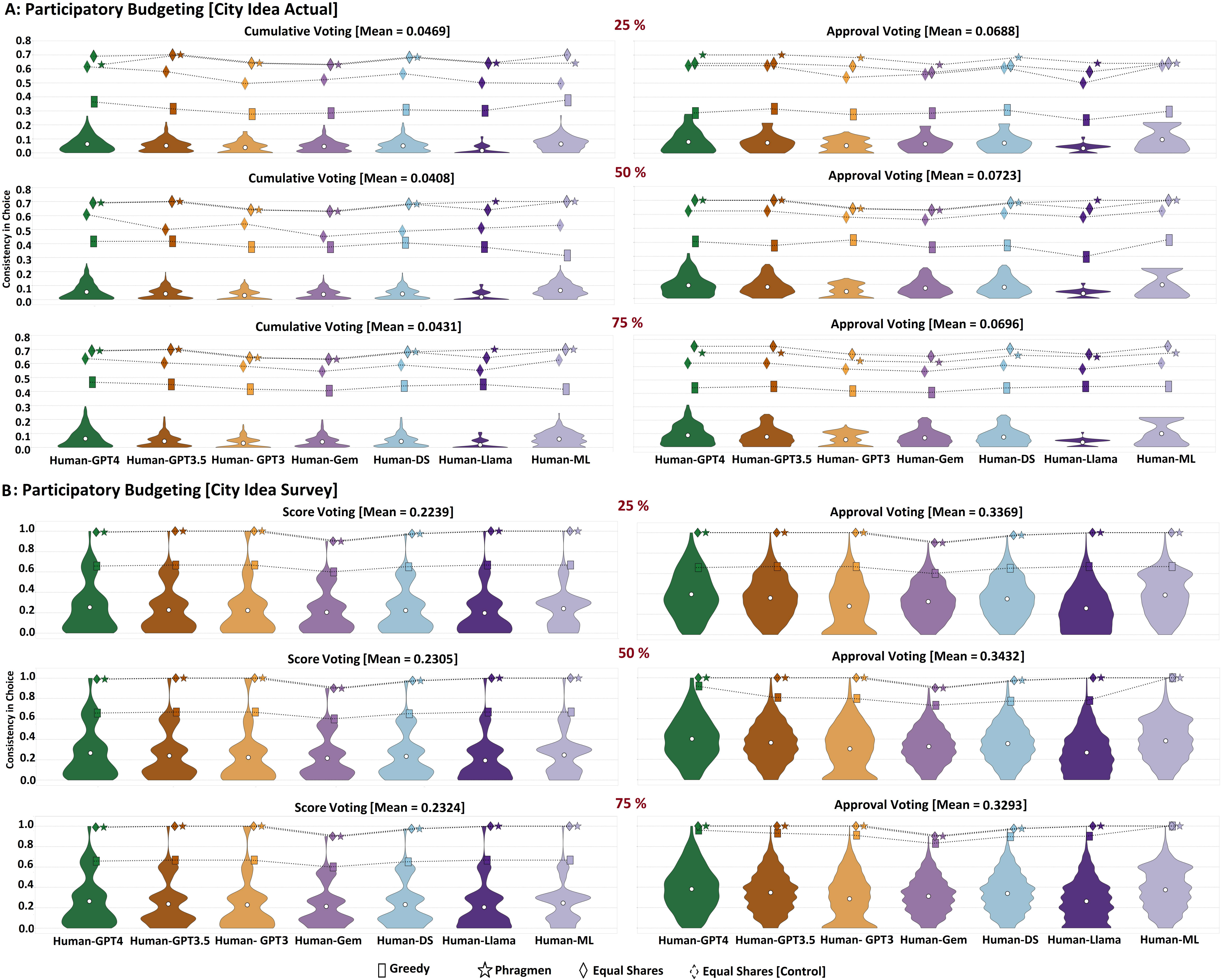}
    \caption{\textcolor{black}{\textbf{The consistency of collective decision-making is significantly higher than that of individual AI choices, especially under fairer ballot aggregation rules such as equal shares and Phragmén's methods. This holds true even within voter subpopulations and when selecting from 33 voting alternatives.} The consistency (y-axis) in individual and collective choice is shown for different AI models (x-axis) for six large language models - {\texttt{GPT 4-o Mini} }, {\texttt{GPT3.5} },  {\texttt{GPT3} }, \texttt{Gemini 1.5 Flash}  (Gem),  { \texttt{Deepseek R1}  (DS)}  and {\texttt{Llama3-8B} (Llama)} along with the predictive AI model ({\em ML}), across the (A) actual and (B) survey participatory budgeting campaign of City Idea for  25\%, 50\% and 75\% of the population. For participatory budgeting, the ballot formats of cumulative (left) and approval (right) are shown, including the ballot aggregation methods of equal shares, Phragmén's and utilitarian greedy. In case of equal shares in the actual voting, the accuracy of winners is calculated for all winners a controlled number of winners (as many as utilitarian greedy) for a fairer comparison.}}
    \label{fig:consistency-biases_actual_division}
\end{figure}

The Condorcet method of pairwise comparisons~\cite{kulakowski2022similarity,navarrete2024understanding} is used to assess the accuracy of human-AI choices at both individual and collective levels. The standardization of AI and human ballots into a uniform preference matrix for project pairs is detailed in Section 4.2 (main paper).  We further analyze  Figure 2 (main paper) using Figure~\ref{fig:consistency-biases_actual_division}  to show the individual consistency representations for the different population sizes.

\begin{figure}[!htb]
    \centering
    
    \includegraphics[scale = 0.235]{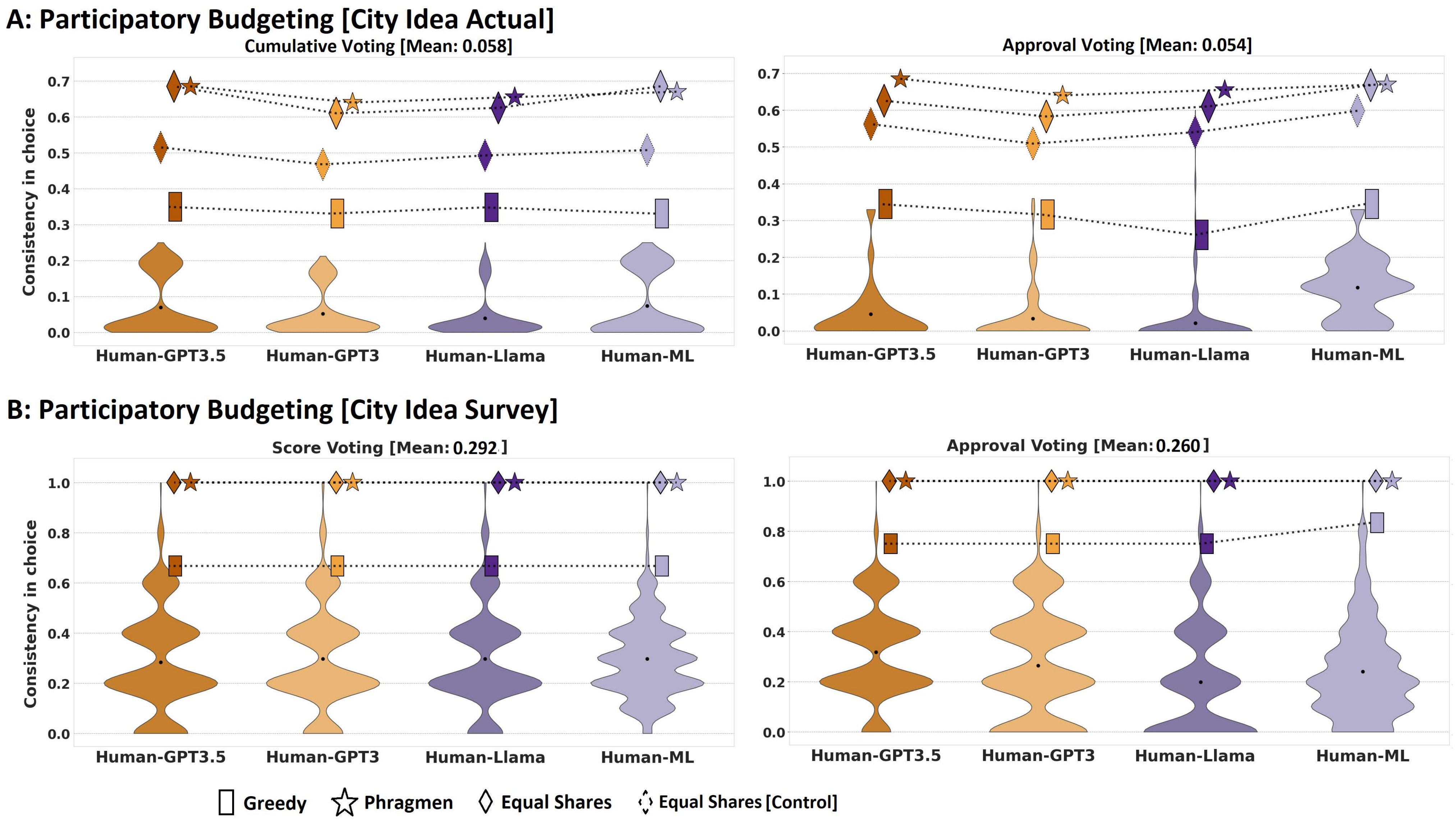}
    \caption{\textcolor{black}{\textbf{The consistency of collective choice is higher than individual choice, particularly for the fairer ballot aggregation rules of equal shares and Phragmén's.} The consistency (y-axis) in individual and collective choice is shown using the Kemeny distance for different AI models (x-axis), across two real-world voting scenarios: The participatory budgeting campaign of City Idea, (A) actual and (B) survey. For participatory budgeting, the ballot formats of cumulative/score (left) and approval (right) are shown, including the ballot aggregation methods of equal shares, Phragmén's and utilitarian greedy. For the actual voting of City Idea, the consistency of equal shares is calculated for all winners and a controlled number of winners (as many as utilitarian greedy) for a fairer comparison. }}
    \label{fig:kemeny}
\end{figure}

In addition to the Condorcet method, we also evaluate consistency using other similarity metrics, such as the Kemeny distance (Figure~\ref{fig:kemeny})~\cite{anshelevich2018approximating}. The Kemeny distance metric measures the number of pairwise inversions needed to align the choice preference orders in two ballots. The trends in collective and individual consistency between human and AI choices using the Kemeny distance and the Condorcet methods are similar (Figures 2 (main paper) and~\ref{fig:kemeny}).

\begin{figure}[!htb]
    \centering
    
    \includegraphics[scale = 0.40]{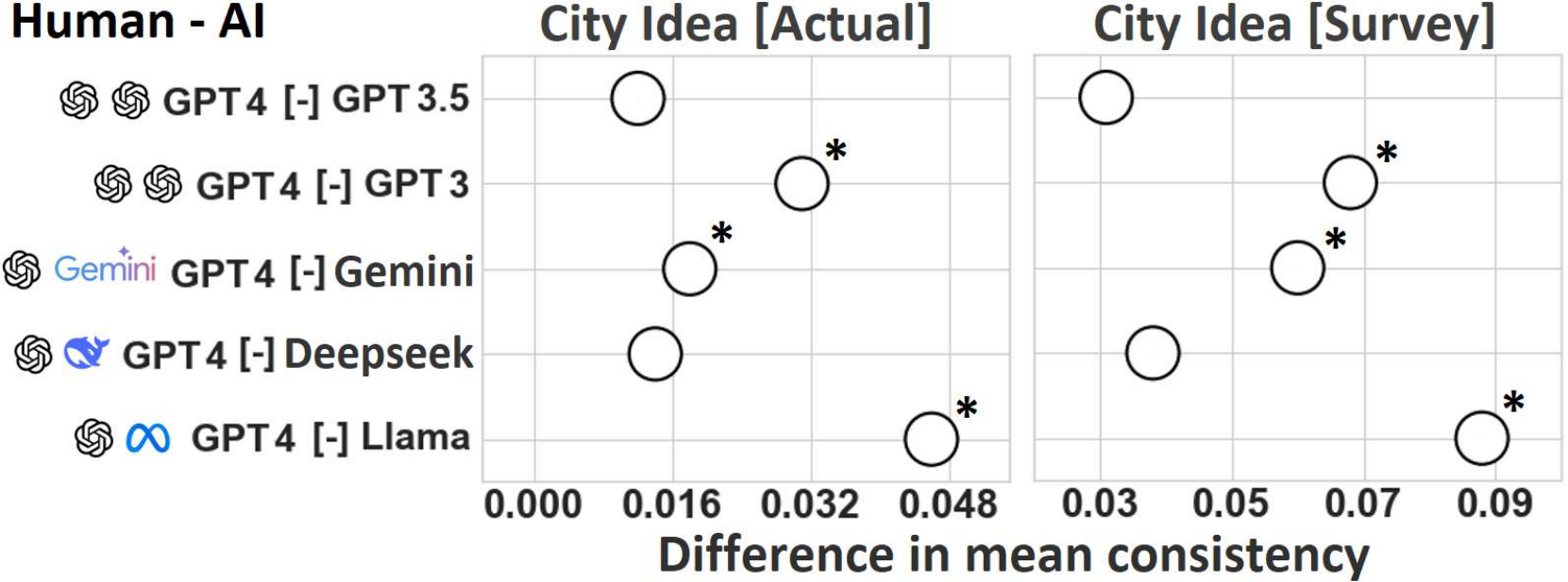}
    \caption{\textcolor{black}{{The difference in mean human-AI consistency (x-axis) for individual choice is shown for {\texttt{GPT 4-o Mini}} and five other large language models: {\texttt{GPT 4-o Mini}}, {\texttt{GPT3.5}}, {\texttt{GPT3}}, \texttt{Gemini 1.5 Flash} (Gemini), {\texttt{Deepseek R1}} (Deepseek), and {\texttt{Llama3-8B}} (Llama). This represents the average difference in consistency, considering probable and score/cumulative ballots. $^*$ indicates that the difference is statistically significant.}}}
    \label{fig:consistency-summary}
\end{figure}

\subsection{Consistency across  LLMs}\label{sec:compareLLM}

{\em Human-AI Consistency}: The consistency between human and AI choices are shown in Figure 2 (main paper). For ballots with a significant number of alternatives, \texttt{GPT 4-o Mini}, the entry level reasoning model, achieves the highest consistency, outperforming \texttt{GPT3.5} and \texttt{GPT3} by 4.7\% (combined p<0.03) and 6.9\% (combined p<0.02), respectively. Compared with open source models, \texttt{GPT 4-o Mini} achieves 4.72\% (combined p<0.04) and 10.2\% (combined p<0.02) higher consistency than \texttt{Deepseek R1}  and \texttt{Llama3-8B}, respectively. In case of ballots with fewer alternatives, \texttt{Deepseek R1}  shows relatively higher consistency but remains 3.9\% lower than \texttt{GPT 4-o Mini}. \texttt{GPT 4-o Mini} further achieves 4.6\%, 4.9\%, and 8.04\% higher consistency than \texttt{Gemini 1.5 Flash}, \texttt{GPT3.5}, and \texttt{GPT3}, respectively. Compared to the proprietary reasoning model \texttt{Gemini 1.5 Flash}, \texttt{GPT 4-o Mini} demonstrates 4.92\% (combined p<0.03) and 4.69\% (combined p<0.04) higher consistency. Overall, \texttt{GPT 4-o Mini} exhibits inconsistencies that are comparable to those of the predictive machine learning model (see Figure~\ref{fig:consistency-summary}).

\noindent {\em Consistency between AI ballots}: The consistency between the different AI ballots has been demonstrated in Figure 3 (main paper). Among the open-source models, \texttt{Llama3-8B} achieves the highest consistency across different ballot formats for AI choices, with an average of 76.2\%.  This is followed by \texttt{GPT 4-o Mini} (74.3\%), \texttt{GPT3.5} (72.1\%), \texttt{Gemini 1.5 Flash}  (71.23\%) and \texttt{Deepseek R1}  (68.7\%).

 \subsection{Abstaining models} \label{sec:abstainingStats}

We present the degree of overlap between the abstaining models. Among the 252 voters who took part in both the pre- and post-voting surveys and the actual voting, 115 have low digital literacy, 126 have low engagement interest, and 106 have low trust in institutions. 10 voters have all three traits, low digital skills, low engagement, and low trust. Additionally, 25 voters have both low trust and low engagement, 23 have low digital skills and low trust, and 28 have low digital skills and low engagement. The minimal overlap among these groups validates the approach of separate abstaining groups in the voting scenario.
    
\subsection{Consistency recovery using AI representation}\label{sec:recoverysupple}
\begin{figure}[!htb]
    \centering
    \includegraphics[scale = 0.29]{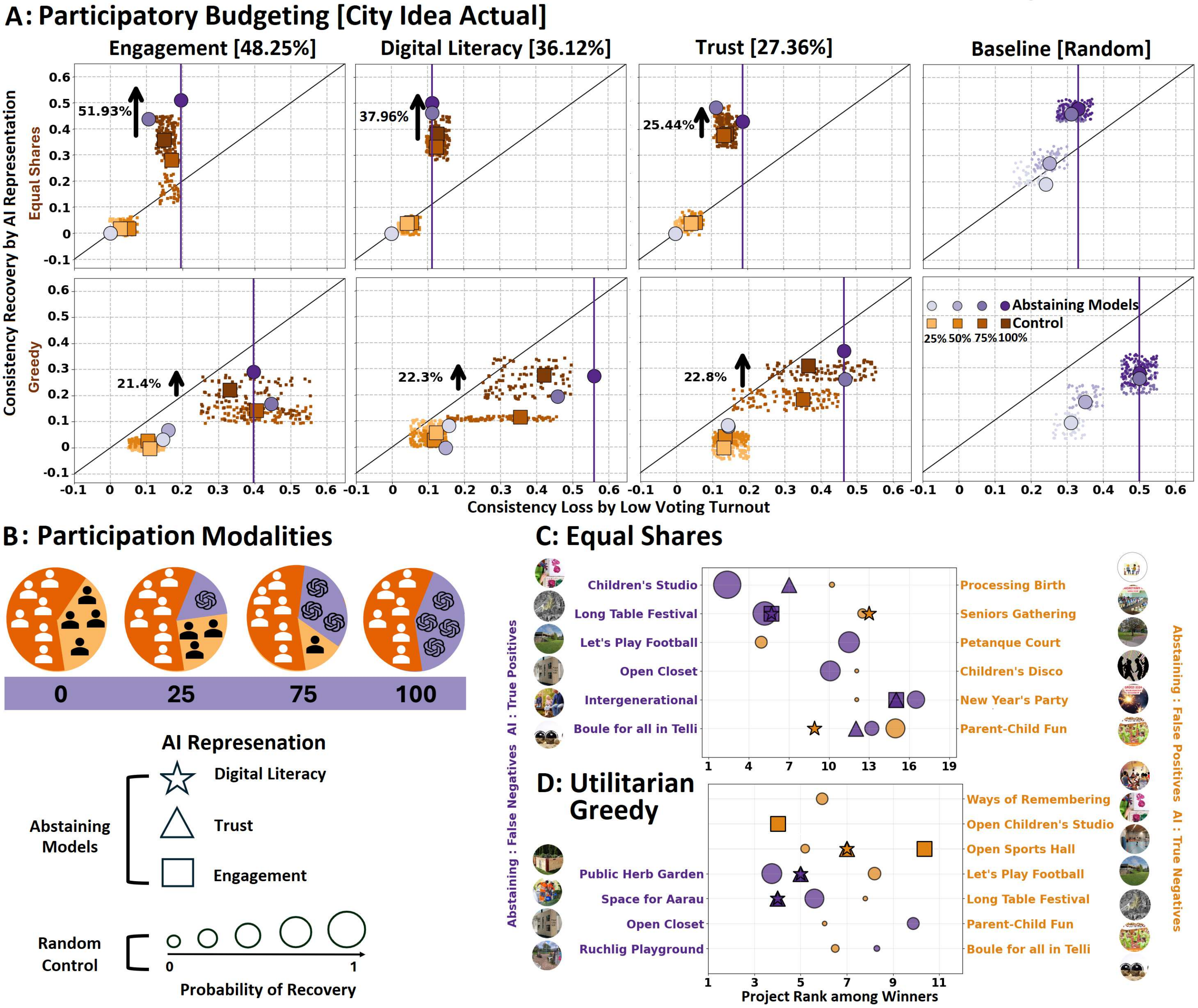}
    \caption{\textcolor{black}{\textbf{Representing more than half of human abstaining voters with AI results in significant consistency recovery, in particular for fair ballot aggregation methods. }
    The consistency loss in voting outcomes by low voters turnout (x-axis) is emulated by removing different ratios of human voters (25\%, 50\%, 75\% and 100\%) among the whole population (baseline) and those who are likely to abstain: low engagement, trust and digital literacy profile (\% of the abstaining populations in the brackets on top). A consistency recovery (y-axis) is hypothesized by AI representation using {\bf \texttt{GPT 4-o Mini}} for the  (A) actual participatory budgeting campaign of City Idea,  (B) studied participation modalities, (C)-(D)  origin of consistency recovery in participatory budgeting for utilitarian greedy and equal shares respectively. Abstaining voters result in falsely removing (left) and erroneously adding (right) winning projects. AI representatives add back and remove these projects respectively to recover consistency. The projects and their probability to recover consistency under random control are shown for comparison.}}
    \label{fig:consistency-recovery-GPT4}
\end{figure}

\begin{figure}[!htb]
    \centering
    \includegraphics[scale = 0.327]{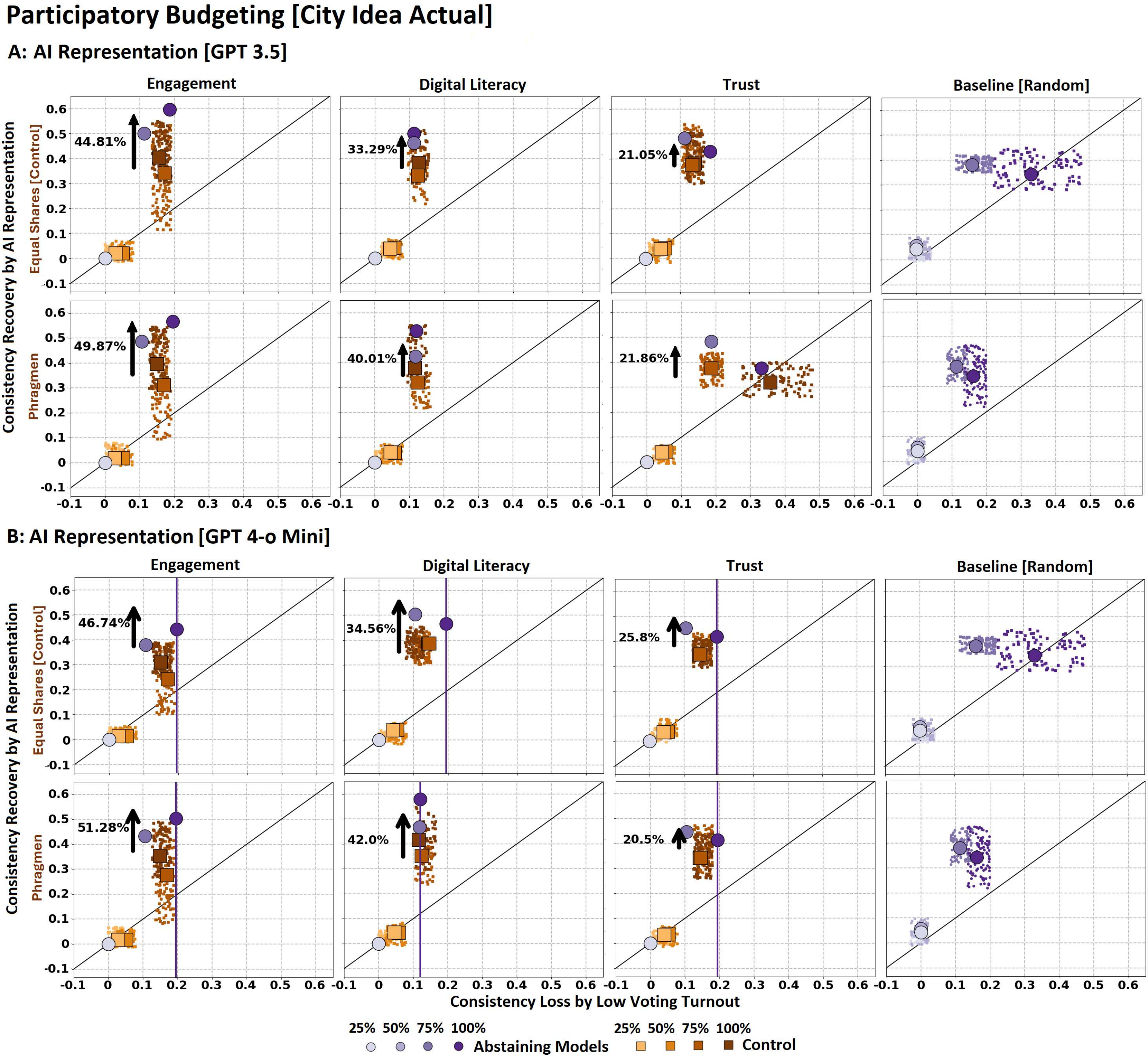}
    \caption{\textcolor{black}{{\textbf{AI representation of abstaining voters is more effective than representing arbitrary voters (random control) under the fair aggregation rules of Phragmén's method and equal shares (controlled settings with number of winners same as utilitarian greedy).}} The consistency loss in voting outcomes by low voters turnout (x-axis) is emulated by removing different ratios of human voters (25\%, 50\%, 75\% and 100\%) among the whole population (baseline) and those who are likely to abstain: low engagement, trust and digital literacy profile. A consistency recovery (y-axis) is hypothesized by AI representation using (A) \texttt{GPT3.5}  and (B) \texttt{GPT 4-o Mini} for the actual participatory budgeting campaign of City Idea. }}
    \label{fig:consistency-recovery_remain}
\end{figure}

\begin{figure}[!htb]
    \centering

    \includegraphics[scale = 0.327]{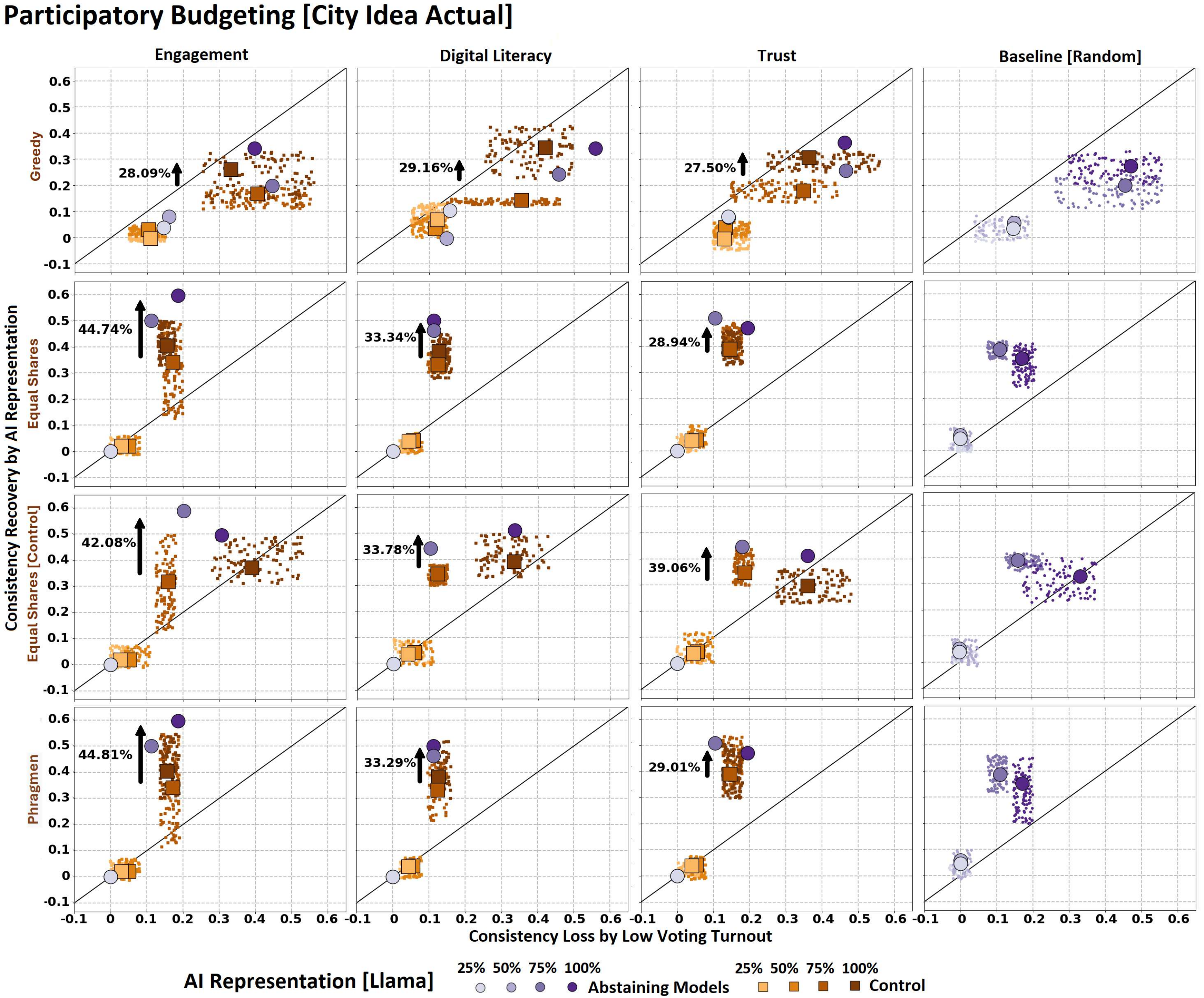}
    \caption{\textcolor{black}{
\textbf{AI representation of abstained voters is more effective than representing arbitrary voters (random control).  }The consistency loss in voting outcomes by low voters turnout (x-axis) is emulated by removing different ratios of human voters (25\%, 50\%, 75\% and 100\%) among the whole population (baseline) and those who are likely to abstain: low engagement, trust and digital literacy profile. A consistency recovery (y-axis) is hypothesized by AI representation using {\bf \texttt{Llama3-8B}} for the  actual  participatory budgeting campaign of City Idea. 
}}
    \label{fig:consistency-recovery_llama}
\end{figure}

\begin{figure}[!htb]
    \centering

    \includegraphics[scale = 0.327]{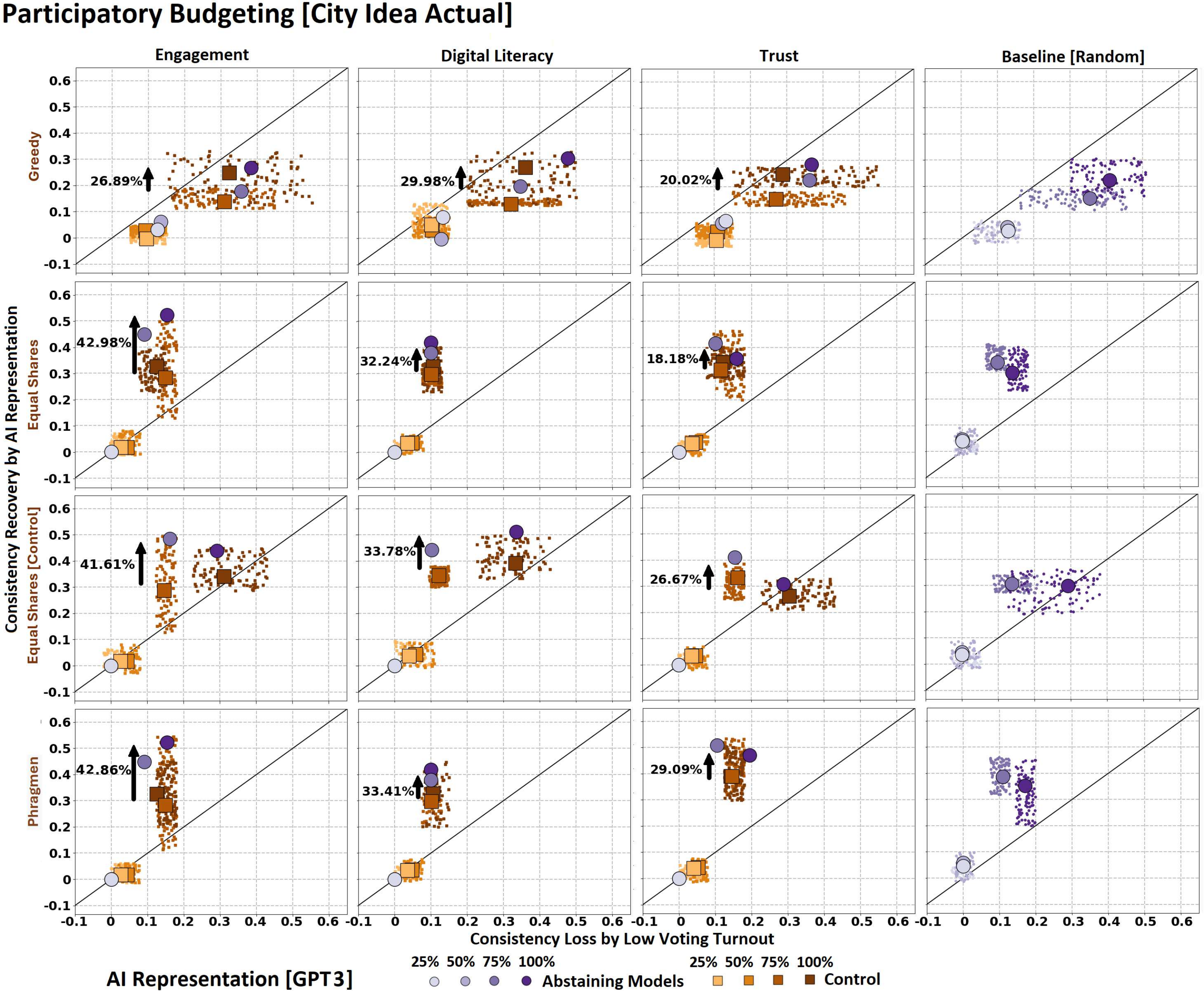}
     \caption{\textcolor{black}{
\textbf{AI representation of abstained voters is moderately effective than representing arbitrary voters (random control).   }The consistency loss in voting outcomes by low voters turnout (x-axis) is emulated by removing different ratios of human voters (25\%, 50\%, 75\% and 100\%) among the whole population (baseline) and those who are likely to abstain: low engagement, trust and digital literacy profile. A consistency recovery (y-axis) is hypothesized by AI representation using { \texttt{GPT3} } for the  actual participatory budgeting campaign of City Idea. 
}}
    \label{fig:consistency-recovery_GPT3}
\end{figure}

\begin{figure}[!htb]
    \centering

    \includegraphics[scale = 0.36]{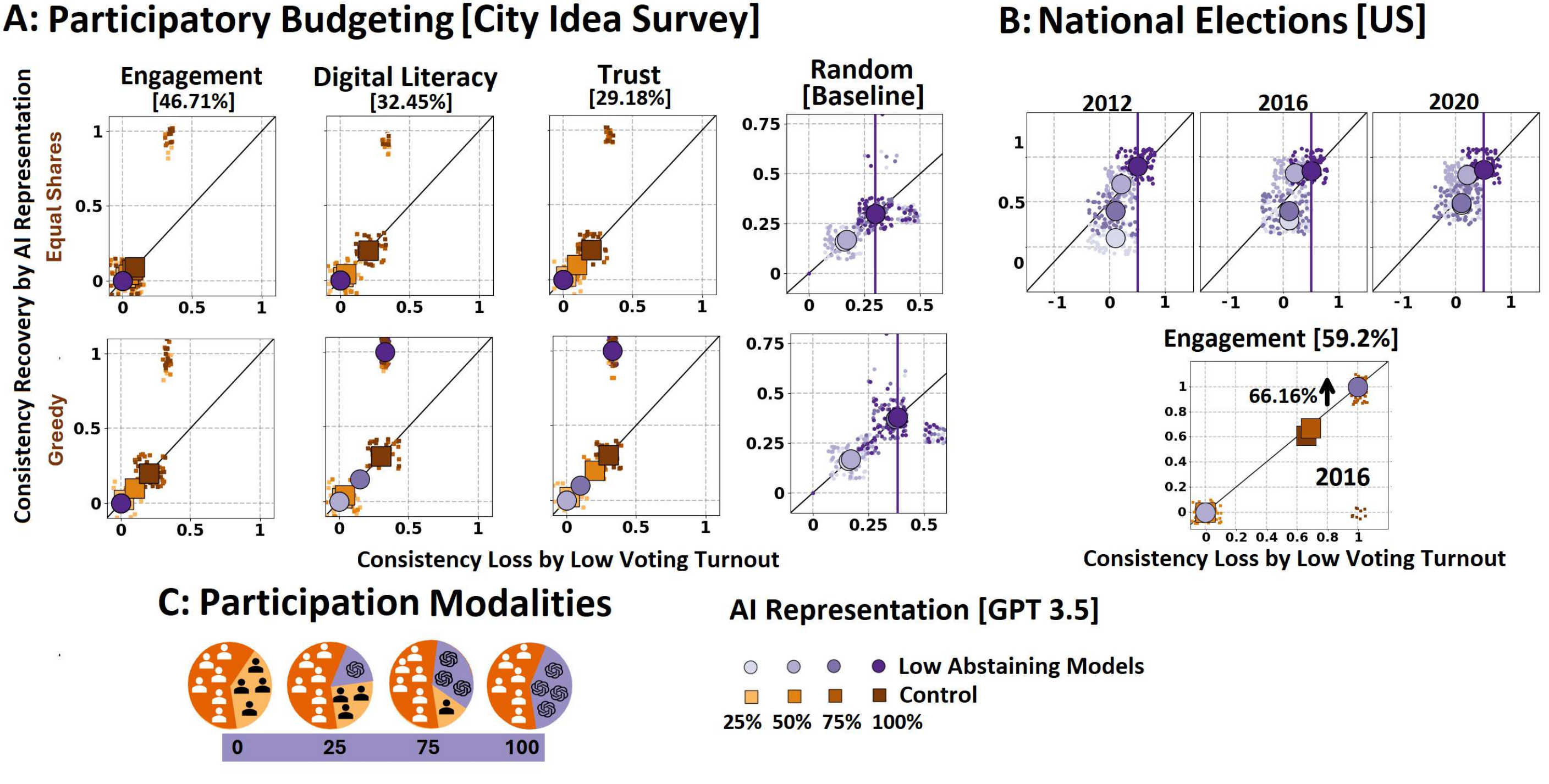}
    \caption{\textcolor{black}{\textbf{Representing more than half of human abstaining voters with AI results in significant recovery of consistency, in particular for fair ballot aggregation methods such as equal shares in participatory budgeting. Strikingly, for voters likely to abstain, collective consistency would remain intact when using equal shares without any AI representation. However, the consistency loss under the utilitarian greedy approach is recovered through AI representation, proving more effective than representing an equivalent number of random voters.}
    The consistency loss in the voting outcome by low voters turnout (x-axis) is emulated by removing different ratios of human voters (25\%, 50\%, 75\% and 100\%), who are likely to abstain with a low engagement, trust and digital literacy profile. A recovery of consistency (y-axis) is hypothesized by AI representation using \texttt{GPT3.5}. The (A) survey voting in the participatory budgeting campaign of City Idea, (B) US elections and (C) the studied participation modalities. }}
    \label{fig:consistency-recovery-survey-US}
\end{figure}

\begin{figure}[!htb]
    \centering

    \includegraphics[scale = 0.327]{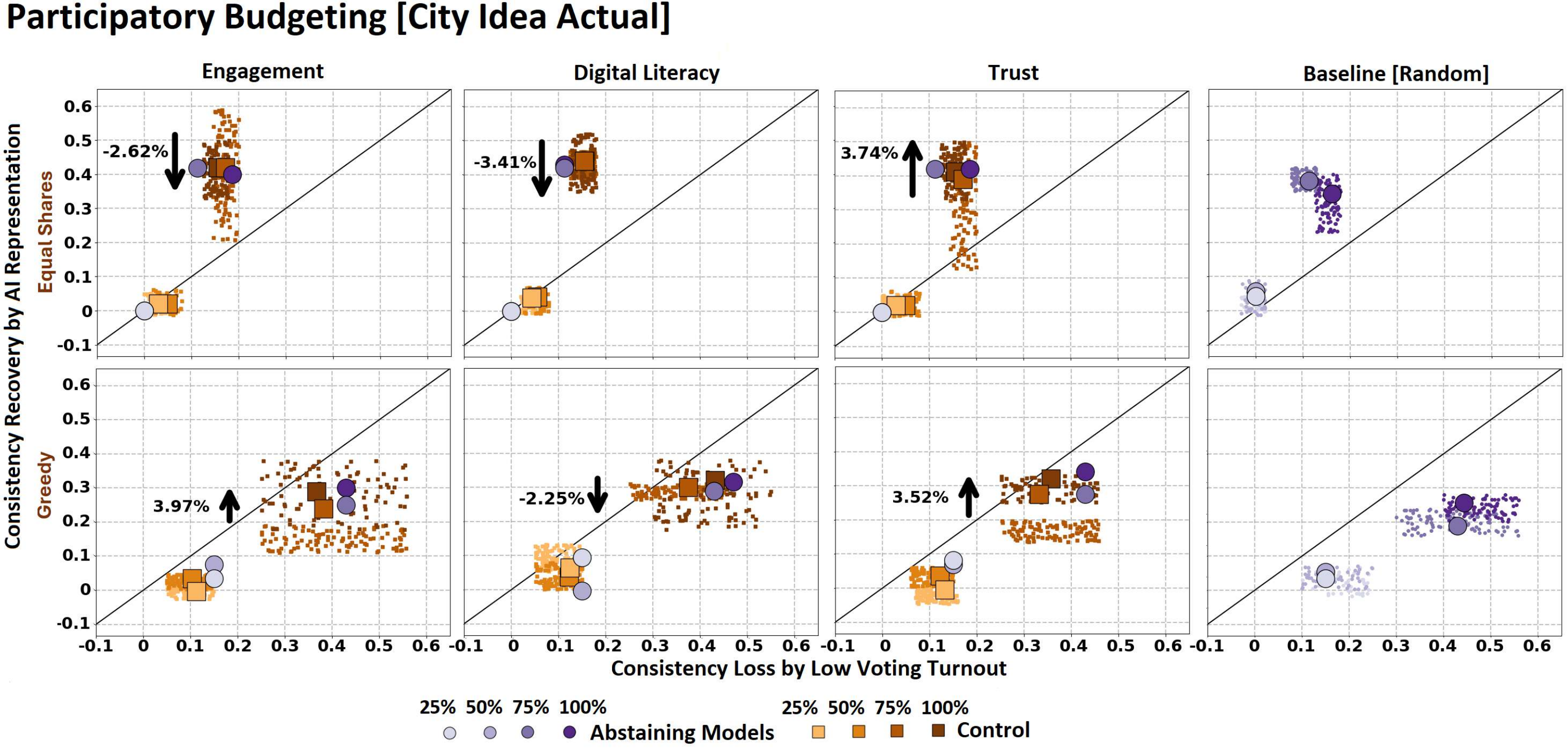}
    \caption{\textcolor{black}{\textbf{The AI representation of voters who come with a more active participation profile, without typical features of abstaining voters,  is not significantly more effective than representing arbitrary voters (random control).} The consistency loss in voting outcomes by low voters turnout (x-axis) is emulated by removing different ratios of human voters (25\%, 50\%, 75\% and 100\%) among the whole population (baseline) and those who are likely to abstain: low engagement, trust and digital literacy profile. A consistency recovery (y-axis) is hypothesized by AI representation using \texttt{GPT3.5}  for the actual participatory budgeting campaign of City Idea. }}
    \label{fig:consistency-recovery_opp}
\end{figure}

In this section, we present additional findings on assessing consistency recovery by AI representatives, which extend the findings shown in Section 2.2 (main paper). Figures 4 (main paper) and~\ref{fig:consistency-recovery-GPT4} illustrate the consistency recovery using AI representation of voters who are likely to abstain, and two aggregation methods: equal shares~\cite{peters2021proportional} and utilitarian greedy~\cite{peters2020proportional}, for the actual voting of City Idea, modeled using \texttt{GPT3.5} and \texttt{GPT 4-o Mini}, respectively. Additionally, Figure~\ref{fig:consistency-recovery_remain} demonstrates consistency recovery for another fair aggregation method, Phragmén’s~\cite{brill2024phragmen} method, alongside a controlled instance of equal shares, ensuring the number of winners is the same as utilitarian greedy aggregation, using \texttt{GPT3.5} and \texttt{GPT 4-o Mini}. Similarly, Figures~\ref{fig:consistency-recovery_llama} and~\ref{fig:consistency-recovery_GPT3} depict the consistency recovery for the actual voting scenario using utilitarian greedy, equal shares, Phragmén’s, and equal shares with controlled winners for \texttt{GPT3} and \texttt{Llama3-8B}, respectively. The results on consistency recovery by AI representatives in the survey voting scenario of City Idea have been shown in Figure~\ref{fig:consistency-recovery-survey-US}.
Our findings reveal that AI representation substantially enhances consistency recovery for abstaining voter groups but has a negligible effect when applied to voters who come with a more active participation profile, and without typical features of abstaining voters (Figure~\ref{fig:consistency-recovery_opp}).

\textcolor{black}{Consistency recovery is at two levels: (i) False negative projects removed under abstaining but added back by AI representatives, which are are higher in ranking and number than (ii) false positive projects added under abstaining but removed by AI representatives. Detailed comparisons of false-negative and false-positive projects are provided in Figure 4 (main paper) for \texttt{GPT3.5} and in Table~\ref{tab:false_positives_negatives_all_models} for \texttt{Llama3-8B} and \texttt{GPT3}. The average project recovery rates for false negatives and false positives, based on abstention models and their respective random control populations, are presented in Table~\ref{tab:recovery}.}

\textcolor{black}{We further analyze the recovery of voting outcomes by examining abstention patterns across different regions (Table~\ref{table:region}). }

We also compare how the pre-election predictions fare against actual polls with human voters and 100\% AI representation for the US national elections of 2012, 2016, and 2020. Interestingly, for the partisan dataset, the predicted winners in the pre-election closely match the actual election winners (for both humans and AI representation), except in 2016, where the pre-election prediction differed. We have taken a subset of the actual election votes and show the relative percentage of votes each candidate received in Table~\ref{tab:USpre}.

  \section{The machine learning framework}\label{sec:mlo}
    
  We discuss the machine learning architecture for predicting consistency gain or loss for individual voters based on their personal human traits in this section. The relevant personal human traits are mapped to cognitive biases for further analysis.

    \subsection{Human cognitive biases in AI collective decision making}\label{sec:bias}
Human choices are significantly influenced by potential cognitive biases that are often a manifestation of socio-economic characteristics, conditions of life quality, (dis)satisfaction with the available public amenities and the overall life experiences of an individual~\cite{korteling2023cognitive}. We map self-reported personal traits to potential underlying human cognitive biases. These traits are part of the input context for ballot generation in large language models. Our goal is to explore whether these biases are reinforced by the models. If so, they may become more likely to manifest under AI representation. Figure 1d (main paper) outlines the mapping we study based on a systematic and comprehensive review of relevant literature. The following types of biases are determined:

\cparagraph{Time-discounting biases} These are characterized by the tendency to receive immediate gratification over a larger but future reward. Projects related to public spaces or culture often focus on events such as annual festivals or cinema nights (alternatives proposed in the participatory budgeting campaign in Aarau~\cite{CityIdeaReport2025}). These projects have a quick turnaround time, offer direct and tangible rewards, and may also create long-term, repeatable impacts. Similarly, the welfare projects proposed in Aarau~\cite{CityIdeaReport2025} involve small-scale initiatives such as educating asylum-seeking children, commemorative activities, and bread tours for the elderly, all of which yield rapid benefits. Therefore, such project are subject of time-discounting biases~\cite{korteling2023cognitive}.

\cparagraph{Optimism bias} Projects such as road construction require significant time and investment costs for resources even before implementation begins. Additionally, uncertainties and challenges related to costs, infrastructure, and planning may arise during execution and delay the project from materialising. Despite these hurdles, such investments contribute to sustainable reforms that benefit society in the long run, fostering optimism among people who continue to support them. Even though 72\% of public transportation infrastructure projects in European cities experience cost overruns, voters still back these initiatives due to an inherent optimism about improving transportation~\cite{korteling2023cognitive,maharjan2024fair}. We refer to this tendency to prioritize long-term sustainability despite economic uncertainties as ‘optimism’~\cite{gollier2011pricing,percoco2006individuals}.

\cparagraph{Surrogation biases} This reflects how humans favor simpler measures to assess the impact over ones that are more precise and harder to evaluate.
Korteling et al.~\cite{korteling2023cognitive} argue that these biases manifest when deciding projects with large societal impact such as health or education, while their outcome is subject of different satisfaction levels among citizens. The project outcomes may be perceived as successful by part of society using easy-to-evaluate metrics instead of looking at long-term effects on the community. For instance, a timely vaccination drive may be preferred over significant changes to vaccination protocols or health insurance policies covering vaccination. Hence these projects are likely to be more preferred as they come with more intuitive ways to assess for the broader population. This is reflected by the average winning rate of 38.1\% and 36.2\% for health and education-related projects in Poland~\cite{maharjan2024fair}, where participatory campaigns have been actively hosted in the last decade.

\cparagraph{Conformity biases} These biases arise out of group pressure under which people make decisions to be socially desirable~\cite{conniff2009using,culiberg2016going}. 
It is argued that a conformity bias may induce voting for green alternatives~\cite{korteling2023cognitive}. Green-themed participatory budgeting campaigns have been adopted in European cities such as Lisbon~\cite{falanga2021green}, to promote green initiatives, aligning to a culture for more sustainable behavior. Poland runs participatory budgeting campaigns at large scale, which include environment and urban greenery projects. These are within the top-5 most popular projects with an average of 22.5\% and 26.5\% respectively~\cite{maharjan2024fair}. Even in Aarau we observe the same trend wherein, with environmental friendly projects accounting for the top-10 most popular projects~\cite{CityIdeaReport2025}.

\cparagraph{Affect heuristic biases} This is the tendency to make decisions based on what intuitively or emotionally feels right. Affect biases have been studied to analyze the inclusive attitudes most people show towards elderly people~\cite{fiander1999role}. Similar biases also manifest in welfare of children and in inter-generational communication~\cite{korteling2023cognitive,acciarini2021cognitive}.  In Aarau, we observe that 71.3\% of the voters prefer projects for younger and elderly people. 

\cparagraph{Biases for altruism and egotism} Individual interest is often in conflict with the community interest in participatory and collective decision-making processes. Intrinsic altruism of citizens influences voting choices. As a result, altruism and egotism are influential for the fairness of voting outcomes and how these outcomes benefit the city in overall~\cite{fehr2006economics,korteling2023cognitive,acciarini2021cognitive}. In Aarau, we observe that 67.1\% of the voters, who prefer better representation in the outcome are prosocial and prioritize city-wide benefit (altruism bias) over individual benefit (egotism bias). 

\cparagraph{Unconscious biases} Human choices are influenced by socio-economic and demographic traits such as race, ethnicity, citizenship, household size and income~\cite{colombo2020heuristics}. Specifically, political ideology and belief shape to a high degree decisions for candidates in elections~\cite{kuklinski2007belief}.

\begin{table}[h]
\raggedright
\caption{\textbf{Voter abstention can cause incorrect removal of projects (false negatives), or incorrect addition of projects (false positives), in the winning set compared to the original winner set at 100\% turnout.} The findings are shown for the AI representation using \texttt{Llama3-8B} and \texttt{GPT3}.}

\label{tab:false_positives_negatives_all_models}

\resizebox{0.9\textwidth}{!}{
\begin{tabular}{lllll}
\hline
\textbf{Project types} &  &  &  &  \\
\textbf{Aggregation, AI models} & \textbf{Projects} & \textbf{Probability} & \textbf{Rank} & \textbf{Abstaining models, Rank} \\
\hline

False negatives & Boule for all in Telli & 0.34 & 14 & Digital literacy, 13 \\

Equal shares, \texttt{Llama3-8B} & New edition of Telli map & 0.42 & 12.5 & Digital literacy, 13; Engagement, 12 \\
 & Open Sports Hall & 0.83 & 10.7 &  \\
 & Long Table Festival & 0.81 & 5.4 & Digital literacy, 13; Engagement, 12; Trust, 13 \\
 & Let's Play Football & 0.65 & 9.5 &  \\
\hline

False positives & A Garden for all & 0.49 & 12.6 &  \\
Equal shares, \texttt{Llama3-8B} & Petanque Court & 0.64 & 8.1 & Digital literacy, 9; Engagement, 9 \\
 & New Year's Party & 0.12 & 9.2 &  \\
 & Children's Disco & 0.37 & 9.6 &  \\
 & Parent-Child Fun & 0.29 & 12.4 & Digital literacy, 9; Engagement, 11; Trust, 11 \\
\hline

False negatives & Boule for all in Telli & 0.14 & 14.2 &  \\
Equal shares, \texttt{GPT3} & Intergenerational Project & 0.52 & 9.5 & Digital literacy, 10; Engagement, 11; Trust, 11 \\
 & Open Closet & 0.73 & 10.7 &  \\
 & Long Table Festival & 0.81 & 7.2 & Digital literacy, 7; Engagement, 8 \\
 & Let's Play Football & 0.62 & 11.5 &  \\
 & Open Children's Studio & 0.75 & 2.8 &  \\
\hline

False positives & Seniors Gathering 70+ & 0.29 & 15.2 & Digital literacy, 14; Trust, 15 \\
Equal shares, \texttt{GPT3} & Petanque Court & 0.52 & 8.8 & Engagement, 8 \\
 & New Year's Party & 0.73 & 10.1 &  \\
 & Children's Disco & 0.62 & 7.2 & Digital literacy, 7; Engagement, 6; Trust, 6 \\
 & Parent-Child Fun & 0.75 & 3.4 &  \\
\hline
\end{tabular}}
\end{table}

\begin{table}[h]
\caption{\textbf{Consistency recovery by abstaining models is more salient for true positives (1.66 vs. 1.0 for true negatives), whereas in random control populations, it favors true negatives (2.27 vs. 1.89 for true positives).} The recovery for abstaining and control populations (randomly sampled 40 times based on the size of the abstaining group) is analyzed across false negatives, false positives, and different aggregation methods. Recovery is evaluated both for all instances and specifically for cases where project changes occur.}
\label{tab:recovery}
\centering
\resizebox{\textwidth}{!}{%
\begin{tabular}{lcccccc}
\toprule
 & \textbf{Digital literacy} & \textbf{Control [digital literacy]} & \textbf{Engagement} & \textbf{Control [engagement]} & \textbf{Trust} & \textbf{Control [trust]} \\
 & \textbf{\# projects} & \textbf{Avg. projects \%} & \textbf{\# projects} & \textbf{Avg. projects \%} & \textbf{\# projects} & \textbf{Avg. projects \%} \\
\midrule
\multicolumn{7}{c}{All instances} \\ \hline
Equal shares [abstaining: false negatives; AI: true positives] & 1 & 1.81 & 2 & 1.36 & 3 & 2.04 \\
Utilitarian greedy [abstaining: false negatives; AI: true positives] & 2 & 2.43 & 0 & 1.36 & 2 & 2.31 \\
Equal shares [abstaining: false positives; AI: true negatives] & 2 & 2.49 & 0 & 1.31 & 0 & 2.74 \\
Utilitarian greedy [abstaining: false positives; AI: true negatives] & 1 & 2.44 & 2 & 1.98 & 1 & 2.67 \\
Equal shares [all additions and removals] & 3 & 2.15 & 2 & 1.34 & 3 & 2.39 \\
Utilitarian greedy [all additions and removals] & 3 & 2.43 & 2 & 1.67 & 3 & 2.49 \\
\midrule
\multicolumn{7}{c}{Instances where project changes occur} \\ \hline
Equal shares [abstaining: false negatives; AI: true positives] & 1 & 2.01 & 2 & 1.55 & 3 & 2.33 \\
Utilitarian greedy [abstaining: false negatives; AI: true positives] & 2 & 2.43 & 0 & 1.57 & 2 & 2.31 \\
Equal shares [abstaining: false positives; AI: true negatives] & 2 & 2.54 & 0 & 1.31 & 0 & 2.81 \\
Utilitarian greedy [abstaining: false positives; AI: true negatives] & 1 & 2.44 & 2 & 1.98 & 1 & 2.67 \\
Equal shares [all additions and removals] & 3 & 2.27 & 2 & 1.43 & 3 & 2.57 \\
Utilitarian greedy [all additions and removals] & 3 & 2.43 & 2 & 1.77 & 3 & 2.49 \\
\bottomrule
\end{tabular}%
}
\end{table}

\begin{table}[!htb]
    \caption{\textcolor{black}{{ \bf
Collective consistency recovery is higher through AI representation in large districts such as Telli, Zelgli, Schachen, and Innenstadt, where at least 25\% of the 33 proposed projects originate. Additionally, Altstadt and Scheibenschachen, where more than 30\% of the proposed projects are up for voting, also exhibit positive consistency recovery with AI representation.} The consistency recovery is calculated based on the district-specific abstaining voters, adjusted by subtracting the recovery observed in randomly sampled voters of equivalent size under a scenario of 100\% AI representation. The reported recovery figures pertain to \texttt{GPT3.5}, which demonstrates, on average, 18.2\% higher representation than \texttt{Llama3-8B} and 20.14\% higher than \texttt{GPT3}.}}\label{table:region}
    \centering
    \resizebox{0.5\textwidth}{!}{%
    \begin{tabular}{crr}\toprule
    \multicolumn{1}{l}{\textcolor{black}{}} &     \multicolumn{2}{c}{\textcolor{black}{Consistency recovery}}  \\ \hline
  {\bf District}  &  \textcolor{black}{Equal shares}  &  \textcolor{black}{Utilitarian greedy}   \\ \hline
    
    \textcolor{black}{Alstadt}	&	\textcolor{black}{6.11}	&	\textcolor{black}{6.51}	\\ \hline	
    \textcolor{black}{Ausserfield}	&	\textcolor{black}{8.89}	&	\textcolor{black}{0.91}	\\ \hline	
    \textcolor{black}{Binzenhof}	&	\textcolor{black}{-1.55}	&	\textcolor{black}{-8.03}	\\ \hline	
    \textcolor{black}{Damn}	&	\textcolor{black}{8.21}	&	\textcolor{black}{-17.21}	\\ \hline	
    \textcolor{black}{Goldern}	&	\textcolor{black}{5.16}	&	\textcolor{black}{-3.48}	\\ \hline	
    \textcolor{black}{Gonhard}	&	\textcolor{black}{13.22}	&	\textcolor{black}{2.43}	\\ \hline	
    \textcolor{black}{Hinterdorf}	&	\textcolor{black}{-0.9}	&	\textcolor{black}{3.48}	\\ \hline	
    \textcolor{black}{Hungerberg}	&	\textcolor{black}{7.76}	&	\textcolor{black}{1.31}	\\ \hline	
    \textcolor{black}{Innenstadt}	&	\textcolor{black}{8.94} & \textcolor{black}{10.33}	\\ \hline	
    \textcolor{black}{Rossligut}	&	\textcolor{black}{2.28}	&	\textcolor{black}{-0.40}	\\ \hline	
    \textcolor{black}{Schachen}	&	\textcolor{black}{6.14}	&	\textcolor{black}{-7.98}	\\ \hline	
    \textcolor{black}{Scheibenschachen}	&	\textcolor{black}{9.94} & \textcolor{black}{6.54}	\\ \hline	
    \textcolor{black}{Seibenmatten}	&	\textcolor{black}{4.78}	&	\textcolor{black}{-3.31}	\\ \hline	
    \textcolor{black}{Torfeld Nod}	&	\textcolor{black}{3.21}	&	\textcolor{black}{-2.18}	\\ \hline	
    \textcolor{black}{Torfeld Sud}	&	\textcolor{black}{4.88}	&	\textcolor{black}{3.01}	\\ \hline	
    \textcolor{black}{Telli}	&	\textcolor{black}{10.34}	&	\textcolor{black}{4.85}	\\ \hline	
    \textcolor{black}{Zelgi}	&	\textcolor{black}{22.22} & \textcolor{black}{2.84}	\\ \hline	
    \end{tabular}
    }
\end{table}

\begin{table}[ht] \centering \caption{\textcolor{black}{Comparison of Human, \texttt{GPT3.5} representation, \texttt{GPT 4-o Mini} representation, and and the actual Pre-election predictions (US National elections 2012–2020): \% of votes in favor of each candidate shown. The AI representation is emulated for 100\% of the sampled population. Candidates 1 and 2 are the electoral candidates contesting the election.}} \label{tab:USpre} \resizebox{0.6\textwidth}{!}{ \begin{tabular}{lcccc} \hline \textcolor{black}{\textbf{Year}} & \textcolor{black}{\textbf{Human}} & \textcolor{black}{\textbf{\texttt{GPT3.5}}} & \textcolor{black}{\textbf{\texttt{GPT 4-o Mini}}} & \textcolor{black}{\textbf{Pre-Elections}} \\ \hline \multicolumn{5}{l}{\textcolor{black}{\textbf{2012}}} \\ \textcolor{black}{Candidate 1} & \textcolor{black}{55.25\%} & \textcolor{black}{61.49\%} & \textcolor{black}{59.14\%} & \textcolor{black}{48.80\%} \\ \textcolor{black}{Candidate 2} & \textcolor{black}{37.45\%} & \textcolor{black}{31.21\%} & \textcolor{black}{33.55\%} & \textcolor{black}{48.10\%} \\ \hline \multicolumn{5}{l}{\textcolor{black}{\textbf{2016}}} \\ \textcolor{black}{Candidate 1} & \textcolor{black}{45.70\%} & \textcolor{black}{50.92\%} & \textcolor{black}{48.11\%} & \textcolor{black}{43.60\%} \\ \textcolor{black}{Candidate 2} & \textcolor{black}{41.73\%} & \textcolor{black}{36.51\%} & \textcolor{black}{39.32\%} & \textcolor{black}{48.60\%} \\ \hline \multicolumn{5}{l}{\textcolor{black}{\textbf{2020}}} \\ \textcolor{black}{Candidate 1} & \textcolor{black}{53.31\%} & \textcolor{black}{51.47\%} & \textcolor{black}{50.59\%} & \textcolor{black}{51.30\%} \\ \textcolor{black}{Candidate 2} & \textcolor{black}{37.33\%} & \textcolor{black}{39.18\%} & \textcolor{black}{45.30\%} & \textcolor{black}{46.80\%} \\ \hline \end{tabular}} \end{table}

\begin{figure}[!htb]
    \centering
    \includegraphics[scale = 0.20]{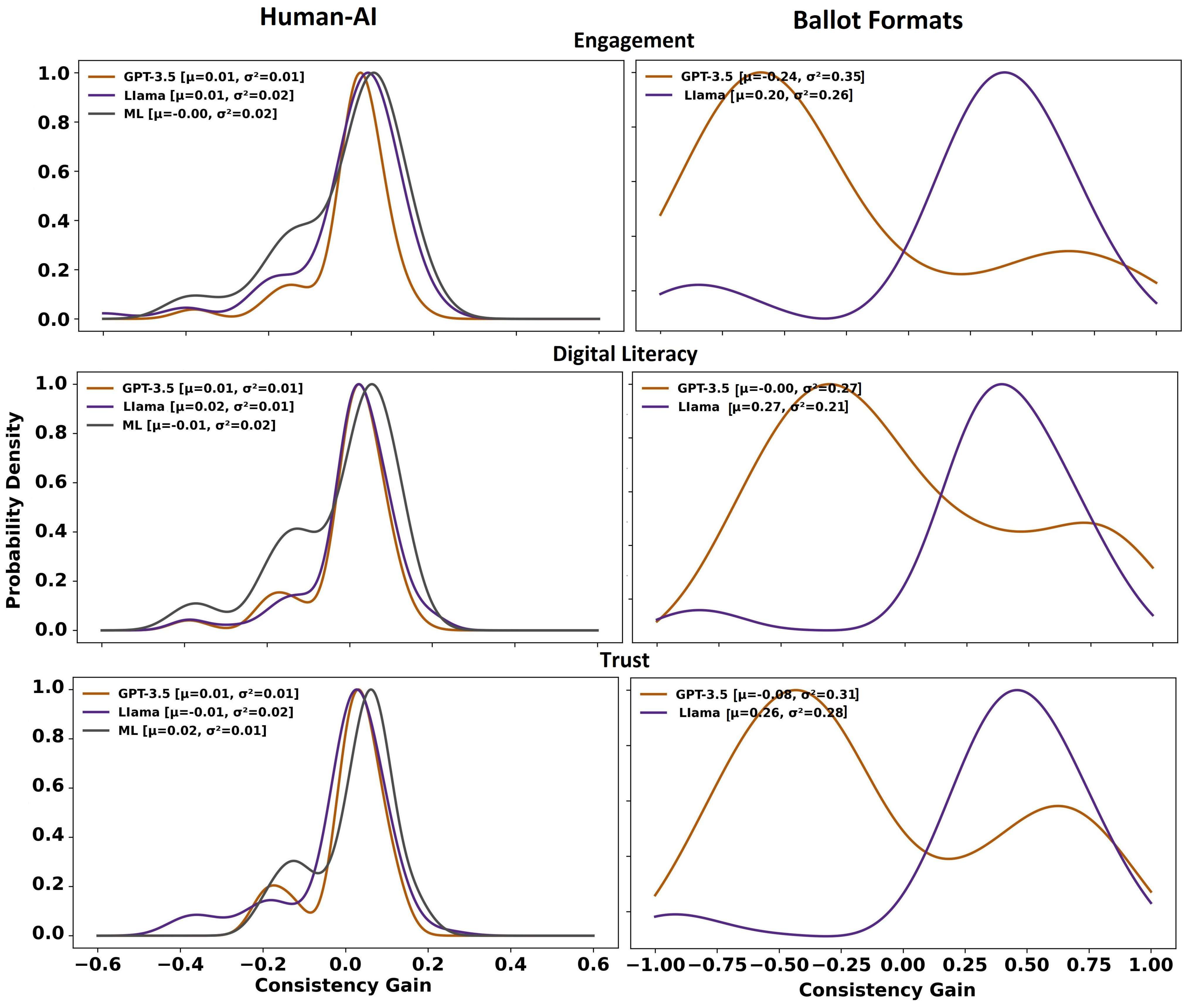}
    \caption{\textcolor{black}{\textbf{ 
The divergence in AI choices for abstaining voters, relative to the random baseline population, tends to be neutral, with a slight inclination towards gain. }
   The AI representation of three abstaining models (low engagement, trust, digital literacy) is evaluated for accuracy against human choices and transitivity across ballot formats for each voter by comparison with random voters. For voters in the abstaining group, the average difference in human–AI accuracy or transitivity between ballot formats is computed by randomly sampling voters, using sample sizes of 20, 30, and 40.
}}
    \label{fig:consistency-gain}
\end{figure}

    \subsection{Fairness in machine learning architectures} \label{sec:fair}
   In this section, we discuss the approaches adopted to reduce prediction bias in our machine learning framework, which can arise due to
sensitive personal traits such as gender, age, education, and household size~\cite{johnson2020fairkit}.   To reduce the impact of the bias from these traits, we augment the approaches suggested by Johnson et al.~\cite{johnson2020fairkit} and formulate an approach based on hyperparmeter optimization and synthetic minority oversampling~\cite{chakraborty2021bias}.

The voter data collected through the field study is first analyzed for unequal distributions. We observe that the distributions are quite balanced for gender, age groups, and household size, but unbalanced for political orientation and basic education. 
    
As an example, around 78.3\% of the participants are aligned with left-political beliefs, and 66.7\% of the participants are at the highest and second highest levels of education for the actual City Idea voting dataset. We mark the data corresponding to individuals with left political orientation as a privileged group and with right political orientation as a non-privileged group. The same technique is applied to segregate high and low education levels. We randomly sample data separately from these groups, keeping the sample sizes equal, and train a decision tree model to predict the independent variable. We repeat this process for a fixed number of iterations as a stopping criterion and select the final model that achieves the highest recall and the lowest average odds difference~\cite{chakraborty2021bias}.

    Recall is calculated as $TP/(TP+FN)$ where TP is the true positive, FP is the false positive, TN is the true negative, and FN is the false negative. The average odds difference is calculated as the average difference in the false positive rates and true positive rates for the privileged and non-privileged groups. The false positive rate (FPR) is defined as $\mathrm{FPR} = FP/(TP+FN)$, and the true positive rate (TPR) is defined as $\mathrm{TPR} = TP/(FP+TN)$~\cite{chakraborty2021bias}.

    Apart from addressing the biases for sensitive personal traits,  the datasets are  also finally checked for a class-wise imbalance, and synthetic minority oversampling~\cite{chakraborty2021bias} is applied for the classes that still remain a minority. This process is helpful for the actual City Idea voting dataset where the number of unique classes is over 25, and even after mitigating the possible biases in the protected variables using oversampling, some classes remain a minority, which can impact the overall prediction capability of the model~\cite{chakraborty2020fairway}.

    \subsection{Incremental prediction of AI choice consistency based on personal human traits groups}\label{sec:ml}

In the machine learning architecture, personal human traits are used as features, serving as independent variables, while the consistency gain of voters belonging to the abstaining group is treated as the dependent variable. We have experimented with different supervised machine learning models, including decision trees~\cite{de2013decision}, support vector machines~\cite{mammone2009support}, and multilayer perceptrons~\cite{wu2018development}, which do not have long term memory, as well as recurrent neural networks~\cite{mandic2001recurrent}, which have long term memory and process and learn information in short interrelated sequences. This capacity of recurrent neural networks to store and remember interpretations from sets of sequences that correspond to groups of perusal human traits helps to mimic human decision making, which is a function of the traits related to socio-demographic characteristics, preferences, political inclination, etc.~\cite{colombo2020heuristics}. Hence among the machine learning models, recurrent neural networks provide the best prediction performance. 

       Consequently, predicting consistency gain / loss becomes a joint probability distribution function ($\mathbb{P}$) of  the personal human traits of voters:
    
      {\footnotesize
    $$\mathbb{P}(\text{ballot})\ =\ \mathbb{P} (\text{socio-demographics}) \cdot  \mathbb{P}(\text{political interests}) \cdot \mathbb{P}(\text{project preferences})  \cdot \mathbb{P}(\text{outcome expectations})$$
    }
  
    We further test recurrent neural networks with all subsets of the personal human trait groups and hyperparameters, and we observe that holistic integration of all groups (see Table~\ref{tab:set}) provides the best performance. The performance of consistency gain prediction for the abstaining groups and the entire population is enumerated in Table~\ref{tab:result_all}, considering all personal trait groups and recurrent neural networks. The datasets used are obtained from actual and survey voting of the City Idea campaign and the US elections. Our findings indicate that consistency gains can be more accurately predicted for abstaining groups compared to the overall population.

\begin{table}[!htb]
    
     \caption{\textcolor{black}{{{\bf Using all the personal human traits as features helps in achieving the optimum prediction performance of consistencies of AI choices.} Recurrent Neural Networks to predict  (i) the consistency difference between the three abstaining models and their random control and (ii) the (in)consistency of AI representation and transitivity for the whole population.  For each dataset, the prediction metrics shown are averaged across both experiments for the datasets for the abstaining groups and the baseline.
    {\em Parameters of the best model extracted from hyperparameter tuning}: dense layer of 16 neurons; leaky Relu activation function; categorical cross-entropy loss; adam optimiser; synthetic minority oversampling technique to increase 20\% data for all classes; iterations: 600.   }}}
     \centering
    \resizebox{\textwidth}{!}{  
    \begin{tabular}{llllll}
    \hline
    \multicolumn{1}{c} {\bf \textcolor{black}{Personal Human Traits}} & \multicolumn{1}{c} {\bf \textcolor{black}{Model}} & \multicolumn{2}{c}{\bf \textcolor{black}{Survey voting}} & \multicolumn{2}{c}{\bf \textcolor{black}{Actual voting}} \\ \hline
    \multicolumn{2}{c}{} & \multicolumn{1}{c}{\bf \textcolor{black}{F1-score }} 
     & \multicolumn{1}{c}{\bf \textcolor{black}{Accuracy}} &  \multicolumn{1}{c}{\bf \textcolor{black}{F1-score }}  & \multicolumn{1}{c}{\bf  \textcolor{black}{Accuracy}}  \\ 
      \hline

    \multirow{ 3}{*}{\textcolor{black}{All traits}} &
    \textcolor{black}{\texttt{Llama3-8B}} & \textcolor{black}{0.830} & \textcolor{black}{0.836} & \textcolor{black}{0.816} & \textcolor{black}{0.819} \\ 
    & \textcolor{black}{\texttt{GPT3} } & \textcolor{black}{0.820}  & \textcolor{black}{0.799} & \textcolor{black}{0.811}  & \textcolor{black}{0.818} \\ 
    & \textcolor{black}{\texttt{GPT3.5}} & \textcolor{black}{0.845} & \textcolor{black}{0.838} &  \textcolor{black}{0.821} & \textcolor{black}{0.825} \\ \hline
    
    \multirow{ 3}{*}{\textcolor{black}{Socio-demographics and political interests}} &
    \textcolor{black}{\texttt{Llama3-8B}} & \textcolor{black}{0.610} & \textcolor{black}{0.618} &  \textcolor{black}{0.616} & \textcolor{black}{0.613} \\
    & \textcolor{black}{\texttt{GPT3} } & \textcolor{black}{0.642} & \textcolor{black}{0.640}  & \textcolor{black}{0.635}  & \textcolor{black}{0.634} \\ 
    & \textcolor{black}{\texttt{GPT3.5} } & \textcolor{black}{0.612} & \textcolor{black}{0.602}  & \textcolor{black}{0.616}  & \textcolor{black}{0.603} \\ \hline

    \multirow{ 3}{*}{ \textcolor{black}{Socio-demographics, project preferences and outcome expectations}} &
    \textcolor{black}{\texttt{Llama3-8B}} & \textcolor{black}{0.714} & \textcolor{black}{0.719} &  \textcolor{black}{0.698} & \textcolor{black}{0.700} \\
    & \textcolor{black}{\texttt{GPT3} } & \textcolor{black}{0.753} & \textcolor{black}{0.760} & \textcolor{black}{0.712}  & \textcolor{black}{0.721} \\ 
    & \textcolor{black}{\texttt{GPT3.5} } & \textcolor{black}{0.687} & \textcolor{black}{0.679} & \textcolor{black}{0.721}  & \textcolor{black}{0.725} \\ 
    \hline

    \multirow{ 3}{*}{\textcolor{black}{Socio-demographics, political interests and outcome expectations}} &
    \textcolor{black}{\texttt{Llama3-8B}} & \textcolor{black}{0.661} & \textcolor{black}{0.657} & \multirow{ 3}{*} {\textcolor{black}{Only survey voting}} & \\
    & \textcolor{black}{\texttt{GPT3} } & \textcolor{black}{0.685} & \textcolor{black}{0.688}  &  & \\ 
    & \textcolor{black}{\texttt{GPT3.5} } & \textcolor{black}{0.709} & \textcolor{black}{0.715}  & &  \\ \hline

    \multirow{ 3}{*}{\textcolor{black}{Socio-demographics, political interests and project preferences}} &
    \textcolor{black}{\texttt{Llama3-8B}}  & \textcolor{black}{0.672} & \textcolor{black}{0.689} &  \multirow{ 3}{*} {\textcolor{black}{only survey voting}} &\\ 
    & \textcolor{black}{\texttt{GPT3} } & \textcolor{black}{0.646} & \textcolor{black}{0.656} &  &\\ 
    & \textcolor{black}{\texttt{GPT3.5} } & \textcolor{black}{0.6652} & \textcolor{black}{0.663} & &\\ \hline
    
    \end{tabular}
    }
    \label{tab:set}
    
    \end{table}

\begin{table}[!htb]
    
     \caption{\textcolor{black}{{\bf  The performance statistics for every abstaining group and baseline for predicting the consistency of AI choices with respect to human choices and within ballot formats.} The F1-Score reported is based on the experiment conducted using all the traits using recurrent neural networks - dense layer of 16 neurons; leaky Relu activation function; categorical cross-entropy loss; adam optimiser; synthetic minority oversampling technique to increase 20\% data for all classes; epoch: 600.  }}
     \centering
    \resizebox{\textwidth}{!}{  
    \begin{tabular}{l|rrrrrrrrrrrr|rrrrrrrrrrrr}
    \hline
    \multicolumn{25}{c}{\textcolor{black}{\bf Human-AI (F1-Scores)}} \\ \hline
 \multicolumn{1}{c}{} &  \multicolumn{12}{c}{\textcolor{black}{\bf City Idea [Actual]}} &  \multicolumn{12}{c}{\textcolor{black}{\bf City Idea [Survey]}} \\ \hline

\multicolumn{1}{c}{\textcolor{black}{\bf Ballots}} & \multicolumn{3}{c}{\textcolor{black}{\bf Engagement}}  & \multicolumn{3}{c}{\textcolor{black}{\bf Digital literacy}} & \multicolumn{3}{c}{\textcolor{black}{\bf Trust}} & \multicolumn{3}{c}{\textcolor{black}{\bf Baseline}} & \multicolumn{3}{c}{\textcolor{black}{\bf Engagement}}  & \multicolumn{3}{c}{\textcolor{black}{\bf Digital literacy}} & \multicolumn{3}{c}{\textcolor{black}{\bf Trust}} & \multicolumn{3}{c}{\textcolor{black}{\bf Baseline}} \\ \hline

\textcolor{black}{Score} &	\textcolor{black}{0.86} &	\textcolor{black}{0.83} &	\textcolor{black}{0.86} &	\textcolor{black}{0.88} &	\textcolor{black}{0.88} &	\textcolor{black}{0.88} &	\textcolor{black}{0.86} &	\textcolor{black}{0.88} &	\textcolor{black}{0.87} &	\textcolor{black}{0.78} &	\textcolor{black}{0.78} &	\textcolor{black}{0.83} &	\textcolor{black}{0.83} &	\textcolor{black}{0.83} &	\textcolor{black}{0.82} &	\textcolor{black}{0.87} &	\textcolor{black}{0.88} &	\textcolor{black}{0.88} &	\textcolor{black}{0.85} &	\textcolor{black}{0.84} &	\textcolor{black}{0.86} &	\textcolor{black}{0.74} &	\textcolor{black}{0.71} &	\textcolor{black}{0.74} \\ \hline

\textcolor{black}{Approval} &	\textcolor{black}{0.85} &	\textcolor{black}{0.85} &	\textcolor{black}{0.87} &	\textcolor{black}{0.88} &	\textcolor{black}{0.9} &	\textcolor{black}{0.87} &	\textcolor{black}{0.89} &	\textcolor{black}{0.86} &	\textcolor{black}{0.89} &	\textcolor{black}{0.79} &	\textcolor{black}{0.76} &	\textcolor{black}{0.81} &	\textcolor{black}{0.82} &	\textcolor{black}{0.82} &	\textcolor{black}{0.83} &	\textcolor{black}{0.86} &	\textcolor{black}{0.84} &	\textcolor{black}{0.86} &	\textcolor{black}{0.85} &	\textcolor{black}{0.83} &	\textcolor{black}{0.85} &	\textcolor{black}{0.73} &	\textcolor{black}{0.74} &	\textcolor{black}{0.75} \\ \hline

\multicolumn{25}{c}{\textcolor{black}{\bf Within Ballot Formats}} \\ \hline
\textcolor{black}{Single Choice - Score} &	\textcolor{black}{0.83} &	\textcolor{black}{0.83} &	\textcolor{black}{0.82} &	\textcolor{black}{0.87} &	\textcolor{black}{0.83} &	\textcolor{black}{0.83} &	\textcolor{black}{0.85} &	\textcolor{black}{0.84} &	\textcolor{black}{0.86} &	\textcolor{black}{0.74} &	\textcolor{black}{0.71} &	\textcolor{black}{0.74}  &	\textcolor{black}{0.81} &	\textcolor{black}{0.83} &	\textcolor{black}{0.84} &	\textcolor{black}{0.87} &	\textcolor{black}{0.88} &	\textcolor{black}{0.88} &	\textcolor{black}{0.85} &	\textcolor{black}{0.84} &	\textcolor{black}{0.88} &	\textcolor{black}{0.74} &	\textcolor{black}{0.73} &	\textcolor{black}{0.76} \\ \hline

\textcolor{black}{Single Choice - Approval}  &	\textcolor{black}{0.83} &	\textcolor{black}{0.83} &	\textcolor{black}{0.81} &	\textcolor{black}{0.86} &	\textcolor{black}{0.82} &	\textcolor{black}{0.85} &	\textcolor{black}{0.84} &	\textcolor{black}{0.81} &	\textcolor{black}{0.85} &	\textcolor{black}{0.73} &	\textcolor{black}{0.71} &	\textcolor{black}{0.73} &	\textcolor{black}{0.83} &	\textcolor{black}{0.84} &	\textcolor{black}{0.84} &	\textcolor{black}{0.86} &	\textcolor{black}{0.86} &	\textcolor{black}{0.85} &	\textcolor{black}{0.84} &	\textcolor{black}{0.83} &	\textcolor{black}{0.86} &	\textcolor{black}{0.75} &	\textcolor{black}{0.73} &	\textcolor{black}{0.73} \\ \hline
  \multicolumn{25}{c}{\bf \textcolor{black}{US Elections - \texttt{GPT3.5}  = 0.89; \texttt{Llama3-8B} = 0.86; ML= 0.89 (averged over all three years) for single choice ballots}} \\ \hline
    \end{tabular}
    }
    \label{tab:result_all}
    
\end{table}

    \subsection{Explainability of choices} \label{sec:explain}
\begin{figure}[!htb]
        \centering
     \includegraphics[scale = 0.27]{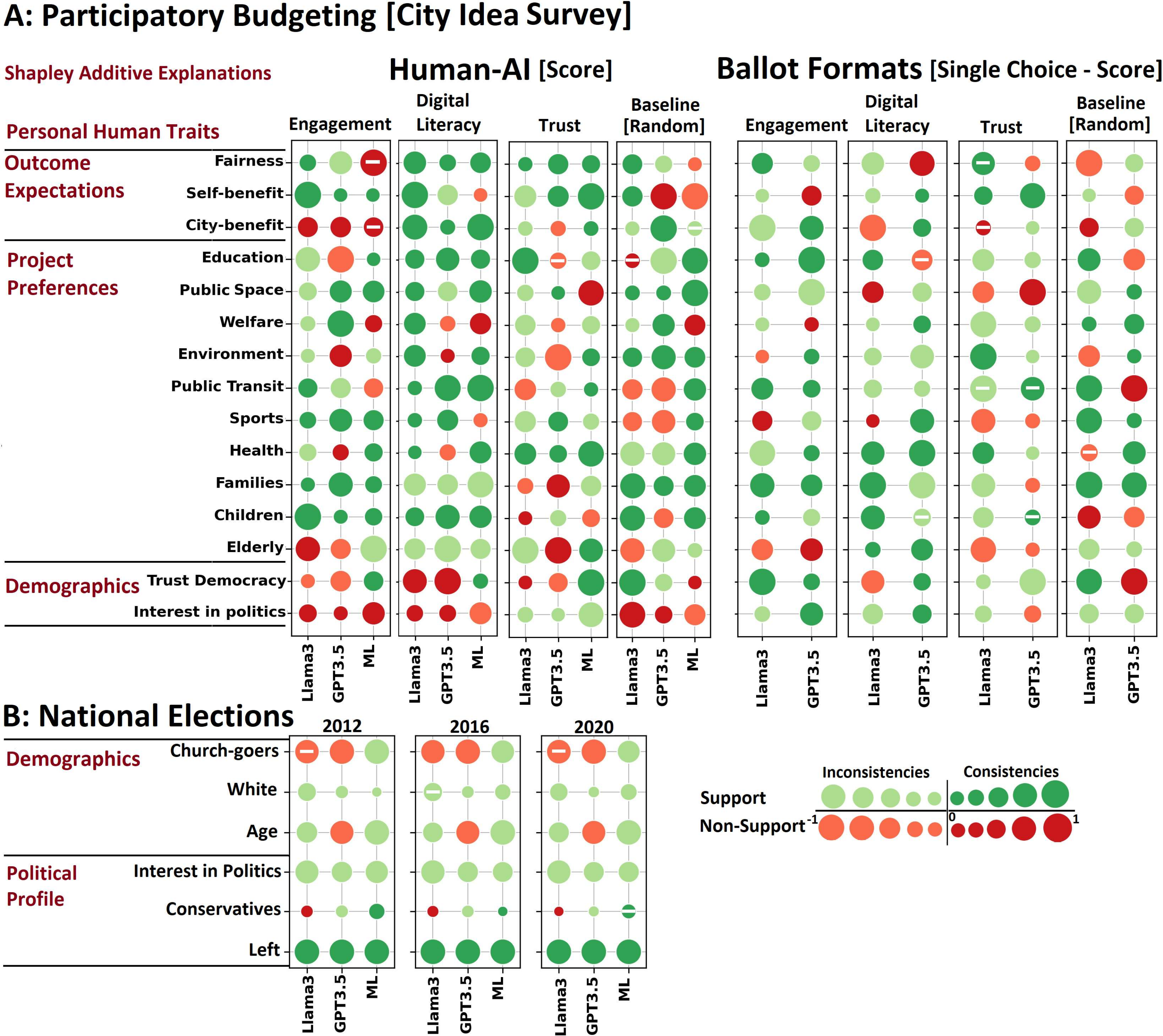}
        \caption{\textcolor{black}{ {\bf 
Compared to an arbitrary abstaining voter, those with low engagement and digital literacy exhibit characteristics that explain the consistency of human-AI representation and ballot formats, for instance, no interest in politics and support to education/health projects related to unconscious and surrogation biases.  Time discounting, affect and conformity biases, such as preference for public space and environmental projects as well as support to families contribute to the consistency of human-AI choice. For US elections, unconscious bias such as political beliefs positively impacts the human-AI consistency.} The relative importance of the personal human traits (y-axis) are shown for the  (A) survey participatory budgeting campaign of City Idea and the (B) US Elections using the size of the bubbles and it is calculated using Shapley Additive Explanations. The AI representation is shown for \texttt{GPT3.5}  and {\texttt{Llama3-8B}  (Llama)} along with the predictive model (ML) (x-axis). The consistency of human-AI representation (score ballots) and ballot formats  (single choice vs. score) is assessed.  For each of these, the personal human traits explain the following: (i) The consistency difference between the three abstaining models and their random control. (ii) The (in)consistency of AI representation and transitivity for the whole population. The `-' sign indicates non-significant values (p>0.05).  }}
        \label{fig:ML_supple}
    \end{figure}

\begin{figure}[!htb]
        \centering
     \includegraphics[scale = 0.245]{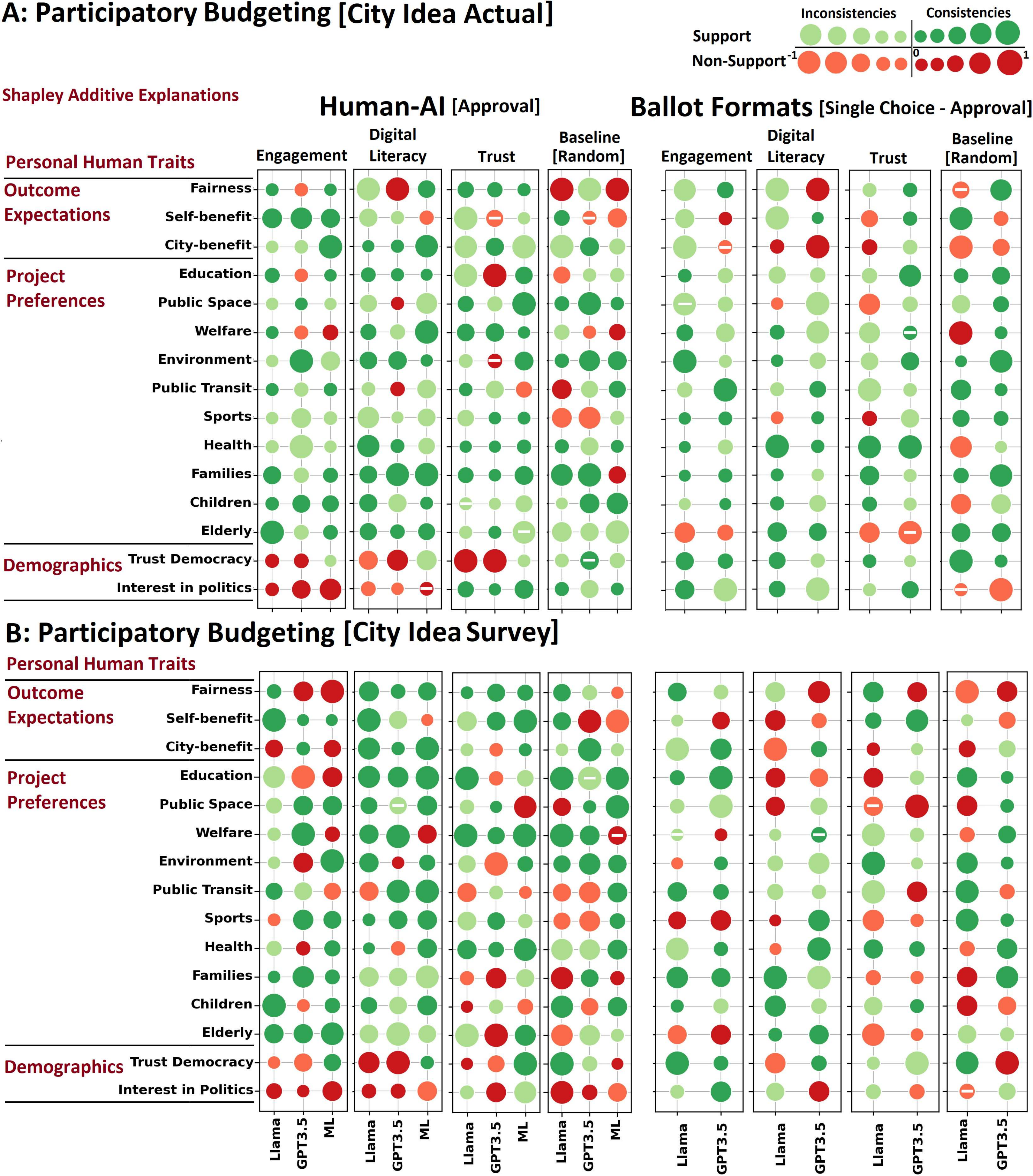}
        \caption{\textcolor{black}{ {\bf 
  Voters with low engagement and digital literacy exhibit traits that explain ballot format consistency and human-AI choices, such as no interest in politics or supporting initiatives with citywide benefits related to unconscious and altruism bias.} The relative importance of the personal human traits (y-axis) for the  (A) actual and (B) survey participatory budgeting campaign of City Idea  for \texttt{GPT3.5}  and {\texttt{Llama3-8B}  (Llama)} along with the predictive model (ML)  (x-axis)  are depicted by the size of the bubbles and it is calculated using { Shapley Additive Explanations}. The consistency of human-AI representation ({approval ballots})  and ballot formats  ({single choice vs. approval} is assessed.  For each of these, the personal human traits explain the following: (i) The consistency difference between the three abstaining models and their random control. (ii) The (in)consistency of AI representation and transitivity for the whole population. The `-' sign indicates non-significant values (p>0.05). }}
        \label{fig:approval}
    \end{figure}

\begin{figure}[!htb]
        \centering
     \includegraphics[scale = 0.225]{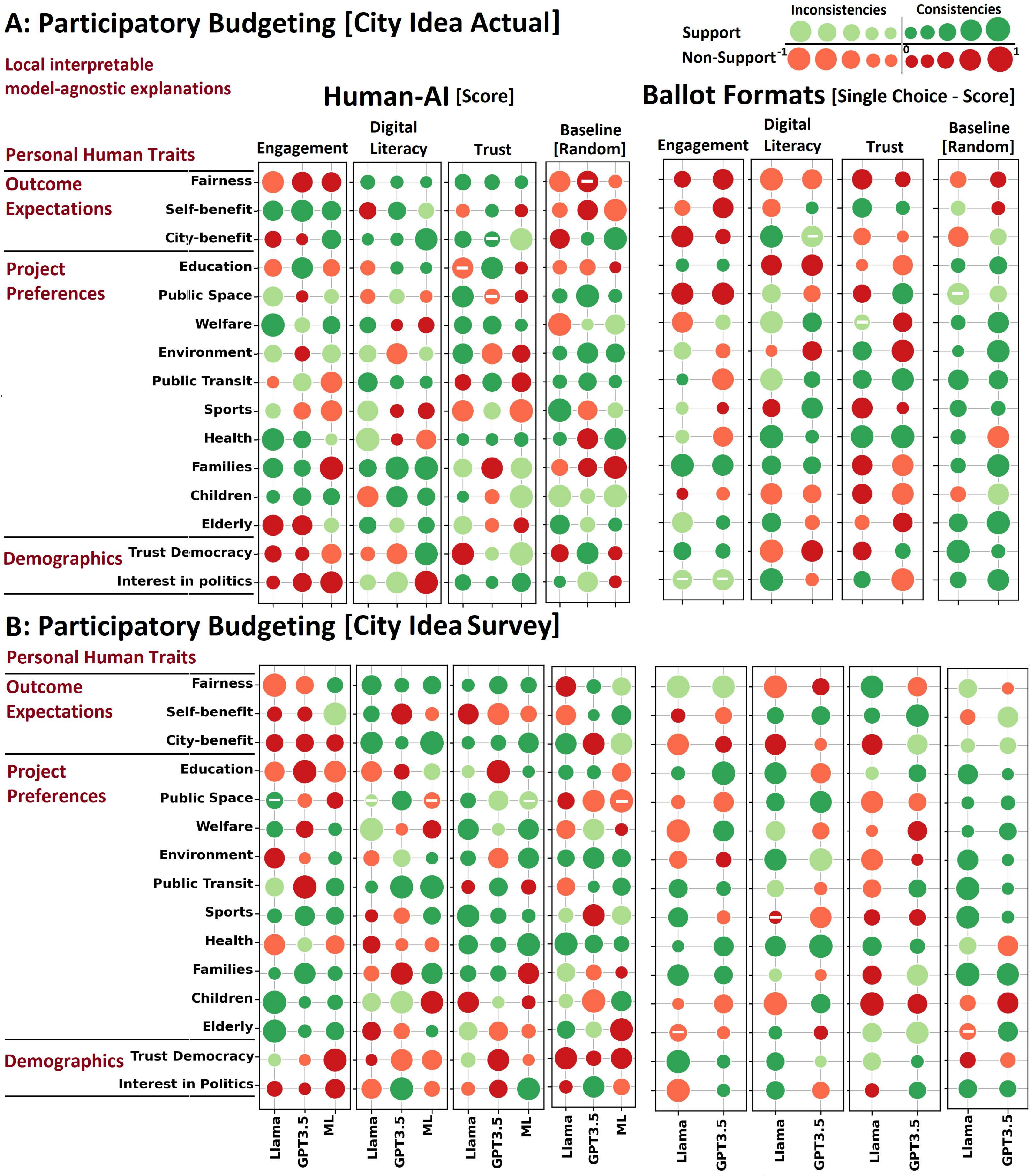}
        \caption{\textcolor{black}{{\bf Voters with low engagement and digital literacy exhibit traits that explain human-AI choices and ballot format consistency, 
        such as no interest in politics or supporting health initiatives related to unconscious and surrogation bias.} The relative importance of the personal human traits (y-axis) for the  (A) actual and (B) survey participatory budgeting campaign of City Idea  for \texttt{GPT3.5}  and {\texttt{Llama3-8B}  (Llama)} along with the predictive model (ML)  (x-axis)  are depicted by the size of the bubbles and it is calculated using {Local Interpretable Model-agnostic Explanations}. The consistency of human-AI representation ({score / cumulative ballots})  and ballot formats  ({single choice vs. score \ cumulative}) is assessed.  For each of these, the personal human traits explain the following: (i) The consistency difference between the three abstaining models and their random control. (ii) The (in)consistency of AI representation and transitivity for the whole population. The `-' sign indicates non-significant values (p>0.05).
 }}
        \label{fig:Lime_score}
    \end{figure}

\begin{figure}[!htb]
        \centering
     \includegraphics[scale = 0.225]{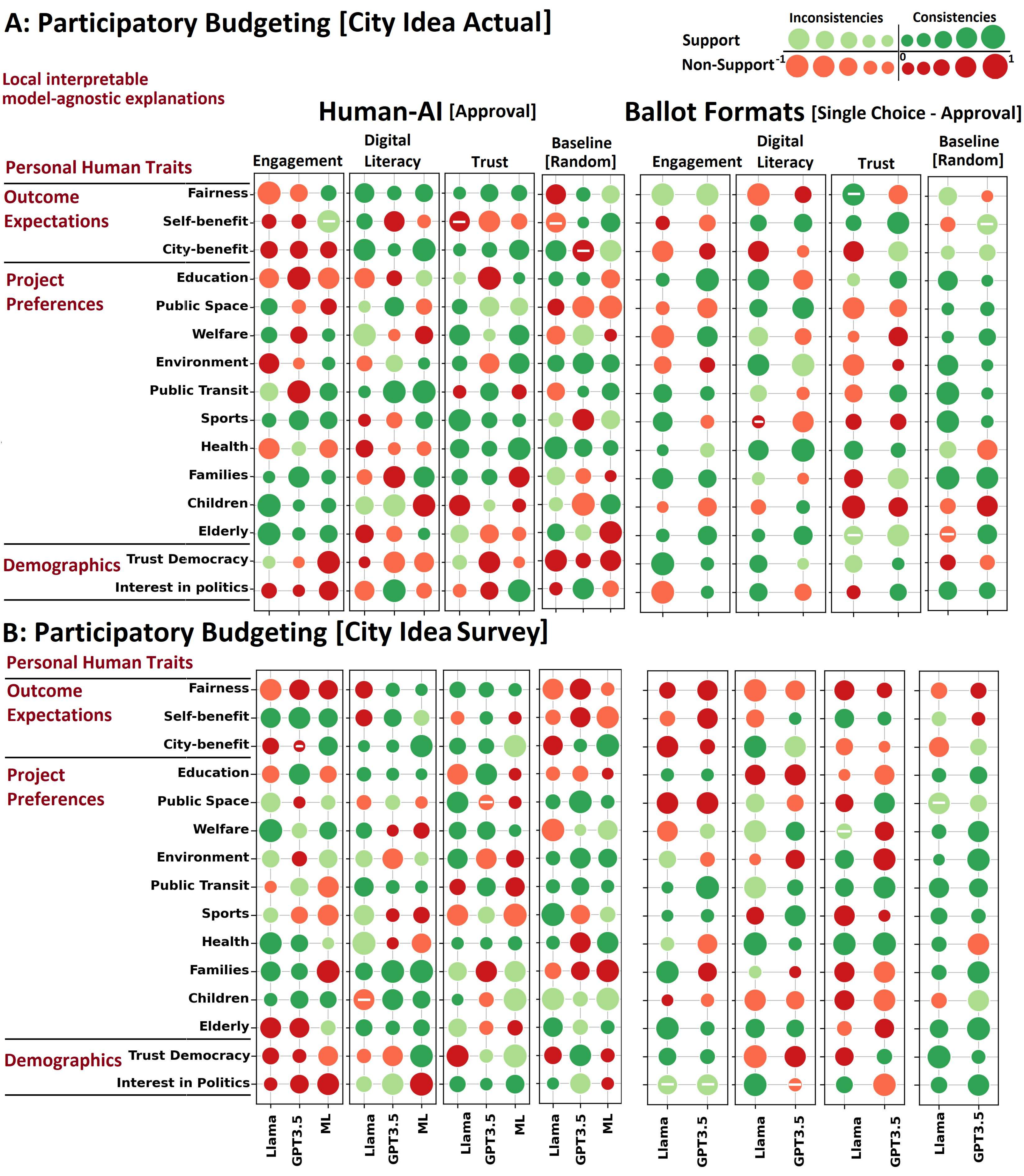}
        \caption{\textcolor{black}{{\bf 
Voters with low engagement and digital literacy exhibit traits that explain human-AI choices and ballot format consistency, 
        such as no interest in politics or supporting health initiatives related to unconscious and surrogation bias.} The relative importance of the personal human traits (y-axis) for the  (A) actual and (B) survey participatory budgeting campaign of City Idea for \texttt{GPT3.5}  and {\texttt{Llama3-8B}  (Llama)} along with the predictive model (ML)  (x-axis)  are depicted by the size of the bubbles and it is calculated using { Local Interpretable Model-agnostic Explanations}. The consistency of human-AI representation ({approval ballots})  and ballot formats  ({single choice vs. approval}) is assessed.  For each of these, the personal human traits explain the following: (i) The consistency difference between the three abstaining models and their random control. (ii) The (in)consistency of AI representation and transitivity for the whole population. The `-' sign indicates non-significant values (p>0.05).  }}
        \label{fig:Lime_approval}
    \end{figure}

\begin{figure}[!htb]
        \centering
     \includegraphics[scale = 0.225]{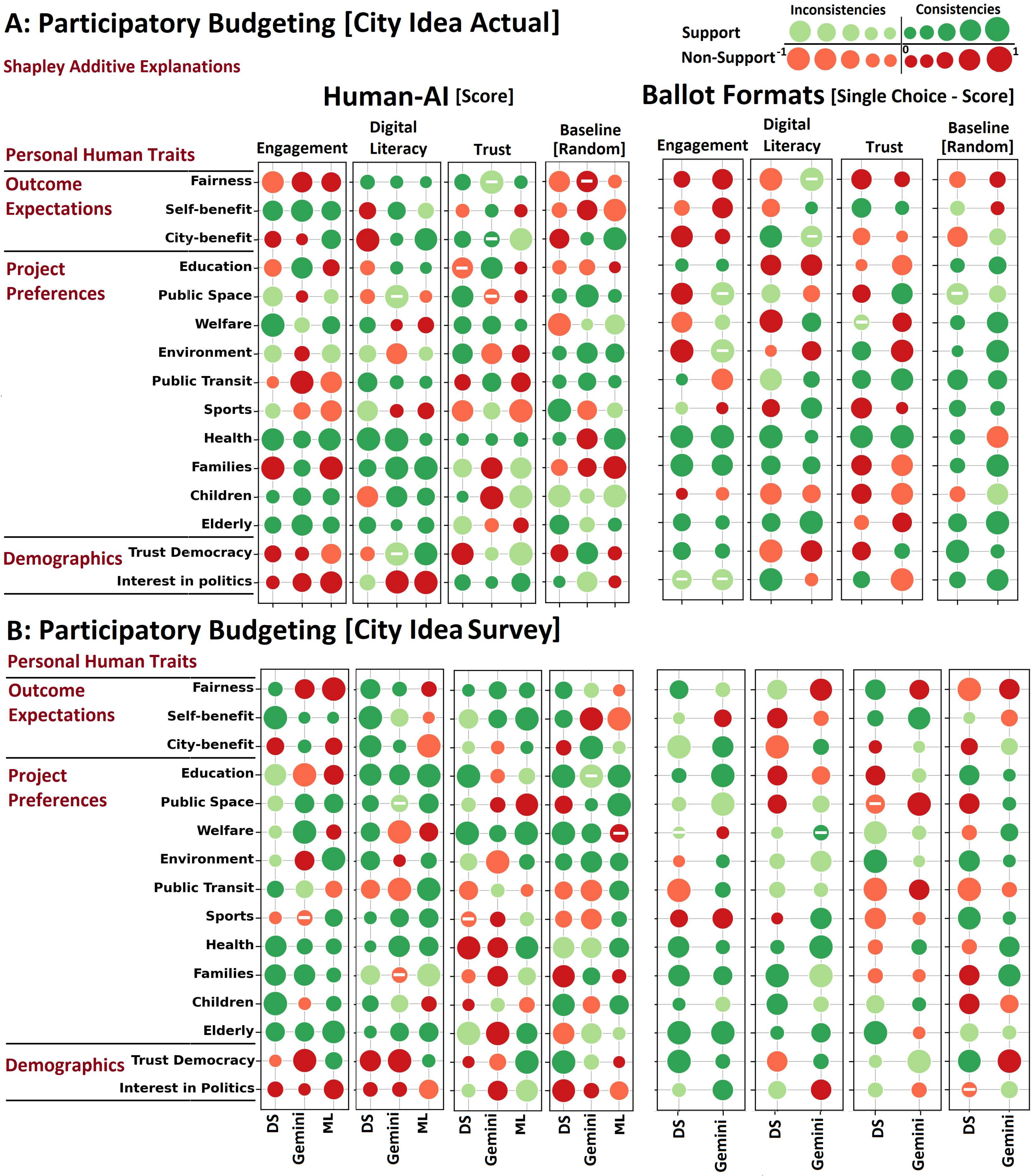}
        \caption{\textcolor{black}{{\bf 
Voters with low engagement and digital literacy exhibit traits that explain human-AI choices and ballot format consistency, 
        such  supporting health and elderly initiatives related to  surrogation bias.} The relative importance of the personal human traits (y-axis) for the  (A) actual and (B) survey participatory budgeting campaign of City Idea for \texttt{Gemini 1.5 Flash}  (Gemini) and {\texttt{Deepseek R1}  (DS)} along with the predictive model (ML)  (x-axis)  are depicted by the size of the bubbles and it is calculated using {Shapley Additive Explanations}. The consistency of human-AI representation ({score ballots})  and ballot formats  ({single choice vs. score}) is assessed.  For each of these, the personal human traits explain the following: (i) The consistency difference between the three abstaining models and their random control. (ii) The (in)consistency of AI representation and transitivity for the whole population. The `-' sign indicates non-significant values (p>0.05).  }}
        \label{fig:score_SHAP_DS_Gem_v1}
    \end{figure}

We causally analyze personal human traits and their contribution to consistency for each voter at the individual level using local explainable AI methods such as Shapley Additive Explanations (SHAP) and Local Interpretable Model Agnostic Explanations (LIME)~\cite{framling2021comparison}, along with a relative analysis across all voters using a global feature ablation study~\cite{hameed2022based}. The findings using SHAP and LIME methods are outlined in Figures 5 (main paper),~\ref{fig:ML_supple},~\ref{fig:approval},~\ref{fig:Lime_score},~\ref{fig:Lime_approval} and~\ref{fig:score_SHAP_DS_Gem_v1} for all types of ballots. The observations from the feature ablation study are detailed in Table~\ref{table:error}. The mapping of relevant personal human traits to cognitive biases is discussed for score or cumulative ballots in Figure 5 (main paper) and for approval ballots in Table~\ref{tab:approval_tab}.

\begin{table}[!htb]
\caption{\textcolor{black}{{\bf Preference for fairness and welfare positively contribute to the Human-AI consistencies for voters with low digital literacy and low trust, respectively. preference for family projects positively contributes to within-ballot format consistencies for all three abstaining populations.} the traits are tested for their relative importance using feature ablation methods~\cite{hameed2022based} to extract the mean decrease in accuracy after removing them from the model. the top 3 important features with high errors and the bottom 2 features with the least errors are noted.}}
\label{table:error}

\centering
\resizebox{\textwidth}{!}{%
\begin{tabular}{llllll}
\toprule
\multicolumn{1}{l}{\textcolor{black}{Model}} &
\textcolor{black}{Top 1} &
\textcolor{black}{Top 2} &
\textcolor{black}{Top 3} &
\textcolor{black}{Bottom 1} &
\textcolor{black}{Bottom 2} \\ \hline

\multicolumn{6}{c}{\textbf{\textcolor{black}{City Idea [Actual] - human AI consistency}}} \\ \hline
\textcolor{black}{Engagement} &
\textcolor{black}{Public transit (0.17)} &
\textcolor{black}{Self benefit (0.15)} &
\textcolor{black}{Children (0.10)} &
\textcolor{black}{Interests in politics (-0.003)} &
\textcolor{black}{Sports (-0.0025)} \\

\textcolor{black}{Digital literacy} &
\textcolor{black}{Families (0.16)} &
\textcolor{black}{Fairness (0.10)} &
\textcolor{black}{Children (0.08)} &
\textcolor{black}{Trust democracy (-0.005)} &
\textcolor{black}{Culture (0.0010)} \\

\textcolor{black}{Trust} &
\textcolor{black}{Welfare (0.14)} &
\textcolor{black}{Fairness (0.12)} &
\textcolor{black}{Health (0.11)} &
\textcolor{black}{Urban greenery (0.002)} &
\textcolor{black}{Sports (-0.001)} \\ \hline

\multicolumn{6}{c}{\textbf{\textcolor{black}{City Idea [Actual] - within ballot formats}}} \\ \hline
\textcolor{black}{Engagement} &
\textcolor{black}{Families (0.09)} &
\textcolor{black}{Education (0.07)} &
\textcolor{black}{Welfare (0.07)} &
\textcolor{black}{Interest in politics (-0.002)} &
\textcolor{black}{Public space (0.004)} \\

\textcolor{black}{Digital literacy} &
\textcolor{black}{Education (0.10)} &
\textcolor{black}{Families (0.05)} &
\textcolor{black}{Health (0.04)} &
\textcolor{black}{Interest in politics (-0.002)} &
\textcolor{black}{Public space (-0.004)} \\

\textcolor{black}{Trust} &
\textcolor{black}{Welfare (0.12)} &
\textcolor{black}{Health (0.11)} &
\textcolor{black}{Families (0.09)} &
\textcolor{black}{Fairness (-0.003)} &
\textcolor{black}{Public space (-0.001)} \\ \hline

\multicolumn{6}{c}{\textbf{\textcolor{black}{City Idea [Survey] - human AI consistency}}} \\ \hline
\textcolor{black}{Engagement} &
\textcolor{black}{Public transit (0.20)} &
\textcolor{black}{Self benefit (0.19)} &
\textcolor{black}{Health (0.18)} &
\textcolor{black}{Interest in politics (-0.004)} &
\textcolor{black}{Urban greenery (-0.003)} \\

\textcolor{black}{Digital literacy} &
\textcolor{black}{City benefit (0.18)} &
\textcolor{black}{Education (0.13)} &
\textcolor{black}{Health (0.11)} &
\textcolor{black}{Public space (0.005)} &
\textcolor{black}{Urban greenery (-0.001)} \\

\textcolor{black}{Trust} &
\textcolor{black}{Welfare (0.18)} &
\textcolor{black}{Health (0.17)} &
\textcolor{black}{Interest in politics (0.16)} &
\textcolor{black}{Elderly (-0.0012)} &
\textcolor{black}{Environment (0.003)} \\ \hline

\multicolumn{6}{c}{\textbf{\textcolor{black}{City Idea [Survey] - within ballot format}}} \\ \hline
\textcolor{black}{Engagement} &
\textcolor{black}{Public space (0.19)} &
\textcolor{black}{Environment (0.17)} &
\textcolor{black}{Families (0.16)} &
\textcolor{black}{Elderly (-0.005)} &
\textcolor{black}{Interests in politics (-0.001)} \\

\textcolor{black}{Digital literacy} &
\textcolor{black}{Families (0.19)} &
\textcolor{black}{Elderly (0.15)} &
\textcolor{black}{Welfare (0.15)} &
\textcolor{black}{Public space (0.006)} &
\textcolor{black}{Interests in politics (0.006)} \\

\textcolor{black}{Trust} &
\textcolor{black}{Families (0.15)} &
\textcolor{black}{Health (0.14)} &
\textcolor{black}{Fairness (0.14)} &
\textcolor{black}{Interests in politics (0.003)} &
\textcolor{black}{Sports (0.002)} \\ \hline

\end{tabular}
}
\end{table}

\begin{table}[ht]
\caption{\textcolor{black}{{\bf Compared to an arbitrary abstaining voter, those with low engagement and digital literacy exhibit characteristics that explain the consistency of human-AI representation and ballot formats, for instance no interest in politics and support to family initiatives corresponding to unconscious and surrogation biases.} The significant biases observed in approval ballots across all abstention models, based on both survey data and actual City Idea campaign, have been aggregated using relative importance scores and significance values. The explainable AI methods used are Shapley Additive Explanations (SHAP) and Local Interpretable Model Agnostic Explanations (LIME)~\cite{framling2021comparison}.}}
\label{tab:approval_tab}

\raggedright
\resizebox{\textwidth}{!}{
\begin{tabular}{lllllll}
\hline
\textcolor{black}{\textbf{Features}} &
\textcolor{black}{\textbf{Relative importance [\%]}} &
\textcolor{black}{\textbf{p-value}} &
\textcolor{black}{\textbf{Explainable AI method}} &
\textcolor{black}{\textbf{Type of consistency}} &
\textcolor{black}{\textbf{Ballot formats}} &
\textcolor{black}{\textbf{[Abstaining models]}} \\
\hline
\textcolor{black}{Not interested in politics} & 14.2 & 0.031 & SHAP & Human - AI & Approval & [Engagement] \\
\textcolor{black}{Not interested in politics} & 12.4 & 0.024 & LIME & Human - AI & Approval & [Engagement] \\
\textcolor{black}{Interested in self benefit} & 16.7 & 0.038 & SHAP & Human - AI & Approval & [Engagement] \\
\textcolor{black}{Interested in self benefit} & 17.3 & 0.041 & LIME & Human - AI & Approval & [Engagement] \\
\textcolor{black}{Support to family initiatives} & 19.2 & 0.002 & SHAP & Ballot formats & Single choice - approval & [Engagement] \\
\textcolor{black}{Support to family initiatives} & 18.8 & 0.045 & LIME & Ballot formats & Single choice - approval & [Engagement] \\
\textcolor{black}{Support to city benefits} & 13.4 & 0.003 & SHAP & Human - AI & Approval & [Digital literacy] \\
\textcolor{black}{Support to city benefits} & 14.6 & 0.004 & LIME & Human - AI & Approval & [Digital literacy] \\
\textcolor{black}{Support to health initiatives} & 12.3 & 0.003 & SHAP & Ballot formats & Single choice - approval & [Digital literacy] \\
\hline
\end{tabular}
}
\end{table}

 \clearpage
    \bibliographystyle{plain}
    \bibliography{sample}
   \makeatletter\@input{xx.tex}\makeatother